\documentclass{article}
\usepackage{times}
\usepackage[letterpaper,textwidth=5.5in,textheight=9in,centering]{geometry}

\usepackage[utf8]{inputenc}
\usepackage[T1]{fontenc}
\usepackage{amsmath,amssymb,amsthm,mathtools}
\usepackage{booktabs}
\usepackage{graphicx}
\usepackage{pgfplots}\pgfplotsset{compat=1.16}
\usepackage{microtype}
\usepackage{enumitem}
\usepackage[ruled,vlined]{algorithm2e}
\usepackage{xcolor}
\usepackage{natbib}
\usepackage{hyperref}

\newtheorem{theorem}{Theorem}
\newtheorem{proposition}{Proposition}
\newtheorem{lemma}{Lemma}
\newtheorem{corollary}{Corollary}
\theoremstyle{definition}
\newtheorem{definition}{Definition}
\newtheorem{example}{Example}
\theoremstyle{remark}
\newtheorem{remark}{Remark}
\newtheorem{openproblem}{Open Problem}

\newcommand{\lhm}{\operatorname{lhm}}
\newcommand{\im}{\operatorname{im}}
\newcommand{\BK}{\mathrm{BK}}
\newcommand{\Om}{\Omega_L}
\newcommand{\Bb}{\mathcal{B}}
\newcommand{\R}{\mathbb{R}}

\newcommand{\E}{\mathbb{E}}

\newcommand{\TP}{T_P}
\newcommand{\reg}{\mathrm{Regret}}

\newcommand{\CL}{C(L)}
\newcommand{\cheat}[1]{\mathrm{cheat}(#1)}
\newcommand{\friendf}[2]{\mathrm{friend}(#1,#2)}

\DeclareMathOperator{\Ldim}{Ldim}
\DeclareMathOperator{\VC}{VC}
\DeclareMathOperator{\TC}{TC}
\DeclareMathOperator{\Pois}{Poisson}
\newcommand{\Tt}{\mathcal{T}}

\usepackage{caption}
\usepackage{tikz}
\usetikzlibrary{arrows.meta}

\title{Optimal Learning of Logical Semantics from Data}

\author{%
  Anagha Tiwari\textsuperscript{1} \quad
  Alexander G.~Gray\textsuperscript{2,3} \quad
  Jonathan Lenchner\textsuperscript{2} \\
  Sasha Rubin\textsuperscript{4} \quad
  Nick Feamster\textsuperscript{1} \quad
  Wojciech Szpankowski\textsuperscript{3,2} \\[6pt]
  {\small \textsuperscript{1}University of Chicago, Department of Computer Science} \\
  {\small \textsuperscript{2}Centaur AI Institute} \\
  {\small \textsuperscript{3}Purdue University, Department of Computer Science} \\
  {\small \textsuperscript{4}University of Sydney, School of Computer Science} \\[6pt]
  {\small \texttt{anaghatiwari@uchicago.edu}, \texttt{alexander.gray@centaurinstitute.org},} \\
  {\small \texttt{jon.lenchner@gmail.com}, \texttt{sasha.rubin@sydney.edu.au},} \\
  {\small \texttt{feamster@uchicago.edu}, \texttt{szpan@purdue.edu}}
}

\date{}
\usepackage{placeins}
\begin{document}
\raggedbottom
\maketitle

\begin{abstract}
A learner that induces logic programs from a template $L$ of $N$ candidate Horn clauses
is handed the $2^N$ clause-sets $S\subseteq L$, but most of them deduce the same facts.
What it can actually learn is the set $\Om$ of realizable deductive closures, the distinct
\emph{meanings} (semantic contents) the template can express. Learning theory takes the
hypothesis class as given; here it is not. $\Om$ exists only implicitly, as the image of a
least-fixpoint map, and its size, its VC and Littlestone dimensions, and even a way to run
the optimal learners over it become available only once it is constructed. We construct it
for two recursive families. For the bandwidth-$d$ reachability template on $n$ atoms the
\emph{semantic capacity} $\CL=\log_2|\Om|$ is $n\log_2\rho_d$ bits, with $\rho_d$ the
$d$-bonacci constant, against about $dn$ syntactic bits; for nonlinear transitive closure on
$m$ constants it is $\tfrac14 m^2(1+o(1))$ bits. Standard finite-class results applied to
$\Om$ then become closed-form rates: Halving makes at most $\CL$ mistakes, the minimax
log-loss (Shtarkov) regret equals $\CL$ once the queries separate the contents, and the
agnostic $0/1$ regret is $\Theta(\sqrt{\CL\,T})$ for both families, whose Littlestone
dimension is of order $\CL$. 
Under random base facts, the size of a
content and its \emph{semantic compression} (how much of a random program its meaning makes
redundant) share a sharp threshold at one base fact per constant: below it the meaning
determines the program up to $O(\log m)$ bits, above it a constant fraction of the program is
redundant. On real data (Padgett's Florentine families and WordNet) the $2^{40}$ clause-sets
of the Florentine template collapse onto $4{,}104$ contents. Under label noise, version-space
pruning loses the target in every run, while exponential weights over $\Om$ stays within its
regret bound. %
\end{abstract}

\vspace{-0.1in}
\section{Introduction}

Inductive logic programming and, more recently, neuro-symbolic learning ask a learner to
choose a set of rules (a hypothesis) from a fixed \emph{template} (language bias) of
candidate clauses. It is tempting to measure the difficulty of this choice by the number
of syntactic hypotheses, $2^N$ for a template of $N$ clauses. This count is misleading,
and the difference matters for semantic information, for code length, and for any
guarantee stated through the parameterization.
The reason is denotational. For a definite (Horn) program the model-theoretic, fixpoint,
and proof-theoretic semantics coincide \citep{vanemden1976semantics,lloyd1987foundations}.
A program's meaning is its \emph{least Herbrand model} and the set of ground atoms it
lets one deduce. Two clause-sets with the same least Herbrand model make identical
predictions on every atom, hence they are semantically indistinguishable; this is what
motivates the notion of semantic information.

Fix a set of background clauses $\BK$ (facts and/or rules)
and a finite \emph{template}
$L=\{c_1,\dots,c_N\}$ of definite (Horn) clauses, the rules the language bias makes
available. A \emph{hypothesis} is a subset $S\subseteq L$; adding it to $\BK$ yields a
definite logic program. Such a program has one well-defined meaning, its least Herbrand model
$\lhm(\BK\cup S)$: exactly the ground atoms one can \emph{deduce} from it.
The object a logic-based learner searches is therefore not the syntax $2^L$ but the
set of \emph{realizable deductive closures}
\[
\mu:2^L\to 2^{\Bb},\qquad R:=\mu(S)=\lhm(\BK\cup S),\qquad
\Om:=\im\mu=\{\lhm(\BK\cup S):S\subseteq L\}\subseteq 2^{\Bb},
\]
where $\Bb$ is the (finite) Herbrand base. 
The map $\mu$ is
many-to-one.
An element of $\Om$ is a \emph{realizable closure}: a set of ground atoms that some hypothesis
actually deduces.
Note the two kinds of object: a hypothesis $S\in2^L$ is a set of \emph{clauses}, its image
$R=\mu(S)\in2^{\Bb}$ a set of \emph{atoms}; $\Om$ collects atom-sets, never rule-sets.
We call the map $\mu$ the \emph{semantics} of the template and $R=\mu(S)$ the \emph{meaning}
(semantic content) of $S$; $\Om$ is the set of meanings the template can express, and $S,S'$ are
\emph{synonyms} if $\mu(S)=\mu(S')$. We call $\CL:=\log_2|\Om|$ the \emph{semantic information},
or \emph{capacity}, of the template.

\paragraph{What is standard and what is new.} Learning theory measures a class by the
distinct functions it contains, so VC dimension, Littlestone dimension and the Shtarkov sum
do not depend on how the class is parameterized. Once $\Om$ is known, the learning results
of Section~\ref{sec:learn} are therefore standard finite-class facts, and we state them only
to make the rates explicit. The difficulty is that a rule learner is never given $\Om$. It is
given the template $L$, i.e.\ the parameterization $2^L$, and $\Om$ exists only as the image
of a least-fixpoint map. Nothing in $L$ reveals $|\Om|$, $\VC(\Om)$ or $\Ldim(\Om)$, and brute
force means running the fixpoint on all $2^N$ clause-sets ($2^{40}$ for the Florentine
template of Section~\ref{sec:exp}, $2^{365}$ for WordNet). Our contribution is to construct
$\Om$ for recursive templates:
\emph{(i)} we characterize membership in $\Om$ by a reachability condition
(Proposition~\ref{prop:realize}) and count $\Om$ exactly: $n\log_2\rho_d$ bits for the
linear bandwidth family (Theorem~\ref{thm:count}) and $\tfrac14m^2(1+o(1))$ bits for nonlinear
transitive closure (Theorem~\ref{thm:nonlinear});
\emph{(ii)} we compute the VC dimension of the bandwidth family and show that its Littlestone
dimension is of order $\CL$ (Corollary~\ref{cor:agnostic-bandwidth}), so the standard bounds
become explicit rates;
\emph{(iii)} we show that the same structure lets Halving, the Bayes mixture and Hedge run
\emph{implicitly} in $O(n)$ time per round, never listing a content
(Appendices~\ref{app:ewa} and~\ref{app:wordnet});
\emph{(iv)} we quantify what a learner pays for placing its prior or description length on
clause-sets rather than contents, which is relevant to Bayesian and MDL program induction and
to differentiable ILP, while VC- and Littlestone-based guarantees are unaffected
(Section~\ref{sec:exp}, Appendix~\ref{app:baseline});
\emph{(v)} under random base facts, we show that the extent of a content, i.e.\ how many facts it
deduces (Theorem~\ref{thm:phase}),
and the semantic compression $H(F\mid R)$, the part of a random program that its meaning makes
redundant (Theorem~\ref{prop:collapse}), undergo the same sharp transition at one base fact per
constant.
We carry one running example, cheating spreading through a class of friends, through all
three models, and test the framework on Padgett's Florentine families and on WordNet,
including two settings, held-out atoms and label noise, where the choice between contents
and clause-sets changes the learner's behaviour (Section~\ref{sec:exp} and Appendix~\ref{app:exps}). Because the paper
uses the vocabulary of both logic programming and learning theory, Appendix~\ref{app:dict} gives a
self-contained primer for readers from either community, with a three-atom example worked through
every quantity.

The classical theory
of semantic information of \citet{carnap1952semantic}, and its modern philosophical
development by \citet{floridi2011philosophy}, sought to measure the \emph{content} of a
proposition rather than the Shannon bits of a signal. Our $\CL$ is a concrete, computable
realization of that intuition for a recursive template: it counts, in bits, how many
distinct \emph{meanings} the template can hold. More directly, this paper can be read as a
combinatorial case study within the program of \citet{lastras2026}, which
incorporates logical reasoning into Shannon's information theory; there, logic enters the
information-theoretic picture in general, whereas here a single recursive family is worked
out end to end, and its semantic information is pinned down exactly. The analytic
machinery consists of rational generating functions, singularity analysis, and precise
minimax redundancy \citep{szpankowski2001average,flajolet2009analytic,drmota2004precise,wu2026foundations}.

\vspace{-0.1in}
\paragraph{Related work}\label{sec:related}
This paper sits at the meeting point of five strands, and its contribution is to connect
them through the single object $\Om$:
{\bf 1.} \textit{Semantic information.} The idea that information has a \emph{content} distinct
from Shannon's signal bits originates with \citet{carnap1952semantic} and is developed
philosophically by \citet{floridi2011philosophy}. That tradition largely measures the
content of a single proposition; we instead count, in bits, how many distinct contents a
recursive template can hold, and identify that count with the capacity of a (noiseless)
closure channel, a learning rate, and a coding redundancy. The most direct antecedent is the program of
\citet{lastras2026} incorporating logical reasoning into Shannon's
theory; our online reading follows \citet{wu2026foundations}.
{\bf 2.} \textit{Logic-program semantics and (probabilistic) ILP.} That a definite program's
meaning is its least Herbrand model --- the coincidence of model-theoretic, fixpoint, and
proof-theoretic semantics --- is classical \citep{vanemden1976semantics,lloyd1987foundations}.
Inductive logic programming \citep{muggleton1991inductive,muggleton1994ilp} learns such programs from examples, its probabilistic extension places a distribution on clauses \citep{sato1995statistical}, its differentiable and neural successors optimize a soft $2^L$ template \citep{manhaeve2018deepproblog}, and the field is surveyed by \citet{cropper2022ilp30}. The PAC-learnability of recursive logic programs was delineated by \citet{cohen1995efficient,cohen1995negative}, who identifies which classes of recursive clauses are efficiently learnable; we instead fix a template and compute the exact size and dimensions of the class it induces.
Our edge-percolation model of Section~\ref{sec:phase} is exactly the distribution semantics
applied to the transitive-closure template. In the differentiable incarnation of ILP
\citep{evans2018learning}, the soft template is precisely this $2^L$ parameterization, so
the collapse ratio $2^N/|\Om|$ is the slack in a guarantee stated in terms of the
parameterization ($N$ versus $\CL$ bits); the realized overpayment of a learner whose prior
lives on clause-sets is quantified in Appendix~\ref{app:baseline}.
What is new here is not the semantics but constructing the quotient of syntax by it, which
is what makes the class computable.
{\bf 3.} \textit{Analytic combinatorics and minimax redundancy.} The exact counts come from
singularity analysis of rational and infinite-product generating functions
\citep{flajolet2009analytic,szpankowski2001average} and the
asymptotic enumeration of partial orders of \citet{kleitman1970topologies,kleitman1975asymptotic}
governing the nonlinear (transitive-closure) variant. The
coding-theoretic reading rests on Shtarkov's normalized maximum likelihood
\citep{shtarkov1987universal} and the precise minimax-redundancy asymptotics of
\citet{drmota2004precise,jacquet2004markov,rissanen1996fisher,clarke1990information};
$\CL=\log_2|\Om|$ is the Shtarkov sum of the deductive concept class, and its parametric
refinement is a Markov-source redundancy in that line.
{\bf 4.} \textit{Online learning and concept classes.} Reading contents as classifiers puts the
template in the standard mistake-bound and regret framework
\citep{littlestone1988learning,cesabianchi2006prediction,cover2006elements}: Halving gives
the $\log_2|\Om|$ mistake bound, and the Littlestone dimension governs the optimal
agnostic regret, which is $\Theta(\sqrt{\Ldim\,T})$ \citep{bendavid2009agnostic,alon2021adversarial}.
These results apply to any finite class. Our point is that they become \emph{closed-form rates}
in $\rho_d$ and $\varphi$ once $\Om$ is constructed, and that a mixture whose prior is
uniform over clause-sets is minimax-suboptimal, by a gap linear in $n$
(Appendix~\ref{app:baseline}).
{\bf 5.} \textit{Random graphs and phase transitions.} The extent facet reduces to the transitive
closure of a random digraph, whose sharp threshold is Karp's \citep{karp1990transitive},
via the branching-process couplings of \citet{athreya1972branching,vanderhofstad2017random}.
We use that threshold to describe \emph{semantic extent} and, through $H(F\mid R)$,
\emph{semantic compression}.

\vspace{-0.11in}
\section{Preliminaries: realizable closures}\label{sec:object}
\vspace{-0.09in}
Fix a finite set of predicate symbols, each with a fixed arity, a set of constant symbols, and a set of variables, and assume there are no function symbols of positive arity. An \emph{atom} is an expression $P(t_1,\ldots,t_r)$, where $P$ is a predicate symbol of arity $r$ and each $t_i$ is either a constant symbol or a variable; it is a \emph{ground atom} if it contains no variables, and sometimes simply called a \emph{fact}. The \emph{Herbrand base} $\mathcal{B}$ is the set of all ground atoms. A \emph{definite clause} is an expression $h \leftarrow b_1,\ldots,b_k$, where $h,b_1,\ldots,b_k$ are atoms; $h$ is the \emph{head} and $b_1,\ldots,b_k$ form the \emph{body}. A \emph{program} is a finite set of definite clauses.
The immediate-consequence operator $\TP:2^{\Bb}\to2^{\Bb}$ of a program $P$ fires,
in one step, every rule whose body already holds:
$\TP(M)=\{h:(h\leftarrow b_1,\dots,b_k)\in P,\ \{b_1,\dots,b_k\}\subseteq M\}$. Its iterates
from $\emptyset$ increase to the least model $\lhm(P)=\bigcup_{k\ge0}\TP{\uparrow}k$, the
set of atoms $P$ deduces \citep{vanemden1976semantics,lloyd1987foundations}. 
See Appendix~\ref{app:dict} for more explanations.

We work with a set of ground background clauses $\BK$
(ground
facts and/or rules) 
and a finite template
$L=\{c_1,\dots,c_N\}$ of definite clauses. A hypothesis is a subset $S\subseteq L$;
adjoining it to $\BK$ gives a definite program whose meaning is the least Herbrand model
$\lhm(\BK\cup S)\subseteq\Bb$ with $\Bb$ being the Herbrand base. Throughout, we keep the two kinds of object apart: a
\emph{hypothesis} $S\in 2^L$ is a set of clauses (rules the learner switches on); its image
$R:=\mu(S)\in\Om$ is a set of \emph{atoms} (the facts those rules deduce), $R\subseteq\Bb$.
So $\mu$ maps rule-sets to atom-sets, and $\Om$ collects the atom-sets, never the
rule-sets. We often split $\BK$ into a fixed \emph{structural} part and a set
$E$ of declared \emph{evidence} (seed) facts.

\emph{Switchable rules vs.\ switchable facts.} A base fact is a Horn clause with empty body, so
the switchable elements of $L$ may be rules or facts, and $L$ is uniformly a set of Horn clauses.
In the bandwidth family (Section~\ref{sec:count}) the switchable elements are rules, and we write a
hypothesis as $S\subseteq L$; in the transitive-closure family (Theorem~\ref{thm:nonlinear},
Section~\ref{sec:phase}) the composition rule is fixed in $\BK$ and the switchable elements are base
facts, written $F\subseteq L$. The maps $\mu$, $\Om=\im\mu$ and $\CL$ are defined identically in
both cases, so ``$S$ random'' means random rules and ``$F$ random'' means random facts.
For the reachability templates below, every clause has the
shape $r_v\leftarrow r_u$ (a plain edge $u\to v$).

\begin{proposition}[Realizability characterization]\label{prop:realize}
Let the template consist of edge clauses $r_v\leftarrow r_u$ over a directed graph
$G=(V,E)$, with $\BK$ declaring a source set $V_0\subseteq V$ true. Then $R\subseteq V$ is
realizable ($R\in\Om$) if and only if $V_0\subseteq R$ and every vertex of $R$ is reachable
from $V_0$ along a path lying entirely inside $R$.
\end{proposition}

\begin{example}[Cheating in a class]\label{ex:main}
Five students $A,B,C,D,F$ and two predicates, $\mathrm{cheat}/1$ and $\mathrm{friend}/2$,
give the Herbrand base
$
\Bb=\{\cheat{v}:v\in\{A,B,C,D,F\}\}\ \cup\ \{\friendf{u}{v}:u,v\in\{A,B,C,D,F\}\},
$
so that
$|\Bb|=5+25=30$,
where $u,v$ range over the five students. The background holds a friendship graph (each tie declared
in \emph{both} directions, so the relation is symmetric) together with the one student
caught cheating --- the \emph{evidence} (seed) $E=\{\cheat{B}\}$ --- and the template is the
single spreading schema $\cheat{Y}\leftarrow\cheat{X},\friendf{X}{Y}$:
\begin{align*}
\BK&=\underbrace{\{\,\friendf{X}{Y},\friendf{Y}{X}\ :\ (X,Y)\in\{BA,\,BD,\,AC,\,DC,\,CF\}\,\}}_{\text{structure (each tie, both directions)}}\ \cup\ \underbrace{E}_{\text{evidence}},\\[2pt]
L&=\bigl\{\,\cheat{Y}\leftarrow\cheat{X},\friendf{X}{Y}\,\bigr\}\qquad(N=|L|=10\ \text{ground clauses}).
\end{align*}
The $25$ $\mathrm{friend}$-atoms are fixed background: every content contains exactly the ten
declared ties and no other $\mathrm{friend}$-atom, so contents differ only in their
$\mathrm{cheat}$-atoms, the content sub-base on which the learner is queried (a convention we keep throughout the paper). A hypothesis $S\subseteq L$ switches on a set of
\emph{conducting} friendships; its meaning $\mu(S)=\lhm(\BK\cup S)\in\Om$ is the set of
students it concludes cheat. Taking
$S=\{\,\cheat{A}\!\leftarrow\!\cheat{B},\friendf{B}{A};\ \ \cheat{C}\!\leftarrow\!\cheat{A},\friendf{A}{C}\,\}\subseteq L$,
each ground body atom $\friendf{B}{A},\friendf{A}{C}\in\BK$ holds, so the fixpoint iterates
are $\{B\},\{B,A\},\{B,A,C\}$ and $R=\mu(S)=\{B,A,C\}$.
By Proposition~\ref{prop:realize}, 
$\mu$ ranges over all $2^{10}=1024$ hypotheses, but
{\small
\[
\Om=\bigl\{\{B\},\{BA\},\{BD\},\{BAC\},\{BAD\},\{BCD\},\{BACD\},\{BACF\},\{BCDF\},\{BACDF\}\bigr\}
\]
}
so $|\Om|=10$. So $\Om$ collects the deductive closures and only
those --- a strict, structured subset of the atom sets, here a lattice ($\{BA\}$ and
$\{BD\}$ are incomparable: two different outbreaks).
\end{example}

Finally, we define three quantities, none needing probability yet. This will fix
the vocabulary.

\begin{definition}[Semantic content, extent, capacity]\label{def:cec}
For a hypothesis $S\subseteq L$: its \emph{semantic content} is the deductive closure
$\mu(S)=\lhm(\BK\cup S)\in\Om$; its \emph{semantic extent} is the number of new facts that
content adds, $E(S):=|\lhm(\BK\cup S)|-|\lhm(\BK)|$. The \emph{semantic capacity} of the
template is $\CL:=\log_2|\Om|$ bits --- the number of bits needed to name which content was
produced.
\end{definition}

\begin{remark}[Why ``capacity'']
The closure map
$\mu:2^{L}\to\Omega_L$, $S\mapsto R=\mu(S)$, is a deterministic channel, so any
prior $P$ on the clause-sets pushes forward to $\pi=\mu_*P$ on $\Omega_L$ with
$I(S;R)=H(R)$ and $H(S)=H(R)+H(S\mid R)$. The conditional entropy $H(S\mid R)$ (Shannon's
\emph{equivocation}: the uncertainty about a channel's input that remains after seeing its output)
is exactly the syntactic redundancy that $\mu$ compresses away; we call it the \emph{semantic
compression gain}. Maximizing over priors gives
$\sup_P I(S;R)=\log_2|\Omega_L|=C(L)$, uniquely attained by the content-uniform
prior, under which the self-information $-\log_2\pi(R)\equiv C(L)$ is an exact
equipartition over realizable theories; a generic prior falls short of capacity by
$D(\pi\,\|\,\mathrm{unif}_{\Omega_L})$. The channel is noiseless, so its capacity is
simply the Hartley count $\log_2|\Om|$. What the channel view adds is the decomposition
$H(S)=H(R)+H(S\mid R)$: under random base facts the semantic compression gain $H(S\mid R)$
undergoes the same phase transition as the extent (Theorem~\ref{prop:collapse}); see
Appendix~\ref{sec:info}. %
\end{remark}

\vspace{-0.1in}
\section{Combinatorial Model: Counting the contents}\label{sec:count}
\vspace{-0.1in}

\newcommand{\Ld}{L_{n,d}}
\newcommand{\OL}{\Omega_L}

In this section we study a simple template $L$ for which the quantities defined above can be computed exactly.
 The atoms are $r_0,\dots,r_n$; they \emph{are} the Herbrand base $\Bb$, and $\BK=\{r_0\}$ declares the source $r_0$. The
bandwidth-$d$ template is
\[
L_{n,d}=\{\,c_{i\to j}:r_j\leftarrow r_i,\ 0\le i<j\le n,\ j-i\le d\,\},\qquad
N=|L_{n,d}|=dn-\tbinom{d}{2}.
\]
This is the recursive path (transitive-closure) template of ILP, restricted to steps of
length at most $d$. The \emph{bandwidth} $d$ is the largest index distance a single rule can
bridge: rule $c_{i\to j}$ passes a conclusion from atom $r_i$ only to one of
$r_{i+1},\dots,r_{i+d}$. Equivalently, $d$ is the bandwidth of the adjacency matrix of the
rule graph. It is the locality knob of the language bias: $d=1$ allows only a chain, and
$d=n$ allows every forward rule. 
In other words, $L_{n,d}$ is a set of edge-clauses: reading $r_j\leftarrow r_i$ as a directed edge
$i\to j$ makes $\Ld$ a graph on the atoms. A hypothesis is a subset
$S\subseteq\Ld$, a choice of which edges are \emph{switched on}, so $S$ lists the edges that conduct, and there
are $2^N$ such choices.
A hypothesis does not stand on
its own: one adjoins it to $\BK$ and fires the immediate-consequence operator
to a fixpoint, starting from the source. The result is the content
$
  R=\mu(S)=\lhm(\BK\cup S)\ \in\ \OL, r_0\in R,
$
the set of atoms reachable from $r_0$ along the switched-on edges of $S$. So
$S$ is a set of \emph{edges} (in $2^{\Ld}$) and $R$ is a set of \emph{atoms}
(in $2^{\{0,\dots,n\}}$).

\begin{example}[The cheating class, seated in a row] Example~\ref{ex:main} is an instance:
seat the class in a line and let friendship mean ``within $d$ seats.'' Each tie then licenses
one ground rule $\mathrm{cheat}(Y)\leftarrow\mathrm{cheat}(X),\mathrm{friend}(X,Y)$ --- one
edge of $G$ between $\mathrm{cheat}$-atoms --- and the abstract family absorbs the fixed tie
into that edge, writing it as the plain $r_j\leftarrow r_i$ with $r_i$ in the role of
$\mathrm{cheat}(\text{student }i)$.
\end{example}

Write $\Om(n,d):=\Om(L_{n,d})$. We represent a content by the indices of its atoms,
$R=\{0=x_0<x_1<\cdots<x_m\}\subseteq\{0,\dots,n\}$, where $x_\ell$ is the index of the
atom $r_{x_\ell}\in R$; reachability from $r_0$ along edges of length at most $d$ forces
every consecutive gap $x_\ell-x_{\ell-1}\le d$. In view of this, $\Om(n,d)$ is the set of
vertex subsets of $\{0,\dots,n\}$ that contain the source and have no gap exceeding
$d$---not edge-sets, and not arbitrary vertex-sets.
Grouping the realizable sets by their top element $x_m=x$
identifies each with a composition of the integer $x$ into parts in $\{1,\dots,d\}$ (the
gaps). This gives $|\OL(n,d)|=\sum_{x=0}^{n}C_d(x)$, with $C_d(x)$ the number of such
compositions, so that
$\sum_x C_d(x)t^x=(1-(t+\cdots+t^d))^{-1}=(1-t)/(1-2t+t^{d+1})$ \citep{flajolet2009analytic, szpankowski2001average}. This leads to our main result of this  section.

\begin{theorem}[Exact count and precise asymptotics]\label{thm:count}
The counts $a_n=|\Om(n,d)|$ have rational generating function
\[
\sum_{n\ge0}a_n t^n=\frac{1}{1-2t+t^{d+1}},\qquad\text{equivalently}\qquad
a_n=2a_{n-1}-a_{n-d-1}\ (n\ge d+1),
\]
with $a_k=2^k$ for $k\le d$. As $n\to\infty$ with $d\ge2$ fixed,
$a_n=A_d\,\rho_d^{\,n}\bigl(1+O(\gamma_d^{\,n})\bigr)$, where $\rho_d\in(1,2)$ is the
$d$-bonacci constant (the unique root in $(1,2)$ of
$\rho^d=\rho^{d-1}+\cdots+\rho+1$), $A_d$ is an explicit constant, and $\gamma_d<1$. In
particular for $d=2$,
\[
\rho_2=\varphi=\tfrac{1+\sqrt5}{2},\qquad A_2=\frac{\varphi^3}{\sqrt5},\qquad
a_n=F_{n+3}-1\sim\frac{\varphi^{\,n+3}}{\sqrt5}.
\]
\end{theorem}

\begin{corollary}[Collapse ratio and semantic capacity]\label{cor:collapse}
With $N=dn+O(1)$ clauses,
\[
\frac{2^N}{|\Om(n,d)|}=\Bigl(\frac{2^d}{\rho_d}\Bigr)^{n}2^{O(1)},\qquad
\frac{\log_2|\Om(n,d)|}{n}\xrightarrow[n\to\infty]{}\log_2\rho_d\in(0,1).
\]
The template carries $\log_2\rho_d$ bits of content per atom, strictly less than the $d$
bits of syntax per atom; the deficit $d-\log_2\rho_d$ is the redundancy the language bias
builds in by letting several rules compute the same reachability. As $d\to\infty$,
$\log_2\rho_d\uparrow1$: content saturates at one bit per atom.
\end{corollary}

\textbf{The nonlinear jump.} The bandwidth family is \emph{linearly}
recursive (one recursive atom per body), and its count is a one-dimensional
transfer-matrix object with the clean growth constant $\rho_d$. The canonical
ILP target---$\mathrm{path}$, $\mathrm{ancestor}$, transitive closure---is
instead \emph{nonlinear}: genuine transitive closure through the
doubly-recursive composition rule
$\mathrm{tc}(X,Z)\leftarrow\mathrm{tc}(X,Y),\mathrm{tc}(Y,Z)$, whose body
carries two recursive atoms. Here the fixed and switchable parts of the
program trade places. In the bandwidth family $\mathrm{BK}$ is a fact (the
source $r_0$) and $L$ is a set of rules; now $\mathrm{BK}$ is the single
composition rule above---the recursion is fixed, not learned---and $L$ is the
set of ground base facts
$
L=\{\mathrm{tc}(i,j): 1\le i<j\le m\}, ~ N=|L|=\binom{m}{2},
$
each a body-empty Horn clause (cf.\ Section~\ref{sec:object}). A hypothesis is a fact-set
$F\subseteq L$ (which edges $\mathrm{tc}(i,j)$ are declared; we write $F$,
not $S$, since the switchable clauses are facts). The content sub-base is
$\mathcal{B}=L$, the same set of atoms, and the content is
$R=\mu(F)=\mathrm{lhm}(\mathrm{BK}\cup F)=\mathrm{TC}(F)$, obtained by firing
the composition rule on $F$ to a fixpoint; there is no source atom, since
$\mathrm{TC}(F)$ needs none. Read as a graph on $[m]$, a declared fact $\mathrm{tc}(i,j)\in F$ is an
edge $i\to j$, while a derived atom $\mathrm{tc}(i,j)\in R$ is a path from $i$ to $j$: $F$ is a directed
graph and $R$ its transitive closure. Holding the rule fixed and exposing the base
relation is exactly the distribution-semantics reading \citep{sato1995statistical} that
Section~\ref{sec:phase} randomizes. Every transitively closed $R$ satisfies
$\mathrm{TC}(R)=R$, so $R=\mu(R)\in\Om$; hence $\Om$ is precisely the set of
transitively closed subrelations of $\{(i,j): i<j\}$. For a proof of the next result see Appendix~\ref{app:nonlinear}.

\begin{theorem}[Nonlinear capacity: the transitive-closure jump]\label{thm:nonlinear}
In the ordered reading the contents $\Om$ are the strict partial orders on
$[m]$ admitting the identity as a linear extension. Their number
$T(m):=|\Om|$ satisfies
\[
2^{\lfloor m^2/4\rfloor}\ \le\ T(m)\ \le\ 2^{\,m^2/4+o(m^2)},
\qquad\text{so}\qquad
\kappa:=\lim_{m\to\infty}\frac{\log_2 T(m)}{m^2}=\frac14,
\]
and $\CL=\log_2 T(m)=\tfrac14 m^2(1+o(1))$ bits. The same $\kappa=\tfrac14$
governs the unordered variant (base facts $\mathrm{tc}(i,j)$ for all $i\ne j$), whose contents
are in bijection with the preorders on $[m]$. Thus nonlinear recursion has quadratic capacity
$\Theta(m^2)$, against the linear $\Theta(n)$ of the bandwidth family.
\end{theorem}
\vspace{-0.05in}

\vspace{-0.12in}
\section{Probabilistic Model: A phase transition in semantic content}\label{sec:phase}
\vspace{-.1in}

We now switch to a probabilistic model and turn to the semantic extent, that is, how \emph{much} a single content deduces. The natural random model is the
distribution semantics of probabilistic logic programs \citep{deraedt2015probabilistic}:
fix the recursive rule as background, select each candidate base fact independently, and
recompute the least model. The random object is thus the \emph{graph} of selected base
facts, never the rule.
Throughout this section $\mathrm{BK}$ and $L$ are fixed; the random object is
the hypothesis. In the transitive-closure template (Theorem~\ref{thm:phase})
$\mathrm{BK}$ is the composition rule, $L$ is the set of all candidate base
facts $\mathrm{tc}(i,j)$, $i\neq j$, over $m$ constants, and $F\subseteq L$
switches each fact on independently with probability $p$. 

With each base fact selected with probability
$p$ (Erd\H{o}s--R\'enyi model), the selected facts $F$ form a random digraph $\mathcal D(m,p)$, and the content
$R=\mu(F)=\TC(F)$ is its transitive closure. The semantic extent of the draw
(Definition~\ref{def:cec}) is $E(F)=|R|=\#\{(i,j):i\rightsquigarrow j\}$, and $\E|R|$ is the average semantic extent.

By symmetry $\E|R|=m\cdot\E|R_0|$ up to the
diagonal, where $R_0=\{v:0\rightsquigarrow v\}$ is the set reachable from one constant.
Exploring $R_0$ by breadth-first search, each newly reached vertex has
$\mathrm{Binomial}(m-1,p)$ out-neighbours; in the sparse regime $p=c/m$ this is
asymptotically $\Pois(c)$, so $R_0$ is the total progeny of a Poisson($c$)
branching process \citep{karp1990transitive,athreya1972branching}. Branching processes
have a threshold at mean offspring $c=1$, and it transfers verbatim to the deductive
closure. Detailed proof is in Appendix~\ref{app:th2}.

\begin{theorem}[Phase transition of semantic extent]\label{thm:phase}
Let the base facts be $\mathcal D(m,c/m)$ and let $\zeta=\zeta(c)$ be the unique positive
root of $1-\zeta=e^{-c\zeta}$ (so $\zeta=0$ for $c\le1$ and $\zeta>0$ for $c>1$). As
$m\to\infty$,
\[
\E|R_0|\longrightarrow
\begin{cases}\dfrac{1}{1-c}, & c<1,\\[6pt]\zeta(c)^2\,m\,(1+o(1)),& c>1,\end{cases}
\qquad
\E|R|=
\begin{cases}\Theta(m),& c<1,\\[3pt]\zeta(c)^2\,m^2\,(1+o(1)),& c>1.\end{cases}
\]
The average semantic content jumps from linear to quadratic in $m$ at the critical density
$c=1$: below it a random program deduces a bounded number of facts per constant, and its
total content is $\Theta(m)$; above it, the content is a fixed positive fraction
$\zeta(c)^2$ of all $m^2$ possible facts; at $c=1$ exactly, $\E|R|=\Theta(m^{4/3})$.
\end{theorem}

The bandwidth-$d$ template of Section~\ref{sec:count} has no such transition (proof in Appendix~\ref{app:cor2}).

\begin{corollary}[Bounded reach cannot percolate]\label{cor:subcritical}
For the bandwidth-$d$ template of Section~\ref{sec:count} (each atom has at most $d$
in-neighbours, edges on with probability $\theta<1$), the reachable set is a.s.\ finite and,
uniformly in $n$,
\[
\E|R_0|\ \le\ 1+d\,(1-\theta)^{-d(d+1)/2}.
\]
There is no critical $\theta<1$.
\end{corollary}

Corollary~\ref{cor:subcritical} is the
correct reading of the ``bounded content'' phenomenon: a bounded-reach recursive template
is a quasi-one-dimensional reachability process, which in one dimension cannot percolate,
so its semantic extent stays bounded for every fixed reliability $\theta<1$. The bound grows
like $(1-\theta)^{-d(d+1)/2}$ as $\theta\to1$, since emptying a window requires all
$d(d+1)/2$ edges out of it to be off; the exact chain gives $\E|R_0|\approx9.9\times10^5$ for
$d=2$ and $9.9\times10^{11}$ for $d=3$ at $\theta=0.99$, so this order is attained. The phase
transition appears only once rules reach across an unbounded number of constants (the
mean-field regime). The two results together locate the boundary exactly: \emph{semantic
content is a critical phenomenon}, with critical density one base fact per constant, and
bounded-reach language biases sit permanently on its trivial side.

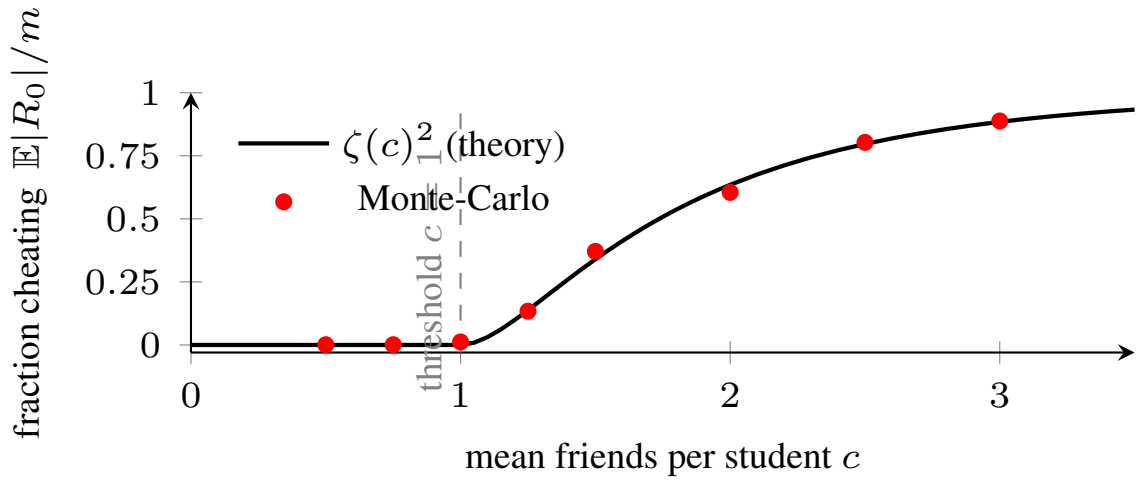
\begin{figure}[t]
\centering
\begin{tikzpicture}
\begin{axis}[
  width=0.56\linewidth, height=3.3cm,
  xlabel={mean friends per student $c$},
  ylabel={fraction cheating $\ \E|R_0|/m$},
  xmin=0, xmax=3.5, ymin=-0.03, ymax=1.0,
  xtick={0,1,2,3}, ytick={0,0.25,0.5,0.75,1.0},
  tick label style={font=\scriptsize}, label style={font=\scriptsize},
  legend style={font=\scriptsize, at={(0.03,0.97)}, anchor=north west, draw=none, fill=none},
  axis lines=left, clip=false,
]
\addplot[no marks, thick, black] coordinates {
(0.0,0.0)(0.1,0.0)(0.2,0.0)(0.3,0.0)(0.4,0.0)(0.5,0.0)(0.6,0.0)(0.7,0.0)(0.8,0.0)(0.9,0.0)(1.0,0.0)
(1.05,0.0088)(1.1,0.031)(1.15,0.062)(1.2,0.0984)(1.25,0.1379)(1.3,0.1789)(1.35,0.2202)(1.4,0.2611)(1.45,0.3011)
(1.5,0.3397)(1.55,0.3767)(1.6,0.4121)(1.65,0.4458)(1.7,0.4777)(1.75,0.5079)(1.8,0.5365)(1.85,0.5633)(1.9,0.5887)
(1.95,0.6125)(2.0,0.6349)(2.1,0.6758)(2.2,0.7119)(2.3,0.7438)(2.4,0.7719)(2.5,0.7968)(2.6,0.8188)(2.7,0.8383)
(2.8,0.8556)(2.9,0.8709)(3.0,0.8845)(3.1,0.8966)(3.2,0.9074)(3.3,0.9169)(3.4,0.9255)(3.5,0.9331)};
\addlegendentry{$\zeta(c)^2$ (theory)}
\addplot[only marks, mark=*, mark size=1.4pt, red] coordinates {
(0.5,0.001)(0.75,0.001)(1.0,0.012)(1.25,0.134)(1.5,0.371)(2.0,0.605)(2.5,0.803)(3.0,0.888)};
\addlegendentry{Monte-Carlo}
\draw[dashed, gray] (axis cs:1,-0.03) -- (axis cs:1,0.92);
\node[font=\scriptsize, gray, anchor=south, rotate=90] at (axis cs:1,0.30) {threshold $c=1$};
\end{axis}
\end{tikzpicture}
\caption{\small The cheating epidemic. Fraction of an $m$-student class implicated after one student
is caught, $\E|R_0|/m$, against the mean number of friends per student $c$, for random
friendships $G(m,c/m)$. The order parameter is $\zeta(c)^2$, with $\zeta$ solving $1-\zeta=e^{-c\zeta}$ (solid). Monte-Carlo ($m=3000$, $250$ trials per point). 
}
\label{fig:epidemic}
\end{figure}

\begin{example}[The cheating epidemic]\label{ex:epidemic}
The same threshold governs cheating at scale, now in the reading of this section.
In Example~\ref{ex:main} the friendships were fixed background and the spreading
rules were switchable; here the roles swap: $\mathrm{BK}$ holds the spreading rule
$\mathrm{cheat}(Y)\leftarrow\mathrm{cheat}(X),\mathrm{friend}(X,Y)$ and the seed
$\mathrm{cheat}(B)$; $L$ is the set of all base facts $\mathrm{friend}(u,v)$ over
$m$ students; and the hypothesis $F\subseteq L$ is drawn at random, each pair
present with probability $c/m$ (mean degree $c$).
The friendship edges play the role of the random base facts $F$ of
Theorem~\ref{thm:phase}. Two things differ: the graph is undirected, $G(m,c/m)$
rather than $D(m,c/m)$, and the content is the connected component $R_0$ of the
seed rather than the transitive closure of all pairs. The argument is unchanged:
the breadth-first exploration of $R_0$ is the same $\mathrm{Poisson}(c)$
branching process, so the same threshold and the same $\zeta(c)^2$ apply
(Figure~\ref{fig:epidemic}). Cheating is a critical phenomenon with threshold one
friend per student: below it contained ($O(1)$ others), above it an epidemic
reaching a fraction $\zeta(c)^2$ of the class.
\end{example}

\paragraph{The same threshold governs semantic compression.} By Remark~1 the closure map is a
deterministic channel $F\mapsto R$, so $H(F)=H(R)+H(F\mid R)$. An optimal code for the random
program uses $H(F)$ bits; one for its meaning uses $H(R)\le H(F)$ bits, since it only identifies the
relation (by its strongly connected components and their order, Appendix~\ref{app:collapse}) rather
than listing the pairs of the denser graph $R$. The difference, the \emph{semantic
compression gain} $H(F\mid R)$, is the part of the program that its meaning makes redundant. As in
source coding, the \emph{semantic compression rate} is $H(R)/H(F)\le1$, the bits of meaning per bit
of program, and its complement $H(F\mid R)/H(F)=1-H(R)/H(F)$ is the \emph{semantic redundancy}, the
fraction of the program that its meaning does not record. (The terms are used here only in this
sense, not in the sense of semantic communication or NLP.) The next result shows that the extent
threshold is also a threshold for semantic compression, which
links the probabilistic model to the capacity and coding results (proof in
Appendix~\ref{app:collapse}).

\begin{theorem}[Semantic compression transition]\label{prop:collapse}
Let $F\sim\mathcal D(m,c/m)$ and $R=\TC(F)$. Then $H(F)=c\,m\log_2 m\,(1+o(1))$, and
(i) for $c<1$, $H(F\mid R)=O(\log m)$, so the compression rate is $H(R)/H(F)=1-O(1/m)$: semantics
compresses essentially nothing;
(ii) for $c>1$, $\limsup_{m}H(R)/H(F)\ \le\ 1-2\zeta(c)^2\bigl(1-\zeta(c)\bigr)<1$, i.e.\ the semantic
redundancy $H(F\mid R)/H(F)$ is at least $2\zeta(c)^2(1-\zeta(c))>0$: semantics compresses a constant
fraction of the program.
For the bandwidth-$d$ template with rules on independently with probability $\theta<1$,
$H(R)=O_{d,\theta}(1)$ while $H(S)=\Theta(n)$, so $H(R)/H(S)\to0$: the compression is total.
[The constant in (ii) is conservative: computing $P(R)$ exactly (Appendix~\ref{app:collapse}) shows
that the true compression rate is about $0.43$ at $c=1.5$ and $0.03$ at $c=3$, i.e.\ a semantic
redundancy of $0.57$ and $0.97$.]%
\end{theorem}

\vspace{-0.05in}
\section{Adversarial Model: Online Learning and minimax regret}\label{sec:learn}
\vspace{-0.05in}

Section~\ref{sec:phase} was about the semantic extent: how much a single content deduces.
We now return to the capacity. $\Om$ is, literally, the hypothesis class (often denoted
$\mathcal H$) a learner searches. The results below are mostly standard finite-class facts applied to
$\Om$; what Sections~\ref{sec:object}--\ref{sec:count} add is that $\Om$, its size and its
dimensions are now known, so the bounds become explicit rates. 
Each content $R\in\Om$ predicts, on every ground
atom, whether it is deducible: $h_R(\gamma)=\mathbf 1[\gamma\in R]$. So
$\Om\subseteq\{0,1\}^{\Bb}$ is a finite concept class, one concept per realizable theory.
The two errors contributing to the loss are logic's own: a false positive is a deduced falsehood (over-general rule), a false negative a missed truth. %

\vspace{-0.05in}
\paragraph{Mismatch loss.}
We score a learner by the $0/1$ (mismatch) loss against the best content in hindsight. As a warm-up we treat the realizable case.
\begin{theorem}[Realizable mistake bound]\label{thm:halving}
In the realizable online setting, the Halving algorithm
\citep{littlestone1988learning,cesabianchi2006prediction} over $\Om$ makes at most
$\log_2|\Om|=\CL$ mistakes. The exact mistake-bound value is the Littlestone dimension
$\Ldim(\Om)$, and $\VC(\Om)\le\Ldim(\Om)\le\log_2|\Om|=\CL$
\citep{littlestone1988learning} (see detailed proof in Appendix~\ref{app:ldim}).
\end{theorem}
\vspace{-.06in}
Now drop realizability: the label sequence $y_{1:T}\in\{0,1\}^T$ is arbitrary (adversarial),
and we measure $0/1$ regret against the best content in hindsight. The regret becomes
\[
\reg^{0/1}_T \;=\; \sum_{t=1}^T \mathbf 1[\hat y_t\ne y_t]
\;-\;\min_{R\in\Om}\sum_{t=1}^T \mathbf 1[h_R(\gamma_t)\ne y_t],
\]
where $\hat{y}_t \in \{0,1\}$ is the learner's predicted label for $\gamma_t$.
\begin{theorem}[Agnostic minimax rate]\label{thm:agnostic}
For every finite content class $\Om$ and every $T\ge\Ldim(\Om)$, the minimax expected $0/1$
regret over $T$ rounds satisfies
\[
c_1\sqrt{\Ldim(\Om)\,T}\;\le\;
\min_{\text{learner}}\max_{y_{1:T}}\E\,\reg^{0/1}_T\;\le\;
\sqrt{\tfrac{\ln 2}{2}\,\CL\,T},
\]
with $c_1=\tfrac14$
(see Algorithm~\ref{alg:mismatch} in Appendix~\ref{app:ewa}).
\end{theorem}
\vspace{-.08in}
The upper bound is Hedge over the $|\Om|$ content-experts (Algorithm~\ref{alg:mismatch} in Appendix~\ref{app:ewa});
the lower bound is the agnostic lower bound of \citet{bendavid2009agnostic}, played on a
Littlestone tree of $\Om$ (full proof in Appendix~\ref{app:agnostic}).
The two bounds ``match'' \emph{iff} $\Ldim(\Om)=\Theta(\CL)$: the Hedge upper bound holds for
any finite class, but without a lower bound on $\Ldim$ the matching regret lower bound could
be as small as $\sqrt{\Ldim\,T}$ with $\Ldim\ll\CL$. In general the optimal agnostic regret
is $\Theta(\sqrt{\Ldim\,T})$ \citep{alon2021adversarial}, so $\CL$ governs the agnostic rate only
when $\Ldim(\Om)=\Theta(\CL)$. For the bandwidth family this holds, as stated below and proved
in Appendix~\ref{app:ldim-d}; for transitive closure the atoms $\mathrm{tc}(a,b)$ across a balanced
bipartition are shattered, so $\Ldim(\Om)\ge\VC(\Om)\ge\lfloor m^2/4\rfloor$ (Appendix~\ref{app:nonlinear}). Whether this can fail for a recursive template is Open Problem~\ref{op:ldim} below.

\begin{corollary}[Bandwidth-$d$ agnostic rate]\label{cor:agnostic-bandwidth}
For the bandwidth-$d$ class, $\VC(\Om(n,d))=n-\lfloor n/d\rfloor+O(1)$, so the sandwich
$\VC\le\Ldim\le\log_2|\Om|$ of Theorem~\ref{thm:halving} has both ends linear in $n$ and
$\Ldim(\Om(n,d))=\Theta(\CL)$. Hence both sides of Theorem~\ref{thm:agnostic} are
$\Theta\!\big(\sqrt{\CL\,T}\big)$, and the minimax agnostic $0/1$ regret is
$\Theta\!\big(\sqrt{\CL\,T}\big)$.
\end{corollary}
More generally, for every \emph{linear} template (one body atom per clause, as in all our
experiments), the VC dimension, the fibre sizes and a linear-time consistent learner are all read
off reachability in the rule graph, and $\Om$ is closed under union (Proposition~\ref{prop:linear},
Appendix~\ref{app:linear}).

\begin{openproblem}\label{op:ldim}
Is there a family of recursive definite Horn templates with $\Ldim(\Om)=o(\CL)$, i.e.\ for which the
optimal agnostic $0/1$ regret is $o\big(\sqrt{\CL\,T}\big)$?
\end{openproblem}

\paragraph{Logarithmic Loss.} Under logarithmic loss each content is the
deterministic predictor $p^R_t(y)=\mathbf1[y=\mathbf1[\gamma_t\in R]]$, and a strategy
$\hat p=(\hat p_t)$ pays 
per round the log loss
$
\ell^{\log}_t(\hat p_t,y_t)\;:=\;-\log_2 \hat p_t(y_t),
$
so that its cumulative loss is the code length
$-\log_2\hat P(y_{1:T})$ with $\hat P(y_{1:T})=\prod_t\hat p_t(y_t)$. The minimax regret over
a query sequence is
$$
\reg^\star_T(\gamma_{1:T})=\min_{\hat p}\max_{y}[\sum_t\ell^{\log}_t(\hat p_t,y_t)-
\min_R\sum_t\ell^{\log}_t(p^R_t,y_t)].
$$

\begin{theorem}[Minimax log-loss regret is the Shtarkov sum --- and it is the capacity]\label{thm:shtarkov}
Write $\Om|_{\gamma_{1:T}}=\{(\mathbf1[\gamma_1\in R],\dots,\mathbf1[\gamma_T\in R]):R\in\Om\}$
for the label patterns the contents induce on the queries. For every query sequence the
minimax regret is the log Shtarkov sum,
\[
\reg^\star_T(\gamma_{1:T})=\log_2\sum_{y\in\{0,1\}^T}\max_{R\in\Om}P_R(y)
=\log_2\bigl|\Om|_{\gamma_{1:T}}\bigr|,
\]
attained uniquely by the normalized-maximum-likelihood (Shtarkov) strategy
\citep{shtarkov1987universal}. Consequently
$\reg^\star_T\le\log_2|\Om|=\CL$, with equality once the queries realize all $|\Om|$
patterns, and by Sauer--Shelah
$\reg^\star_T\le\min\{\CL,\ \VC(\Om)\log_2(T+1)\}$.
\end{theorem}

The proof (Appendix~\ref{app:th3}) is the normalized-maximum-likelihood argument; since each
content is a deterministic predictor, the Shtarkov sum counts the distinct label patterns.

\textbf{Evaluation of Algorithm~\ref{alg:vsp}.} For the bandwidth family, querying the prefix
$r_0,\dots,r_{T-1}$ gives $\reg^\star_T=(T-1)\log_2\rho_d+\log_2A_d+o(1)$: the redundancy grows by
$\log_2\rho_d$ bits per informative query until it saturates at $\CL$. Against this value we place
the horizon-free Bayes mixture under the uniform prior on $\Om$ (Algorithm~\ref{alg:vsp}), whose
hard-label version is Halving; Theorem~\ref{thm:vsp} evaluates it for an \emph{arbitrary} template
and query sequence.

\begin{theorem}[Exact regret of Algorithm~\ref{alg:vsp}]\label{thm:vsp}
Fix a query sequence $\gamma_{1:T}$ and a realizable outcome,
let $w:\Om\to\R_{>0}$ be the prior weight function parameterizing Algorithm~\ref{alg:vsp}
(default $w\equiv1$), extended to sets by $w(A)=\sum_{R\in A}w(R)$;
and let $V_{T+1}$ be the
final version space.
{\rm (i)} (Exact regret.) Under logarithmic loss the cumulative regret of
Algorithm~\ref{alg:vsp} is
\[
\reg_T=\log_2\frac{w(\Om)}{w(V_{T+1})}\ \le\ \log_2\frac{w(\Om)}{w(R^\star)},
\]
The right-hand side is the code length of
the true theory under $w$. Under the uniform prior $w\equiv1$ this reads
$\reg_T=\log_2\frac{|\Om|}{|V_{T+1}|}\le\log_2|\Om|=\CL$: it pays only for the contents it
has not yet ruled out.
{\rm (ii)} (Matching lower bound.) For the same query sequence, every sequential predictor has
worst-case regret at least the minimax value $\reg^\star_T=\log_2\bigl|\Om|_{\gamma_{1:T}}\bigr|$
of Theorem~\ref{thm:shtarkov}. Hence when the queries separate the contents
($\bigl|\Om|_{\gamma_{1:T}}\bigr|=|\Om|$, so $|V_{T+1}|=1$), the uniform
Algorithm~\ref{alg:vsp} attains $\log_2|\Om|=\CL$ and is minimax-optimal.
\end{theorem}

\begin{algorithm}[t]
\caption{Universal deductive predictor (version-space mixture)}\label{alg:vsp}
\KwIn{content class $\Om$; query stream $\gamma_1,\gamma_2,\dots$; prior weights
$w(R)>0$ (default $w\equiv1$)}
$V\leftarrow\Om$\tcp*{version space: contents consistent with labels}
\For{$t=1,2,\dots , T$}{
  receive query $\gamma_t$\;
  $\hat p_t\leftarrow \dfrac{\sum_{R\in V:\,\gamma_t\in R}w(R)}{\sum_{R\in V}w(R)}$
    \tcp*{mixture probability that $\gamma_t$ is entailed}
  predict $\hat p_t\in[0,1]$ (log loss), or $\hat y_t\leftarrow\mathbf1[\hat p_t\ge\tfrac12]$
    (0/1 loss)\;
  observe $y_t=\mathbf1[\gamma_t\in R^\ast]$\;
  $V\leftarrow\{R\in V:\mathbf1[\gamma_t\in R]=y_t\}$\tcp*{prune inconsistent contents}
}
\end{algorithm}

\vspace{-0.05in}
\section{Experiments}\label{sec:exp}
\vspace{-0.05in}
We first illustrate the two exact statements of the framework on a real relational dataset,
Padgett's Florentine families marriage network ($15$ families, $20$ ties): syntax collapses
onto a small content class, and the online cost of Algorithm~\ref{alg:vsp} is exactly
$\CL=\log_2|\Om|$. Both are identities (Proposition~\ref{prop:realize} and
Theorem~\ref{thm:vsp}), so these runs check the constructions and show that they scale; the
tests that can come out either way follow below. We use the template with
the seed $E=\{\textsf{marked}(\textsf{Medici})\}$ and one clause
$\textsf{marked}(Y)\leftarrow\textsf{marked}(X),\textsf{tie}(X,Y)$ per orientation of each
tie, where $\textsf{marked}(X)$ reads ``family $X$ has been reached from the seed''. Thus with
$N=40$ ground clauses, the content sub-base is the $15$ \textsf{marked}-atoms, one per family
(the ties are fixed background). Appendix~\ref{app:exps} gives details and scales the same
experiment to the {\sc WordNet} taxonomy.

\begin{figure}[h]
\centering
\includegraphics[width=0.8\linewidth]{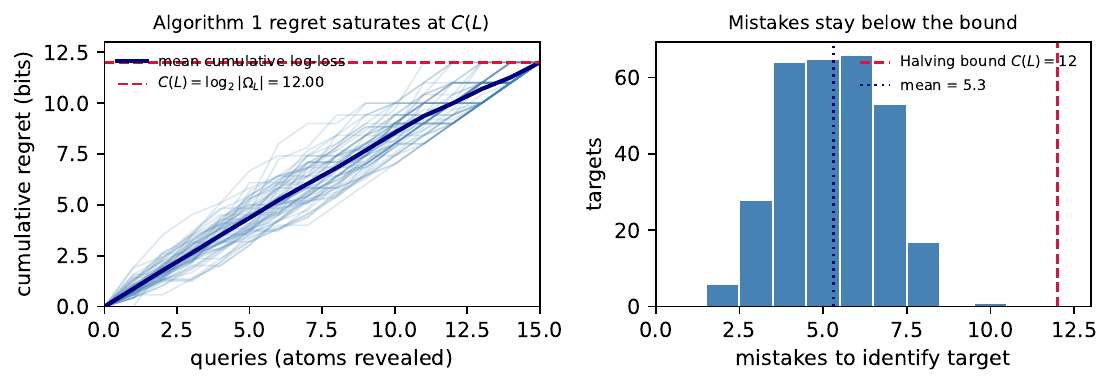}
\caption{\small Algorithm~\ref{alg:vsp} on the Florentine network ($|\Om|=4{,}104$). \textbf{Left:}
cumulative log-loss of $60$ random runs (light) and their mean (dark); every run ends exactly at
$\CL=12.00$ bits, as Theorem~\ref{thm:vsp} predicts. \textbf{Right:} mistakes to identify the
target over $300$ runs (mean $5.3$, at most $10$), all below the Halving bound $\CL\approx12$.}
\label{fig:florentine}
\end{figure}

\textbf{Semantic collapse and exact regret.} Enumerating the least Herbrand model of all
$2^{40}\approx1.1\times10^{12}$ clause-sets, the syntax collapses onto exactly $|\Om|=4{,}104$
contents, the connected atom-sets containing the seed, as Proposition~\ref{prop:realize} predicts:
the template expresses $\CL=\log_2 4{,}104=12.00$ bits, not the $40$ bits a clause-set learner
would budget for. Running Algorithm~\ref{alg:vsp} (uniform prior) with the $15$ atoms revealed in
random order, the cumulative log-loss reaches $\CL$ \emph{exactly} for every target and order
(Figure~\ref{fig:florentine}, left; Theorem~\ref{thm:vsp}); under $0/1$ loss the target is identified
with mean $5.3$ and at most $10$ mistakes over $300$ targets, below the bound $\CL\approx12$
(Figure~\ref{fig:florentine}, right). The regret grows by less than one bit per revealed atom
until the queries separate all contents. The same holds at scale, where the point is that the
mixture can be computed at all: on the
{\sc WordNet} \textsf{carnivore} taxonomy ($366$ concepts, $2^{365}$ clause-sets,
$|\Om|=10^{90.9}$) Algorithm~\ref{alg:vsp}, run implicitly by an $O(n)$ tree dynamic program, pays
exactly $\CL=302$ bits (Table~\ref{tab:wordnet}).

\textbf{Baseline: a clause-prior mixture.} This baseline models learners that place a prior or a
description length on \emph{programs} (Bayesian and MDL program induction, clause-level priors in
probabilistic ILP, soft clause weights in differentiable ILP); it is not a model of search-based
ILP systems. It \emph{is} Algorithm~\ref{alg:vsp} with prior $w(R)=|\mu^{-1}(R)|$, so
Theorem~\ref{thm:vsp}(i) prices it exactly. On Florentine it pays $6$ to $20$ bits ($14.35$ on average, above $\CL$ on $79\%$ of the
contents) against $12.00$ on every target; on the bandwidth-$2$ template it pays $1.01\,n$ bits on
average and $\tfrac43n$ in the worst case against $\CL=0.694\,n$. It wins only on targets with huge
fibres, such as the contents of random clause-sets (Appendix~\ref{app:baseline}).

We now test a setting where the code-length identity does not apply, and check the semantic
compression transition. All runs are exact over $\Om$, implicitly on WordNet
(Appendix~\ref{app:wordnet}); complete results, including held-out atoms, are in
Appendix~\ref{app:newexp}.

\begin{figure}[t]
\begin{minipage}[b]{0.51\linewidth}\centering\footnotesize
\setlength{\tabcolsep}{3.5pt}
\begin{tabular}{@{}lcccc@{}}
\toprule
& \multicolumn{2}{c}{Florentine} & \multicolumn{2}{c}{\textsc{carnivore}}\\
\cmidrule(lr){2-3}\cmidrule(l){4-5}
target from & contents & rules & contents & rules\\
\midrule
Alg.~1 & 110/47 & -- & 1056/45 & --\\
Alg.~2, cont. & 21.6/0.4 & 21.6/0.4 & 363/4.4 & 356/1.1\\
Alg.~2, clause & 25.0/0.8 & 17.8/0.4 & 401/4.6 & 81/0.8\\
\bottomrule\end{tabular}
\vspace{6pt}
\captionof{table}{\small Label noise, $\epsilon=0.1$: $0/1$ regret (guessing $\approx120$ on
Florentine, $\approx1200$ on \textsf{carnivore}) / clean error (\%), by how the target was
generated (contents or random rules).}
\label{tab:noise}
\end{minipage}\hfill
\begin{minipage}[b]{0.45\linewidth}\centering
\includegraphics[width=\linewidth]{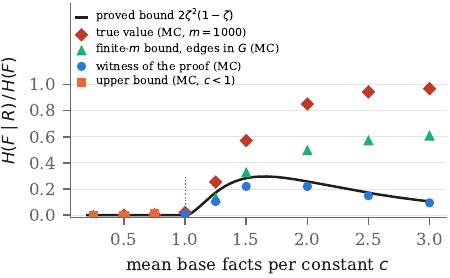}
\captionof{figure}{\small Semantic redundancy $H(F\mid R)/H(F)$: true value (exact $P(R)$, $m=1000$),
lower bounds of the proof, proved bound.}
\label{fig:new}
\end{minipage}
\end{figure}

\textbf{Label noise.} Each label is flipped independently with probability $\epsilon$.
In each of $T$ rounds a random atom is queried about a hidden target content $R^\star$, drawn
either uniformly from $\Om$ (``contents'') or as the content of a uniformly random clause-set
(``rules''), and the learner predicts before seeing the possibly flipped label. The
\emph{regret} is the learner's number of mistakes minus that of the best content in hindsight,
which itself errs on about $\epsilon T$ flipped labels (random guessing: $(\tfrac12-\epsilon)T$). The
\emph{clean error} is the error rate against the true labels over the last quarter of the run,
i.e.\ whether the learner has found the target. Pruning (Algorithm~\ref{alg:vsp}) loses the
target: for every $\epsilon\in\{0.05,0.1,0.2\}$ its version space empties in all runs, and its clean
error is $43$--$47\%$. Hedge over $\Om$ (Algorithm~\ref{alg:mismatch}) stays below its
bound $\sqrt{(\ln2/2)\,\CL\,T}$ ($35$ and $560$ mistakes), implicitly on \textsf{carnivore}
(Appendix~\ref{app:ewa}). With the content-uniform prior the regret does
not depend on how the target arose (Table~\ref{tab:noise}): $21.6$ vs.\ $21.6$ mistakes on Florentine
and $363$ vs.\ $356$ on \textsf{carnivore}. The clause-uniform prior is a bet: it wins on targets
generated by random rules ($17.8$ and $81$) and loses otherwise ($25.0$ and $401$).
Uniform-on-contents is the equalizer (Theorem~\ref{thm:vsp}).

\textbf{Semantic compression.} Computing $P(R)$ exactly from the strongly connected components
(Appendix~\ref{app:collapse}) gives the true redundancy $H(F\mid R)/H(F)$ (Figure~\ref{fig:new}):
small and decreasing in $m$ below $c=1$, but $0.57$, $0.85$, $0.97$ at $c=1.5$, $2$, $3$: semantics
compresses most of a random program.

\textbf{A differentiable-ILP-style learner.} We trained a learner with one soft weight per
clause \citep[in the spirit of][]{evans2018learning} on the same noisy Florentine data as
Algorithm~\ref{alg:mismatch}. Its predictions are as accurate as ours, but the program it
returns is not: rounding its weights to a clause-set gives $5.4\%$ and $13.8\%$ wrong atoms at
$\epsilon=0.1$ and $0.2$ ($T=150$), against $0.0\%$ and $1.6\%$ for our best content, because
its evidence is split across many clause-sets with the same meaning (Appendix~\ref{app:dilp}).

\section{Concluding remarks}\label{sec:position}

For recursive Horn templates the right object is the semantic content class $\Om$. Its
semantic information $\CL=\log_2|\Om|$ is a single invariant, and counting, coding, and
learning all measure it once $\Om$ is constructed. Syntax over-counts by the collapse ratio
$2^N/|\Om|$, so guarantees that count hypotheses through the parameterization (Occam and MDL
bounds, priors, or code lengths over clause-sets) are inflated by up to that factor; guarantees
based on VC or Littlestone dimension are not, since they already measure the induced class. The
experiments show where the choice matters in practice: under label noise, exponential weights
over $\Om$ keeps a regret that does not depend on how the target arose, while a clause-level
prior wins or loses by a wide margin depending on the source. The template poses two very different
questions. Capacity asks how many contents exist; semantic extent asks how much one content deduces.
The two are governed by different mathematics, and conflating them is the error the framework
is built to prevent. The running example makes this concrete. Catch one cheater, and the
possible outbreaks form a $10$-element lattice: $\CL=\log_2 10$ bits, not $2^{10}$
clause-sets. The probabilistic rule reproduces the capacity/entropy/gap triad, and at scale a
sharp epidemic threshold at one friend per student. And Halving learns who cheats in $\CL$
mistakes. This is the concrete, combinatorial face of incorporating logical reasoning into
information theory \citep{lastras2026}. Open directions include the precise second-order
asymptotics of the nonlinear count, the exact constant in $\Ldim(\Om)=\Theta(\CL)$, Open Problem~\ref{op:ldim} (whether some
recursive Horn template has $\Ldim(\Om)=o(\CL)$), a noisy semantic capacity, the complexity of
computing $|\Om|$ for general DAGs, and a head-to-head comparison with search-based ILP
systems.

\section*{AI use statement}

Generative AI tools were used in this work for the following purposes: writing and debugging experiment code, generating figures, drafting and editing parts of the text, and checking some of the proofs. The problem formulation, the main results, and the overall structure of the paper are the authors' own. Every proof was independently verified by the authors, and all AI-assisted text, code, and figures were reviewed and edited by us. The authors take full responsibility for the content of this work, including any text, claims, or artifacts produced with the aid of generative AI.

\section*{Reproducibility statement}
All proofs are in Appendix~\ref{app:proofs}. Code and data for every experiment are provided as
supplementary material; WordNet is accessed through NLTK, and Padgett's network through
NetworkX.

\bibliographystyle{plainnat}
\bibliography{ref-logic4}

\newpage
\appendix

\section{Full proofs}\label{app:proofs}

This appendix gives the complete arguments for the results whose proofs were sketched in
the body. Throughout, $\mu(S)=\lhm(\BK\cup S)$, $\Om=\im\mu$, and for the bandwidth family
$a_n=|\Om(n,d)|$.

\subsection{Realizability (Proposition~\ref{prop:realize})}

\begin{proof}
Let the template consist of edge clauses $r_v\leftarrow r_u$ over $G=(V,E)$ with source set
$V_0$ declared by $\BK$.

\emph{($\Rightarrow$).} Suppose $R=\mu(S)=\lhm(\BK\cup S)$ for some $S\subseteq L$, and let
$v\in R\setminus V_0$. Since $v$ is not a background source, it entered the least model at
some finite stage: $v\in\TP{\uparrow}(k+1)\setminus\TP{\uparrow}k$ for some $k\ge0$, where
$P=\BK\cup S$. By definition of $\TP$, $v$ was produced by firing a clause
$r_v\leftarrow r_u\in S$ whose body $r_u$ already held, i.e.\ $u\in\TP{\uparrow}k\subseteq R$
and $u\to v\in E$. Thus $v$ has an in-neighbour $u\in R$ that entered strictly earlier;
iterating on $u$ terminates at $V_0$ and exhibits a path from $V_0$ to $v$ lying entirely in
$R$, the stated reachability condition. (When $G$ is acyclic, ``every $v$ has an in-neighbour
in $R$'' already forces this, since backward walks cannot cycle.)

\emph{($\Leftarrow$).} Suppose $V_0\subseteq R$ and every $v\in R$ is reachable from $V_0$
inside $R$. Take $S:=\{(u\to v)\in E:u,v\in R\}$, the edges internal to $R$. First,
$\lhm(\BK\cup S)\subseteq R$: every clause of $S$ has its head in $R$ by construction, and
$V_0\subseteq R$, so no atom outside $R$ is ever produced. Second, $R\subseteq\lhm(\BK\cup
S)$: define the within-$R$ distance $\delta(v)$ of $v\in R$ from $V_0$ along paths lying in
$R$ (finite by hypothesis). Induct on $\delta$. If $\delta(v)=0$ then $v\in V_0$, so $v$
holds. If $\delta(v)=\ell+1$, pick an in-$R$-neighbour $u$ with $\delta(u)=\ell$; by
induction $u\in\lhm(\BK\cup S)$, and $(u\to v)\in S$, so firing it adds $v$. Hence
$R=\lhm(\BK\cup S)$ and $R\in\Om$.
\end{proof}

\subsection{Exact count and asymptotics (Theorem~\ref{thm:count})}
\label{app:th1}

\begin{proof}
\emph{Generating function.} The compositions with parts in $\{1,\dots,d\}$ have the
ordinary generating function
\[
\sum_{x\ge0}C_d(x)\,t^x=\frac{1}{1-(t+t^2+\cdots+t^d)}
=\frac{1-t}{1-2t+t^{d+1}},
\]
the last step by $t+\cdots+t^d=t(1-t^d)/(1-t)$ and simplification. Hence
$a_n=\sum_{x=0}^n C_d(x)$ is the partial-sum sequence, and the partial-sum operator
multiplies the generating function by $(1-t)^{-1}$:
\[
\sum_{n\ge0}a_n t^n=\frac{1}{1-t}\cdot\frac{1-t}{1-2t+t^{d+1}}=\frac{1}{1-2t+t^{d+1}}.
\]
Clearing denominators, $\sum_n a_n t^n=(1-2t+t^{d+1})^{-1}$, and reading off the coefficient of
$t^n$ for $n\ge d+1$ gives $a_n-2a_{n-1}+a_{n-d-1}=0$, i.e.\ the stated recurrence; for
$k\le d$ every subset of $\{1,\dots,k\}$ has all gaps $\le d$, so $a_k=2^k$.

\emph{Dominant root.} Write $Q(t)=1-2t+t^{d+1}$. It factors as
\[
Q(t)=(1-t)\,Q_d(t),\qquad Q_d(t)=1-t-t^2-\cdots-t^d .
\]
Substituting $\rho=1/t$, the roots of $Q_d$ are the reciprocals of the roots of the
\emph{$d$-bonacci polynomial} $\rho^d-\rho^{d-1}-\cdots-\rho-1$. This polynomial has distinct roots,
exactly one of which, the $d$-bonacci constant $\rho_d\in(1,2)$, lies outside the unit disk; the
other $d-1$ lie strictly inside it \citep{miles1960generalized}. Hence $t_d:=1/\rho_d\in(\tfrac12,1)$
(indeed $Q_d(\tfrac12)=2^{-d}>0>1-d=Q_d(1)$), all other zeros of $Q_d$ have modulus greater than $1$,
and the factor $1-t$ adds a simple pole at $t=1$. So $t_d$ is the unique singularity of
$\sum a_nt^n=1/Q(t)$ of smallest modulus, and it is simple.

\emph{Singularity analysis.} A rational function is the sum of the principal parts at its poles,
so
\[
a_n\;=\;A_d\,\rho_d^{\,n}\;-\;\frac1{d-1}\;+\;O(\theta_d^{\,n}),
\qquad A_d=\frac{-1}{Q'(t_d)\,t_d}=\frac{1}{\bigl(2-(d+1)t_d^{\,d}\bigr)\,t_d},
\]
where $Q'(t)=-2+(d+1)t^d$, the constant $-1/(d-1)=1/Q_d(1)$ is the contribution of the pole at
$t=1$, and $\theta_d<1$ is the largest modulus of the other $d-1$ roots of the $d$-bonacci polynomial
\citep[Ch.~IV]{flajolet2009analytic}. In particular $a_n=A_d\rho_d^{\,n}(1+O(\gamma_d^{\,n}))$ with
$\gamma_d=1/\rho_d$. For $d=2$, $A_2=\varphi^3/\sqrt5$ and the expansion is the exact identity
$a_n=F_{n+3}-1$.
\end{proof}

\subsection{Nonlinear capacity is quadratic with constant \texorpdfstring{$\kappa=\tfrac14$}{kappa=1/4} (Theorem~\ref{thm:nonlinear})}
\label{app:nonlinear}

\begin{proof}
Fix $m$ constants and the composition rule
$\mathrm{tc}(X,Z)\leftarrow\mathrm{tc}(X,Y),\mathrm{tc}(Y,Z)$ as background; the switchable
clauses are the base facts. A hypothesis $F$ selects a set of base facts, and its content is
$\mu(F)=\TC(F)$, the transitive closure. Since every transitive relation is its own closure,
$\Om$ is exactly the set of realizable transitive relations. We first treat the \emph{ordered}
template (base facts $(i,j)$ with $i<j$, the acyclic/DAG language bias), for which a content
is a transitively closed subrelation of the ground order $\{(i,j):i<j\}$; the unordered
variant is handled at the end. Write $T(m)=|\Om|$.

\emph{Lower bound $T(m)\ge 2^{\lfloor m^2/4\rfloor}$.} Split $[m]=A\sqcup B$ with
$A=\{1,\dots,\lfloor m/2\rfloor\}$ and $B=\{\lfloor m/2\rfloor+1,\dots,m\}$, and let $R$ be an
arbitrary subset of $A\times B$. Every pair $(a,b)\in R$ has $a\in A$, $b\in B$, hence $a<b$,
so $R$ respects the ground order. $R$ is transitive: a violation would need
$(x,y),(y,z)\in R$ with $(x,z)\notin R$, but then $y$ is at once a head (so $y\in B$) and a
tail (so $y\in A$), which is impossible. Hence every such $R$ is a realizable content, and
distinct subsets give distinct relations, so
$T(m)\ge 2^{|A|\,|B|}=2^{\lfloor m/2\rfloor\lceil m/2\rceil}=2^{\lfloor m^2/4\rfloor}$.

\emph{Upper bound $T(m)\le 2^{\,m^2/4+o(m^2)}$.} Each content $R$ is a \emph{strict partial
order} on $[m]$: it is transitive by membership in $\Om$; it is irreflexive and antisymmetric
because $R\subseteq\{(i,j):i<j\}$ forbids every $(i,i)$ and forbids $(i,j)$ and $(j,i)$
together. Distinct contents are therefore distinct labeled strict partial orders, so
$T(m)\le P(m)$, the number of partial orders on $m$ labeled points. By the
Kleitman--Rothschild enumeration \citep{kleitman1975asymptotic},
$\log_2 P(m)=\tfrac{m^2}{4}+o(m^2)$: almost every labeled poset is a three-layer order with
layer sizes $\tfrac m4,\tfrac m2,\tfrac m4$ and about $\tfrac{3}{16}m^2$ comparable pairs, and
these dominate the count. Hence $\log_2 T(m)\le\tfrac14 m^2+o(m^2)$.

\emph{Conclusion (ordered case).} Combining,
$\lfloor m^2/4\rfloor\le\log_2 T(m)\le\tfrac14 m^2+o(m^2)$; dividing by $m^2$ and letting
$m\to\infty$ gives $\kappa=\lim_m\log_2 T(m)/m^2=\tfrac14$, and
$\CL=\log_2 T(m)=\tfrac14 m^2(1+o(1))$.

\emph{Unordered variant.} Now the base facts are all ordered pairs $(i,j)$ with $i\ne j$ (both
directions switchable), and a content is $R=\TC(F)$ for a loop-free digraph $F$ on $[m]$. Such an
$R$ is transitive, and it contains a loop $(i,i)$ exactly when $i$ lies on a directed cycle of
$F$, i.e.\ when the strongly connected class of $i$ has at least two elements. We claim that
$R\mapsto\bar R:=R\cup\Delta$, with $\Delta=\{(i,i):i\in[m]\}$, is a bijection from $\Om$ onto
the preorders on $[m]$. The image is reflexive and transitive, hence a preorder. The map is
injective, because $R$ is recovered from $\bar R$ by deleting the loops $(i,i)$ for which $i$ is
alone in its $\bar R$-equivalence class. It is surjective: given a preorder $Q$, let $F$ be its
off-diagonal part; then $\TC(F)$ agrees with $Q$ off the diagonal (by transitivity of $Q$) and
contains $(i,i)$ exactly when $i$ has a $Q$-equivalent $j\ne i$, so $\TC(F)\cup\Delta=Q$. Hence
$|\Om|=Q(m)$, the number of preorders on $[m]$ (for $m=2$ there are $4$ contents, not the $13$
transitive relations on $[2]$). Preorders correspond to finite topologies, and
$\log_2 Q(m)=\tfrac{m^2}{4}+o(m^2)$ \citep{kleitman1970topologies}, so $\kappa=\tfrac14$ holds for
the unordered variant as well; the bipartite family above gives the matching lower bound.
\end{proof}

\subsection{Phase transition of semantic content (Theorem~\ref{thm:phase})}
\label{app:th2}

\begin{proof}
Write $p=c/m$ and fix the source $0$. By vertex-symmetry of $\mathcal D(m,p)$,
$\E|R|=\sum_i\E|\{j:i\rightsquigarrow j\}|=m\,\E|R_0|$, the diagonal $i=j$ contributing only
$O(m)=o(m^2)$; so it suffices to analyse the forward reachable set
$R_0=\{v:0\rightsquigarrow v\}$.

\emph{Step 1 (exploration process).} Discover $R_0$ by breadth-first search: colour $0$
active and all others neutral; repeatedly take an active vertex $u$, reveal its arcs to the
still-neutral vertices (each present independently with probability $p$), recolour every
newly hit vertex active, and retire $u$. Let $D_t$ be the number discovered after $t$
retirements ($D_0=1$). The number of children of the $t$-th retired vertex is
$\xi_t\sim\mathrm{Binomial}(m-D_t,\,p)$, independent across $t$ given the history.

\emph{Step 2 (branching coupling).} Since $m-D_t\le m-1$, each $\xi_t$ is stochastically
dominated by $\mathrm{Binomial}(m-1,p)$, and while $D_t=o(m)$ it also dominates
$\mathrm{Binomial}(m-1-o(m),p)$; both bounds converge in law to $\Pois(c)$ as
$m\to\infty$. A standard coupling
\citep{vanderhofstad2017random,athreya1972branching} matches the exploration, up to its
first $\varepsilon m$ discoveries, with a Galton--Watson process $T_c$ of offspring law
$\Pois(c)$.

\emph{Step 3 (subcritical $c<1$).} For $T_c$ with mean offspring $c<1$, generation $k$ has
expected size $c^k$, so the expected total progeny is $\sum_{k\ge0}c^k=1/(1-c)<\infty$;
extinction is a.s.\ with exponentially decaying total-progeny tails. The exploration is
dominated above by a $\mathrm{Binomial}(m-1,p)$ branching process, itself subcritical, so
$\{|R_0|\}_m$ is uniformly integrable; combined with $|R_0|\xrightarrow{d}|T_c|$ this gives
$\E|R_0|\to\frac{1}{1-c}$, hence $\E|R|=m\,\E|R_0|=\Theta(m)$.

\emph{Step 4 (supercritical $c>1$).} The extinction probability $q$ of $T_c$ is the smallest
root of $q=e^{-c(1-q)}$; writing $\zeta:=1-q$ recasts this as $1-\zeta=e^{-c\zeta}$, whose
unique positive root exists exactly for $c>1$. The exploration splits along the two fates of
$T_c$: with probability $\to q=1-\zeta$ it dies after $O_P(1)$ discoveries; with probability
$\to\zeta$ it survives and, by Karp's analysis of $\mathcal D(m,c/m)$
\citep{karp1990transitive}, fills the giant out-component, whose relative size equals the
survival probability, so $|R_0|=\zeta m(1+o(1))$ w.h.p. Taking expectations,
$\E|R_0|=\zeta\cdot\zeta m(1+o(1))+(1-\zeta)\cdot O(1)=\zeta^2 m(1+o(1))$, hence
$\E|R|=\zeta^2 m^2(1+o(1))$.

\emph{Step 5 (the critical point $c=1$).} Because the exploration of Step~1 only ever reveals
arcs into still-neutral vertices, $|R_0|$ has exactly the law of the component of a vertex in
the undirected graph $G(m,p)$, so $\E|R_0|$ is the Erd\H{o}s--R\'enyi susceptibility. At
$p=1/m$ the limiting Galton--Watson process is critical ($\zeta(1)=0$), with total-progeny
tail $\Pr[|T_1|\ge k]=\Theta(k^{-1/2})$. The exploration inherits this tail only up to
$k\asymp m^{2/3}$: after $D_t$ discoveries the mean offspring is $1-D_t/m$, and this drift
becomes visible exactly at $D_t\asymp m^{2/3}$, so $\Pr[|R_0|\ge k]=\Theta(k^{-1/2})$ for
$k\le m^{2/3}$ \citep[Ch.~5]{vanderhofstad2017random}, while beyond that scale
$\Pr[|R_0|>A\,m^{2/3}]\le 4m^{-1/3}e^{-A^2(A-4)/32}$ for $A>8$
\citep[Thm.~1]{nachmias2010critical}. Summing the tail,
\[
\E|R_0|=\sum_{k\ge1}\Pr[|R_0|\ge k]
=\sum_{k\le m^{2/3}}\Theta(k^{-1/2})\;+\;O\bigl(m^{2/3}\cdot m^{-1/3}\bigr)
=\Theta(m^{1/3}),
\]
the well-known $\Theta(m^{1/3})$ critical susceptibility. Hence
$\E|R_0|=\Theta(m^{1/3})=o(m)$ and $\E|R|=m\,\E|R_0|=\Theta(m^{4/3})=o(m^2)$: the order of
$\E|R|$ passes from $\Theta(m)$ through $\Theta(m^{4/3})$ to $\Theta(m^2)$ as $c$ crosses $1$,
and the transition is sharp.
\end{proof}

\subsection{Proof of Corollary~\ref{cor:subcritical}}
\label{app:cor2}

\begin{proof}
Fix $d$ and $\theta<1$. On the bandwidth-$d$ template the atoms are $r_0,\dots,r_n$
with source $r_0$, and the only clauses into $r_j$ are the $\le d$ edges $r_j\leftarrow r_i$
with $1\le j-i\le d$, each present independently with probability $\theta$. Whether $r_j$ is
reachable from the source therefore depends on the history only through the reachability
indicators of the preceding $d$ atoms and the (independent) presence of those $\le d$ edges.

\emph{A finite-window Markov chain.} Encode the state at position $j$ as the reachability
pattern of the length-$d$ window ending at $j$,
\[
W_j \;=\; \bigl(\mathbf 1[r_{j-d+1}\text{ reachable}],\dots,\mathbf 1[r_j\text{ reachable}]\bigr)
\in\{0,1\}^d .
\]
Sliding the window from $j$ to $j+1$ drops the oldest coordinate and appends
$\mathbf 1[r_{j+1}\text{ reachable}]$, and $r_{j+1}$ is reachable iff at least one of the
reachable atoms among $r_{j+1-d},\dots,r_j$ has its edge to $r_{j+1}$ switched on. Given
$W_j$, this new bit is an independent Bernoulli whose parameter depends only on $W_j$ and
$\theta$. Hence $(W_j)_{j\ge0}$ is a time-homogeneous Markov chain on the $2^d$ window-states,
with transition matrix $M=M(d,\theta)$.

\emph{The all-zero window absorbs.} Let $\mathbf 0=(0,\dots,0)$ be the all-unreachable
window. If $W_j=\mathbf 0$ then none of the atoms feeding $r_{j+1}$ is reachable, so
$r_{j+1}$ is unreachable and $W_{j+1}=\mathbf 0$: the state $\mathbf 0$ is absorbing. Moreover,
from \emph{any} window $W_j$ the chain is absorbed within $d$ steps with probability at least
\[
\varepsilon:=(1-\theta)^{d(d+1)/2}.
\]
Indeed, the next $d$ atoms $r_{j+1},\dots,r_{j+d}$ are all unreachable if every edge from the
current window into them is off: atom $r_{j+k}$ ($1\le k\le d$) has at most $d-k+1$ in-edges
from $r_{j+k-d},\dots,r_j$, and once $r_{j+1},\dots,r_{j+k-1}$ are unreachable these are the only
edges that matter. There are at most $\sum_{k=1}^d(d-k+1)=d(d+1)/2$ such edges, each off with
probability $1-\theta$. Thus every non-absorbing state is transient, and by the
Markov property $\Pr[W_j\ne\mathbf 0]\le(1-\varepsilon)^{\lfloor j/d\rfloor}$.

\emph{Exponential tail and the bound.} A reachable atom at position $j$ requires
$W_j\ne\mathbf 0$, so
\begin{align*}
\E\,|R_0| \;&=\; 1+\sum_{j\ge1}\Pr[r_j\text{ reachable}]
\;\le\; 1+\sum_{j\ge1}(1-\varepsilon)^{\lfloor j/d\rfloor}
\;\le\; 1+(d-1)+d\,\frac{1-\varepsilon}{\varepsilon}\\
&\le\;1+\frac{d}{\varepsilon}
\;=\;1+d\,(1-\theta)^{-d(d+1)/2},
\end{align*}
uniformly in $n$. In particular $|R_0|$ is a.s.\ finite and has exponentially decaying tails.
Since $\varepsilon>0$ for \emph{every} $\theta<1$, there is no $\theta<1$ at which the
mean diverges: the bandwidth-$d$ process is permanently subcritical, with no critical point.
The order $(1-\theta)^{-d(d+1)/2}$ is attained: solving the chain exactly gives
$\E|R_0|=9.901\times10^{5}$ for $d=2$ and $9.899\times10^{11}$ for $d=3$ at $\theta=0.99$,
against the bounds $2\times10^6$ and $3\times10^{12}$.
\end{proof}

\subsection{The semantic compression transition: proof of Theorem~\ref{prop:collapse}}\label{app:collapse}

\emph{Overview.} The program $F$ is a random directed graph and its meaning $R=\TC(F)$ is the
reachability relation. $H(F\mid R)$ measures how much of the graph is left undetermined once one
knows who reaches whom. Both parts of the proof rest on one observation: $H(F\mid R)$ is small if
$F$ can be rebuilt from $R$ with little extra information, and large if many edges can be changed
without changing $R$.

\emph{Part (i), $c<1$: an upper bound by reconstruction.} A pair $(i,j)\in R$ with no way back
from $j$ to $i$ and no detour $i\to k\to j$ must be an edge of $F$, so these \emph{essential} pairs
can be read off $R$. The remaining edges form a set $K$, and $F$ is determined by $R$ and $K$, so
$H(F\mid R)\le H(K)\approx\E|K|\cdot2\log_2m$. An edge of $K$ lies on a cycle, has a parallel path, or
touches a cycle; below $c=1$ the graph is locally tree-like and the expected number of such
configurations is a convergent geometric series, so $\E|K|=O(1)$ and $H(F\mid R)=O(\log m)$.

\emph{Part (ii), $c>1$: a lower bound by counting interchangeable edges.} A giant strongly
connected component $G$ with about $\zeta^2m$ vertices appears. An edge entering $G$ from a vertex
outside it can be redirected to any vertex of $G$, and an edge leaving $G$ can be made to start at
any vertex of $G$, without changing $R$ (Step~1). Grouping the graphs that differ by such moves
gives equally likely classes with the same $R$, so $H(F\mid R)$ is at least the expected log-size of
a class (Step~2). A vertex outside $G$ sends about $c\zeta^2$ edges into $G$, and a vertex is outside
$G$ with probability about $1-\zeta$, the chance that the vertices reaching it form a finite
branching process. So about $c\zeta^2(1-\zeta)m$ edges enter $G$ from outside, and by reversing all
edges as many leave it (Step~3). Each is worth about $\log_2|G|\approx\log_2m$ bits, which gives the constant
$2\zeta^2(1-\zeta)$ after dividing by $H(F)\approx c\,m\log_2m$ (Step~4). Only two standard facts are
used: the size of the giant component and the branching-process limit of the exploration around
one vertex.

\emph{Bandwidth.} With bounded reach, the meaning is $O(1)$ bits while the program has $\Theta(n)$,
so the redundancy tends to $1$.

\emph{The true value.} After the proof, the exact product formula for $P(R)$ makes
$H(F\mid R)$, and so the compression rate $H(R)/H(F)$, computable by simulation; it shows that the
bound in (ii) is conservative, the missing
part being the edges inside $G$.

\emph{Details.} Throughout, $F\subseteq\{(i,j)\in[m]^2:i\ne j\}$ contains each pair independently with
probability $p=c/m$, and $R=\TC(F)$; $R$ contains a loop $(i,i)$ exactly when $i$ lies on a
directed cycle of $F$. Entropies are in bits and $U:=m(m-1)$. Since
$H(F)=U\,h_2(p)$ with $h_2(p)=p\log_2(1/p)+O(p)$, we have $H(F)=c\,m\log_2 m\,(1+o(1))$.

\emph{(i) $c<1$.} Let $E_R$ be the set of pairs $(i,j)\in R$ with $i\ne j$, $(j,i)\notin R$, and no
$k\notin\{i,j\}$ such that $(i,k),(k,j)\in R$. First, $E_R\subseteq F$: if $(i,j)\in R\setminus F$,
there is a simple path $i=v_0\to v_1\to\cdots\to v_\ell=j$ in $F$ with $\ell\ge2$, and
$k=v_1\notin\{i,j\}$ has $(i,k),(k,j)\in R$. Hence $F=E_R\cup K$ with $K:=F\setminus E_R$, so $F$
is a function of $(R,K)$ and
\[
H(F\mid R)\ \le\ H(K)\ \le\ H(|K|)+\E|K|\,\log_2 U\ \le\ \log_2\!\bigl(e(1+\E|K|)\bigr)+2\,\E|K|\log_2 m ,
\]
using that a nonnegative integer variable with mean $\mu$ has entropy at most $\log_2(e(1+\mu))$.
It remains to show $\E|K|=O(1)$. Let $(i,j)\in K$. Either $(j,i)\in R$, so the edge lies on a
directed cycle; or there is $k\notin\{i,j\}$ with paths $P_1:i\rightsquigarrow k$ and
$P_2:k\rightsquigarrow j$ in $F$. If neither path uses the edge $(i,j)$, their concatenation
contains a second path from $i$ to $j$ of length at least $2$, so the edge is \emph{redundant}. If
$P_1$ uses it, then $j\rightsquigarrow k\rightsquigarrow j$ and $j$ lies on a cycle; if $P_2$ uses
it, then $i\rightsquigarrow k\rightsquigarrow i$ and $i$ lies on a cycle. So every edge of $K$ lies
on a cycle, is redundant, or has an endpoint on a cycle. Let $X_1$, $X_2$, $X_3$ count the
configurations of these three kinds present in $F$ (a cycle with a marked edge; an edge with a
second path between its endpoints; a cycle with a marked vertex and an edge at it). Then
$|K|\le X_1+X_2+X_3$, and each configuration is a fixed set of distinct pairs, present with
probability $p^{\#\text{edges}}$, so linearity of expectation gives
\[
\E|K|\ \le\ \sum_{\ell\ge2}\frac{(mp)^\ell}{\ell}\,\ell
\;+\;\sum_{\ell\ge2}m^{\ell+1}p^{\ell+1}
\;+\;2\sum_{\ell\ge2}\frac{(mp)^\ell}{\ell}\,\ell\,(mp)
\ \le\ \frac{c^2+3c^3}{1-c},
\]
where the three sums count edges on cycles of length $\ell$, edges together with an alternative
path of length $\ell$, and edges with head or tail on a cycle of length $\ell$. Hence
$H(F\mid R)=O(\log m)$, and $H(R)=H(F)-H(F\mid R)=H(F)(1-O(1/m))$.

\emph{(ii) $c>1$.} We use two standard facts about $\mathcal D(m,p)$ with $p=c/m$. Let $\zeta=\zeta(c)$
solve $1-\zeta=e^{-c\zeta}$; it is the survival probability of a Galton--Watson process with
$\mathrm{Poisson}(c)$ offspring, whose total progeny $T$ satisfies $\Pr[T<\infty]=1-\zeta$.
\begin{enumerate}[label=(F\arabic*),leftmargin=*,itemsep=0pt,topsep=2pt]
\item \citep{karp1990transitive,luczak1990phase} If $G_m$ is the vertex set of a largest strongly
connected component, then $|G_m|/m\to\zeta^2$ in probability.
\item For a fixed vertex $u$, let $\mathrm{In}(u)$ be the set of vertices that reach $u$, $u$
included. For every fixed $k$, $\Pr[|\mathrm{In}(u)|\le k]\to\Pr[T\le k]$. Indeed, explore
$\mathrm{In}(u)$ backwards in breadth-first order; while at most $k$ vertices have been found, the
number of new in-neighbours of the current vertex is, given the past, $\mathrm{Bin}(m-j,p)$ with
$j\le k+1$, which tends to $\mathrm{Poisson}(c)$ in total variation, and the event
$\{|\mathrm{In}(u)|\le k\}$ is decided within the first $k$ steps.
\end{enumerate}
Ties between largest components are broken by a fixed rule (say, smallest least vertex), so the
largest strongly connected component $G$ is a function of $R$: the components are the classes of
$\{(i,j):(i,j),(j,i)\in R\}$. Assume $|G|\ge2$ (otherwise there is nothing to count). Call an edge
$(x,y)\in F$ an \emph{entering} edge if $x\notin G$, $y\in G$, and a \emph{leaving} edge if $x\in G$,
$y\notin G$. For $x\notin G$ let $a_x$ be its number of entering edges, and for $y\notin G$ let $b_y$
be its number of leaving edges into it; put $A=\sum_x a_x$ and $B=\sum_y b_y$.

\emph{Step 1: moving a boundary edge keeps the meaning.} Let $(x,y)$ be an entering edge, $y'\in G$
with $(x,y')\notin F$, and $F'=F\setminus\{(x,y)\}\cup\{(x,y')\}$. Put $F_0=F\setminus\{(x,y)\}$. No path
between two vertices of $G$ passes through $x$ (that would put $x$ in $G$), so $G$ is strongly
connected in $F_0$. A simple path uses $(x,y)$ at most once, so $a$ reaches $b$ in $F$ iff $a$
reaches $b$ in $F_0$, or $a=x$ or $a$ reaches $x$ in $F_0$, and $b$ is reached from $y$ in $F_0$
(paths of length $0$ allowed in the last clause). The same holds for $F'$ with $y'$ in place of $y$.
Since $y,y'$ lie in the strongly connected set $G$ of $F_0$, they reach the same vertices, so
$\TC(F')=\TC(F)$; this covers loops $(a,a)$ as well. For a leaving edge $(x,y)$ and $x'\in G$ with
$(x',y)\notin F$, replacing $(x,y)$ by $(x',y)$ also keeps $\TC(F)$: reverse every edge, which
preserves strongly connected components and turns leaving edges into entering ones. After either
move, $R$ is unchanged, hence so is $G$, every edge keeps its type, and every $a_x$ and $b_y$ is
unchanged.

\emph{Step 2: the entropy bound.} Let the class $C$ of $F$ be the set of programs that agree with
$F$ on all non-boundary edges, give each $x\notin G$ an arbitrary set of $a_x$ entering edges
(an $a_x$-subset of $G$ as targets), and give each $y\notin G$ an arbitrary set of $b_y$ leaving
edges (a $b_y$-subset of $G$ as sources). Any two members are joined by a sequence of the moves of
Step~1, so all members have the same $R$, $G$, $a_x$ and $b_y$, and therefore the same class. All
have the same number of edges, hence the same probability $p^{|F|}(1-p)^{U-|F|}$. Since $R$ is a
function of $C$, $H(F\mid R)\ge H(F\mid C)$, and $F$ is uniform on $C$ given $C$, so
\[
H(F\mid R)\ \ge\ \E\log_2|C|\ =\ \E\Bigl[\sum_{x\notin G}\log_2\tbinom{|G|}{a_x}+\sum_{y\notin G}\log_2\tbinom{|G|}{b_y}\Bigr].
\]

\emph{Step 3: the number of boundary edges.} Fix a vertex $u$. Let $F^{u}$ be $F$ with the
out-edges of $u$ removed; the out-edges of $u$ are independent of $F^u$. In $F^u$ the vertex $u$
is a singleton component, so the largest component $G^u$ of $F^u$ is the largest component of a
$\mathcal D(m-1,p)$ graph on the other vertices, and $|G^u|/m\to\zeta^2$ by (F1). A simple path ending
at $u$ uses no out-edge of $u$, so $\mathrm{In}(u)$ is the same in $F$ and $F^u$. Fix $k$ and
$\delta\in(0,\zeta^2)$, and let $\mathcal E=\{|\mathrm{In}(u)|\le k,\ |G^u|\ge(\zeta^2-\delta)m\}$, an event
of $F^u$; for $m$ large, $(\zeta^2-\delta)m>k+1$. On $\mathcal E$, adding the out-edges of $u$ creates new
cycles only through $u$, so it merges into one component $u$ and some vertices of
$\mathrm{In}(u)$, at most $k+1$ vertices in all, and leaves every other component of $F^u$ intact.
Hence $G=G^u$, $u\notin G$, and $a_u=|N^+(u)\cap G^u|$, which given $F^u$ is $\mathrm{Bin}(|G^u|,p)$. Therefore
\begin{align*}
\E\bigl[a_u\mathbf 1[u\notin G]\bigr]\ &\ge\ \E\bigl[p\,|G^u|\,\mathbf 1_{\mathcal E}\bigr]
\ \ge\ c(\zeta^2-\delta)\bigl(\Pr[|\mathrm{In}(u)|\le k]-\Pr[|G^u|<(\zeta^2-\delta)m]\bigr)\\
&\longrightarrow\ c(\zeta^2-\delta)\Pr[T\le k]
\end{align*}
by (F1) and (F2). Letting $k\to\infty$ and $\delta\to0$ gives $\liminf_m\E A/m\ge c\,\zeta^2(1-\zeta)$.
Reversing all edges maps $\mathcal D(m,p)$ to itself and $B$ to $A$, so $\E B=\E A$.

\emph{Step 4: conclusion.} Using $\binom Nk\ge(N/k)^k$, $a_x\le d^+(x)$ and $b_y\le d^-(y)$,
\[
\log_2|C|\ \ge\ (A+B)\log_2|G|\;-\;\sum_{v}\bigl(d^+(v)\log_2d^+(v)+d^-(v)\log_2d^-(v)\bigr),
\]
and the expected sum is at most $2m\,\E[D^2]=O(m)$ for $D\sim\mathrm{Bin}(m-1,p)$. Let
$\mathcal A=\{|G|\ge(\zeta^2-\delta)m\}$. Since $A+B\le|F|$, Cauchy--Schwarz gives
$\E[(A+B)\mathbf 1_{\mathcal A^c}]\le(\E|F|^2)^{1/2}\Pr[\mathcal A^c]^{1/2}=o(m)$, so
\[
\E\bigl[(A+B)\log_2|G|\bigr]\ \ge\ \log_2\bigl((\zeta^2-\delta)m\bigr)\bigl(\E[A+B]-o(m)\bigr)
\ \ge\ 2c\,\zeta^2(1-\zeta)\,m\log_2m\,(1-o(1)).
\]
Dividing by $H(F)=c\,m\log_2m\,(1+o(1))$ gives $\liminf_m H(F\mid R)/H(F)\ge2\zeta^2(1-\zeta)$, and
since $H(R)=H(F)-H(F\mid R)$, $\limsup_m H(R)/H(F)\le1-2\zeta^2(1-\zeta)$.

\emph{Remarks.} (a) Only boundary edges move. Moving an edge \emph{inside} $G$ can change the
meaning: if $G$ is the cycle $v\to v'\to u\to v$, replacing $(u,v)$ by $(u,v')$ disconnects $v$.
(b) The bound on the redundancy tends to $0$ as $c\to\infty$, while the true redundancy should
approach $1$, i.e.\ the compression rate should approach $0$ (the meaning is
then close to the complete relation). The missing part is the edges inside $G$: $R$ constrains them only
through the strong connectivity of $G$, so replacing $F[G]$ by any strongly connected digraph on $G$
keeps $R$. Conditioning on this larger class, and bounding the probability that a
$\mathcal D(|G|,p)$ graph is strongly connected below by $p^{|G|}$ (one Hamiltonian cycle), adds to the
bound of Step~2, for every $m$, the term $\E[(e(G)-|G|)\log_2(1/p)]$, where $e(G)$ is the number of
edges inside $G$. Its Monte-Carlo values are reported in Table~\ref{tab:full-collapse}; we do not
determine its limit.

\emph{Bandwidth.} With rules on independently with probability $\theta$, $H(S)=N h_2(\theta)=\Theta(n)$.
Let $X$ be the largest index of a reachable atom. By the proof of
Corollary~\ref{cor:subcritical}, $\Pr[X\ge j]\le(1-\varepsilon)^{\lfloor j/d\rfloor}$, so $\E X$
and $H(X)$ are $O_{d,\theta}(1)$. Given $X$, $R$ is one of at most $2^X$ subsets of
$\{0,\dots,X\}$, so $H(R)\le H(X)+\E X=O_{d,\theta}(1)$ and $H(R)/H(S)=O(1/n)$.
\hfill$\square$

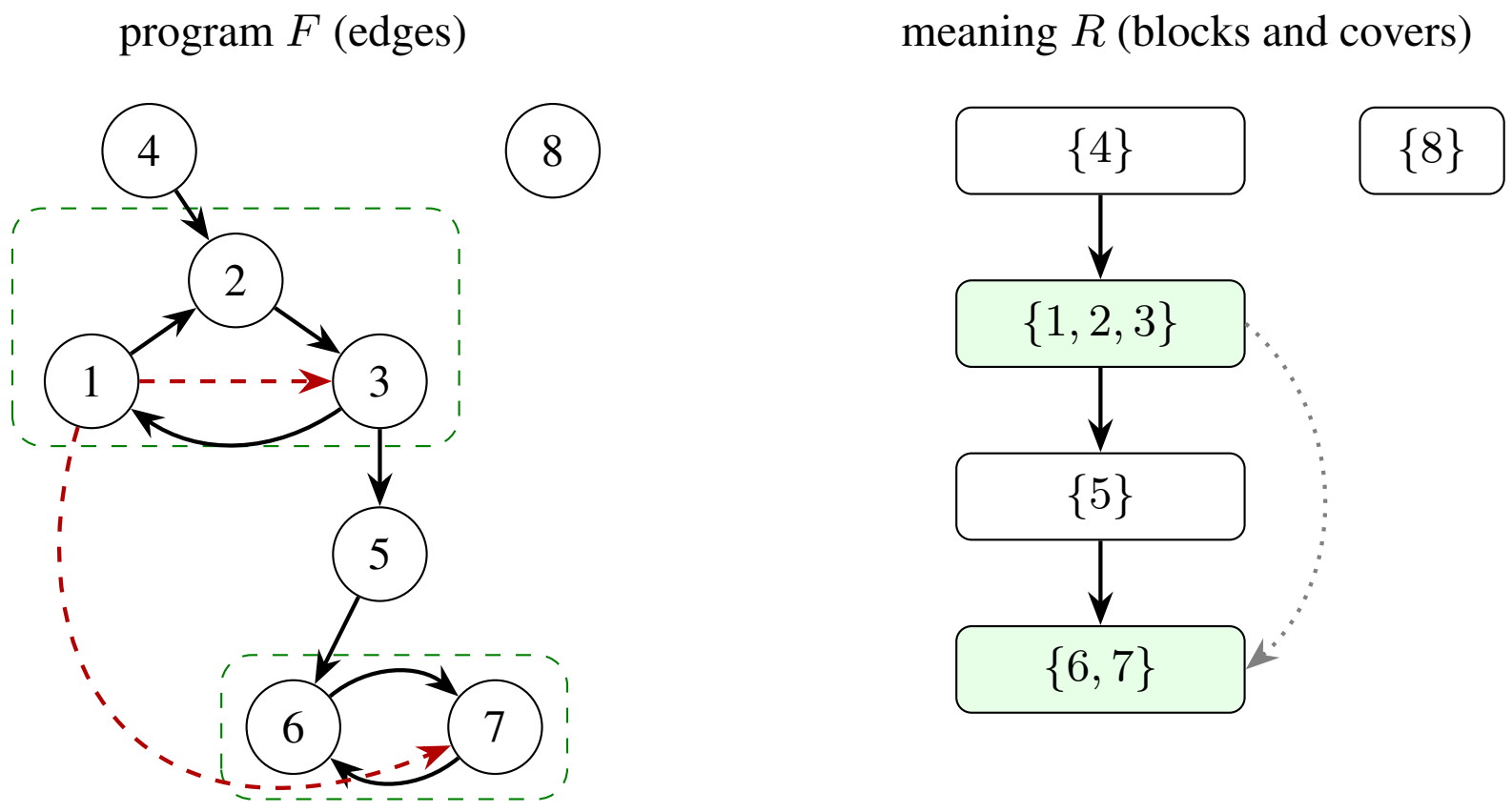
\begin{figure}[t]
\centering
\begin{tikzpicture}[>=Stealth, v/.style={circle,draw,minimum size=6.5mm,inner sep=0pt,font=\small},
  req/.style={->,thick}, red/.style={->,thick,dashed,red!70!black},
  blk/.style={draw,dashed,rounded corners=6pt,green!50!black}]
\node[v] (4) at (0.6,4.2) {4};
\node[v] (8) at (3.4,4.2) {8};
\node[v] (1) at (0.2,2.6) {1};
\node[v] (2) at (1.2,3.3) {2};
\node[v] (3) at (2.2,2.6) {3};
\node[v] (5) at (2.2,1.4) {5};
\node[v] (6) at (1.6,0.2) {6};
\node[v] (7) at (3.0,0.2) {7};
\draw[blk] (-0.35,2.15) rectangle (2.75,3.8);
\draw[blk] (1.1,-0.3) rectangle (3.5,0.7);
\draw[req] (4) -- (2);
\draw[req] (1) -- (2); \draw[req] (2) -- (3);
\draw[req] (3) to[bend left=35] (1);
\draw[red] (1) -- (3);
\draw[req] (3) -- (5); \draw[req] (5) -- (6);
\draw[req] (6) to[bend left=40] (7); \draw[req] (7) to[bend left=40] (6);
\draw[red] (1) .. controls (-0.4,0.6) and (0.6,-0.8) .. (7);
\node[font=\small] at (1.6,5.0) {program $F$ (edges)};
\begin{scope}[xshift=7.2cm]
\tikzset{b/.style={draw,rounded corners=3pt,minimum width=20mm,minimum height=6mm,font=\small}}
\node[b] (B4) at (0,4.2) {$\{4\}$};
\node[b,fill=green!10] (B123) at (0,3.0) {$\{1,2,3\}$};
\node[b] (B5) at (0,1.8) {$\{5\}$};
\node[b,fill=green!10] (B67) at (0,0.6) {$\{6,7\}$};
\node[b,minimum width=10mm] (B8) at (2.3,4.2) {$\{8\}$};
\draw[req] (B4) -- (B123); \draw[req] (B123) -- (B5); \draw[req] (B5) -- (B67);
\draw[->,dotted,thick,gray] (B123.east) to[bend left=50] (B67.east);
\node[font=\small] at (0.6,5.0) {meaning $R$ (blocks and covers)};
\end{scope}
\end{tikzpicture}
\caption{A program and its meaning. Left: $F$ on $8$ vertices; the dashed boxes are the
strongly connected components (blocks) $\{1,2,3\}$ and $\{6,7\}$; solid edges are required (removing
one changes $R$), dashed red edges are redundant ($1\to3$ and $1\to7$). Right: what $R$ records,
namely five blocks and the covers $\{4\}\lessdot\{1,2,3\}\lessdot\{5\}\lessdot\{6,7\}$. The order
$\{1,2,3\}<\{6,7\}$ (dotted) follows by transitivity; $\{8\}$ is incomparable with all others. $R$
does not record the edges inside a block (only that it is strongly connected), nor where an edge
enters or leaves a block (only that it does), nor any edge implied by transitivity.}
\label{fig:blocks}
\end{figure}

\emph{The exact probability of a meaning, and the true compression rate.} The bounds can be checked against
the true value. Given $R$, its strongly connected components $C$ and the strict order $<$ that $R$
induces on them are determined, and $\TC(F)=R$ holds iff (1) each $F[C]$ is strongly connected,
(2) every edge between two components goes from $A$ to $B$ with $A<B$, and (3) every covering pair
$A\lessdot B$ (no component strictly between) carries at least one edge. These conditions concern
disjoint sets of pairs, so with $q=1-p$ (for illustration see Figure~\ref{fig:blocks})
\[
P(R)\;=\;\prod_{C}s_{|C|}\prod_{A\lessdot B}\bigl(1-q^{|A||B|}\bigr)\prod_{A\ne B,\ A\not<B}q^{|A||B|},
\]
where $s_k$ is the probability that $\mathcal D(k,p)$ is strongly connected. Inclusion--exclusion over
the source components gives $s_k$ exactly; this is the classical recursion for strongly connected
digraphs \citep{liskovets1970number,wright1971number}, weighted by edge probabilities: with $\eta_n=1-\sum_{k<n}\binom nk q^{k(n-k)}\eta_k$,
\[
s_n\;=\;\eta_n+\sum_{j<n}\tbinom{n-1}{j-1}s_j\,q^{2j(n-j)}\,\eta_{n-j},
\]
evaluated in high precision. Then $H(F\mid R)=\E[\log_2P(R)-\log_2P(F)]$, which we estimate by
sampling $F$.

The same formula describes an optimal code for the meaning. $R$ is determined by its strongly
connected components and the covering pairs of their order, and every partition of $[m]$ with a
strict order on its blocks arises from exactly one $R$. Codewords of length
$\lceil-\log_2P(R)\rceil$ satisfy Kraft's inequality and have expected length within one bit of
$H(R)$, and by the product formula this length splits into the cost of the blocks being strongly
connected, of each cover carrying an edge, and of no edges against the order. Such a code never
sends the edges inside a block, which vertices carry the edges of a cover, or the edges implied by
transitivity. Sending $F$ instead and letting the decoder compute $\TC(F)$ costs $H(F)$, i.e.\
exactly $H(F\mid R)$ more. The formula reproduces full enumeration for $m\le5$, and the results are stable in the
working precision. Appendix~\ref{app:newexp} reports the true semantic redundancy for $200$, $500$ and $1000$,
Monte-Carlo estimates of both bounds for $m\in\{500,1000,2000\}$, and $\E|K|$ for $m$ up to
$8000$; see also Figure~\ref{fig:new}.

\subsection{The realizable mistake bound and the Littlestone sandwich: Theorem~\ref{thm:halving}}\label{app:ldim}

We prove Theorem~\ref{thm:halving}: Halving over $\Om$ makes at most $\CL$ mistakes, the exact
mistake-bound value is $\Ldim(\Om)$, and $\VC(\Om)\le\Ldim(\Om)\le\log_2|\Om|=\CL$. All three
statements hold for any finite concept class; we state them for $\Om$.

\begin{proof}[Proof of Theorem~\ref{thm:halving}]
\emph{Halving mistake bound.} In the realizable setting the target $R^\ast\in\Om$ labels every
atom by $y=\mathbf 1[\gamma\in R^\ast]$. Halving maintains the version space
$V\subseteq\Om$ of contents consistent with all labels seen and predicts on a queried atom
$\gamma$ by majority vote, $\hat y=\mathbf 1[\,|\{R\in V:\gamma\in R\}|\ge|V|/2\,]$. On a
mistake the true label disagrees with the majority, so at least half of $V$ is inconsistent
with that label and is pruned; hence $|V|$ at least halves on every mistake. Since $R^\ast$ is
never pruned, $|V|\ge 1$ throughout, and $|V|=|\Om|$ initially, so the number of mistakes is at
most $\log_2|\Om|=\CL$.

\emph{$\Ldim$ is the exact mistake bound.} The optimal (adversarial) mistake bound of the
realizable online problem equals the Littlestone dimension $\Ldim(\Om)$
\citep{littlestone1988learning}: an adversary drives a mistake tree of depth $\Ldim(\Om)$ and no
learner can be forced to err more often.

\emph{The sandwich $\VC(\Om)\le\Ldim(\Om)\le\log_2|\Om|$.} For the upper bound, Halving witnesses
a learner with mistake bound $\log_2|\Om|$, and $\Ldim(\Om)$ is the \emph{minimal} achievable
mistake bound, so $\Ldim(\Om)\le\log_2|\Om|$. For the lower bound, let $T\subseteq\Bb$ be
shattered by $\Om$, $|T|=\VC(\Om)$. Ordering $T$ arbitrarily and branching on the two labels at
each atom yields a complete binary mistake tree of depth $|T|$: every root-to-leaf path is a
label pattern on $T$, and shattering supplies a content realizing it, so the tree is
$\Om$-realizable. Hence $\Ldim(\Om)\ge\VC(\Om)$ \citep{littlestone1988learning}.
\end{proof}

\subsection{The agnostic minimax rate: proof of Theorem~\ref{thm:agnostic}}
\label{app:agnostic}

We prove the two-sided bound of Theorem~\ref{thm:agnostic}: for every finite content class $\Om$
and every $T\ge\Ldim(\Om)$,
\[
c_1\sqrt{\Ldim(\Om)\,T}\;\le\;
\min_{\text{learner}}\max_{y_{1:T}}\E\,\reg^{0/1}_T\;\le\;
\sqrt{\tfrac{\ln 2}{2}\,\CL\,T},\qquad c_1=\tfrac14.
\]
Throughout, each content $R\in\Om$ acts as the expert predicting $h_R(\gamma_t)=\mathbf
1[\gamma_t\in R]$, the loss is $0/1$, and the comparator is the best content in hindsight.

\paragraph{Upper bound (Hedge over the content-experts).}
Run the exponential-weights forecaster over the $M=|\Om|$ content-experts with learning rate
$\eta>0$: maintain weights $w_t(R)\propto\exp(-\eta L_{t-1}(R))$, where $L_{t-1}(R)$ is the
cumulative $0/1$ loss of $R$ through round $t-1$, and predict $\hat y_t=1$ with probability
$\sum_{R}w_t(R)\,h_R(\gamma_t)$. With losses in $[0,1]$, the standard exponential-weights
inequality \citep[Thm.~2.2]{cesabianchi2006prediction} gives, against the best expert
$R^\star$,
\[
\E\,\reg^{0/1}_T\;\le\;\frac{\ln M}{\eta}\;+\;\frac{\eta T}{8}.
\]
Optimizing $\eta=\sqrt{8\ln M/T}$ makes the two terms equal and yields
$\E\,\reg^{0/1}_T\le\sqrt{(T/2)\ln M}$. Since $M=|\Om|$ and $\ln M=(\ln 2)\log_2|\Om|=(\ln
2)\CL$, this is $\sqrt{(\ln 2/2)\,\CL\,T}$, holding for \emph{any} finite class regardless of its
Littlestone dimension. (A doubling trick over $\eta$ removes the known-horizon assumption at the
cost of a constant.)

\paragraph{Lower bound (a shattered mistake tree, played in phases).}
Let $d=\Ldim(\Om)$ and let $\Tt$ be a complete binary mistake tree of depth $d$ shattered by
$\Om$: to each internal node $v$ is attached a query atom $\gamma_v\in\Bb$, and for every
root-to-leaf path there is a content $R\in\Om$ whose labels on the path's $d$ atoms are exactly
the branch labels taken along the path. Assume $T\ge d$ and set $L=\lfloor
T/d\rfloor\ge1$; we use the first $dL\le T$ rounds and on the remaining rounds present an
arbitrary atom with fair-coin labels, which contributes zero expected regret.

We describe a \emph{randomized} adversary; since the minimax regret dominates the value against
any fixed adversary (Yao), the resulting expected regret is a lower bound. Play in $d$ phases,
starting at the root $v_1$. In phase $i$ we sit at a node $v_i$ and, for $L$ consecutive rounds,
present the \emph{same} atom $\gamma_{v_i}$ with i.i.d.\ labels $y\sim\mathrm{Unif}\{0,1\}$. Let
$\mathrm{maj}_i\in\{0,1\}$ be the majority label of phase $i$; move to the child of $v_i$ selected
by $\mathrm{maj}_i$, obtaining $v_{i+1}$. This traces one root-to-leaf path $v_1,\dots,v_d$ and a
node-labelling $(\mathrm{maj}_1,\dots,\mathrm{maj}_d)$; by shattering there is a content
$R^\star\in\Om$ with $h_{R^\star}(\gamma_{v_i})=\mathrm{maj}_i$ for every $i$.

\emph{Learner's loss.} Within phase $i$ the atom $\gamma_{v_i}$ is fixed and the $L$ labels are
fresh fair coins independent of everything the learner has observed. Hence every prediction is
correct with probability exactly $\tfrac12$, and the learner's expected loss in phase $i$ is
$L/2$, for a total of $\E[\text{learner}]=dL/2$.

\emph{Comparator's loss.} The content $R^\star$ predicts $\mathrm{maj}_i$ throughout phase $i$, so
its phase-$i$ loss is the minority count $\tfrac{L}{2}-\tfrac12|S_i|$, where $S_i=\sum_{s=1}^L
(2y_{i,s}-1)$ is a sum of $L$ Rademacher signs. By Khintchine's inequality (sharp $L^1$ constant),
$\E|S_i|\ge L^{1/2}/\sqrt2$, so
\[
\E[R^\star]\;=\;\sum_{i=1}^d\Big(\tfrac{L}{2}-\tfrac12\E|S_i|\Big)
\;\le\;\frac{dL}{2}-\frac{d}{2}\cdot\frac{\sqrt L}{\sqrt2}.
\]

\emph{Regret.} Subtracting, and using $\min_{R}L_T(R)\le L_T(R^\star)$,
\begin{align*}
\min_{\text{learner}}\max_{y_{1:T}}\E\,\reg^{0/1}_T
\;& \ge\;\E[\text{learner}]-\E[R^\star]
\;\ge\;\frac{d\sqrt L}{2\sqrt2}\\
\;&=\;\frac{1}{\sqrt8}\sqrt{d\,(dL)}
\;\ge\;\frac{1}{\sqrt8}\sqrt{\tfrac12\,d\,T}
\;=\;\frac14\sqrt{\Ldim(\Om)\,T},
\end{align*}
using $dL\ge T/2$ for $T\ge d$. This is the claimed $c_1\sqrt{\Ldim(\Om)\,T}$ with $c_1=\tfrac14$.

\subsection{Linear templates in general: union-closure, VC dimension, fibres, consistency}\label{app:linear}

All templates in our experiments (bandwidth, Florentine, \textsf{carnivore}) are \emph{linear}: every
clause of $L$ has a single body atom, $r_v\leftarrow r_u$, and $\BK$ declares a source set $V_0$.
Read each clause as an edge $u\to v$ of the \emph{rule graph} $G=(V,L)$; then $\mu(S)$ is the set of
vertices reachable from $V_0$ in $(V,S)$, and Proposition~\ref{prop:realize} describes $\Om$. The
next proposition shows that every learning-theoretic quantity of this paper is then a
reachability property of $G$, computable without listing $\Om$. For $U\subseteq V$, $G-U$ denotes
$G$ with the vertices of $U$ deleted, and $G[U]$ the subgraph induced by $U$.

\begin{proposition}[Linear templates]\label{prop:linear}
Let $L$ be a linear template with rule graph $G$ and sources $V_0$.
\begin{enumerate}[label=(\alph*),leftmargin=*,itemsep=1pt,topsep=2pt]
\item \emph{Union-closure.} If $R_1,R_2\in\Om$ then $R_1\cup R_2\in\Om$.
\item \emph{VC dimension.} A set $A\subseteq V$ is shattered by $\Om$ iff $A\cap V_0=\emptyset$ and every
$a\in A$ is reachable from $V_0$ in $G-(A\setminus\{a\})$. Hence $\VC(\Om)$ is the largest size of
such a set.
\item \emph{Fibres.} For $R\in\Om$,
$\;|\mu^{-1}(R)|=g_G(R)\cdot2^{\,|\{(u,v)\in L:\ u\notin R\}|}$, where $g_G(R)$ is the number of edge sets
$S\subseteq L\cap(R\times R)$ in which every vertex of $R$ is reachable from $V_0$. In particular, for a
uniformly random $S$, $H(S\mid R)=\E\log_2|\mu^{-1}(R)|$.
\item \emph{Consistency.} Given atoms labelled positive ($P$) and negative ($N$), some $R\in\Om$ has
$P\subseteq R$ and $R\cap N=\emptyset$ iff $N\cap V_0=\emptyset$ and $P\subseteq R_{\max}$, where $R_{\max}$ is
the set of vertices reachable from $V_0$ in $G-N$; and $R_{\max}$ is then the largest consistent
content. It is computed in time $O(|V|+|L|)$.
\end{enumerate}
\end{proposition}

\begin{proof}
(a) Every vertex of $R_i$ is reachable from $V_0$ inside $G[R_i]\subseteq G[R_1\cup R_2]$, and
$V_0\subseteq R_1\cup R_2$; apply Proposition~\ref{prop:realize}.

(b) If $A$ is shattered, the pattern $\emptyset$ forces $A\cap V_0=\emptyset$, and the pattern $\{a\}$ gives
some $R\in\Om$ with $R\cap A=\{a\}$; by Proposition~\ref{prop:realize}, $a$ is reachable from $V_0$
inside $G[R]\subseteq G-(A\setminus\{a\})$. Conversely, given a pattern $A'\subseteq A$, choose for each
$a\in A'$ a path $\pi_a$ from $V_0$ to $a$ in $G-(A\setminus\{a\})$, and let $R$ be $V_0$ together with
the vertices of these paths. Each vertex of $R$ is reachable from $V_0$ along its path, which lies
in $G[R]$, so $R\in\Om$; no path visits $A\setminus A'$ and $V_0\cap A=\emptyset$, so $R\cap A=A'$.

(c) $\mu(S)=R$ holds iff (i) $S$ contains no edge from $R$ to $V\setminus R$, since its head would be
reached, and (ii) $S\cap(R\times R)$ reaches every vertex of $R$; edges whose tail lies outside $R$
are never used, so they are unconstrained. Counting the three kinds of edges gives the formula.
For uniform $S$, $S$ is uniform on $\mu^{-1}(R)$ given $R$.

(d) If $R$ is consistent, then $V_0\subseteq R$ gives $N\cap V_0=\emptyset$, and every vertex of $R$ is
reachable inside $G[R]\subseteq G-N$, so $P\subseteq R\subseteq R_{\max}$. Conversely, every vertex of $R_{\max}$
is reachable from $V_0$ along a path in $G-N$ whose vertices are all in $R_{\max}$, so
$R_{\max}\in\Om$ by Proposition~\ref{prop:realize}; it avoids $N$ and contains $P$. One breadth-first
search computes it.
\end{proof}

\emph{Consequences.} By (d), predicting with $R_{\max}$ (reachability with the negatively
labelled atoms deleted) is a linear-time learner that is always consistent in the realizable
case, so its PAC sample complexity is $O\bigl((\VC(\Om)\log\tfrac1\varepsilon+\log\tfrac1\delta)/\varepsilon\bigr)$
\citep{blumer1989learnability}, with $\VC(\Om)$ given by (b). Lemma~\ref{lem:vc-bandwidth} below is
(b) for the bandwidth graph: there, $a$ is cut off from $r_0$ exactly when $A\setminus\{a\}$ contains $d$
consecutive atoms below $a$. The fibre formula (c) is how the $4{,}104$ Florentine fibre sizes of
Appendix~\ref{app:baseline} are computed. We checked (a)--(d) against full enumeration on $300$ random rule
graphs with cycles and up to two sources.

\emph{Beyond linear templates.} For nonlinear templates, (a) and the sufficiency in (b) fail,
because proofs of different atoms combine: for transitive closure, $\{(1,2)\}$ and $\{(2,3)\}$ are
contents but their union is not, as it is not transitively closed. The necessity part of (b)
remains valid, and the VC dimension of transitive closure is handled directly
(Appendix~\ref{app:nonlinear}).

\subsection{The bandwidth-\texorpdfstring{$d$}{d} rate: VC, Littlestone: Corollary~\ref{cor:agnostic-bandwidth}}\label{app:ldim-d}

We prove Corollary~\ref{cor:agnostic-bandwidth}: for the bandwidth-$d$ family
$\Ldim(\Om(n,d))=\Theta(\CL)$, so both sides of Theorem~\ref{thm:agnostic} are
$\Theta(\sqrt{\CL\,T})$. We identify a content with its indicator on the atoms
$r_1,\dots,r_n$; the source $r_0$ lies in every content and carries no shattering information, so
we drop it and write $[n]=\{1,\dots,n\}$. Now, $R\in\Om(n,d)$ iff $0\in R$ and,
listing $R=\{0=x_0<x_1<\dots<x_m\}$, every gap satisfies $x_\ell-x_{\ell-1}\le d$.

\begin{definition}[$d$-gap-free sets]\label{def:gapfree}
A set $T\subseteq[n]$ is \emph{$d$-gap-free} if it contains no $d$ consecutive integers, i.e.\
there is no $t$ with $\{t,\dots,t+d-1\}\subseteq T$. Let
$g(n,d):=\max\{|T|:T\subseteq[n]\text{ is }d\text{-gap-free}\}$.
\end{definition}

\begin{lemma}[VC dimension of the bandwidth family]\label{lem:vc-bandwidth}
For all $n\ge1$, $d\ge1$,
\[
g(n,d)\;\le\;\VC(\Om(n,d))\;\le\;g(n,d)+d,\qquad g(n,d)=n-\Big\lfloor\tfrac nd\Big\rfloor .
\]
In particular $\VC(\Om(n,d))=\big(1-\tfrac1d\big)n+O(1)$.
\end{lemma}
\begin{proof}
\emph{Every $d$-gap-free $T$ is shattered.} Fix such a $T$ with $|T|=k$ and a target
$\sigma\in\{0,1\}^T$; write $P=\sigma^{-1}(1)$. We build $R\in\Om$ with $R\cap T=P$ and
$R\cap\sigma^{-1}(0)=\emptyset$, free on $[n]\setminus T$. Start from $R_0:=\{0\}\cup P$, which
already has the right labels on $T$, and repair realizability without touching $\sigma^{-1}(0)$.
For $v\in R_0$ with $v>0$, the window $I_v:=\{v-d,\dots,v-1\}\cap\{0,\dots,n\}$ must contain an
allowed atom: otherwise all $d$ integers of $I_v$ lie in $\sigma^{-1}(0)\subseteq T$, making
$T\supseteq I_v$ a block of $d$ consecutive integers, contradicting $d$-gap-freeness (when
$v\le d$ the atom $0\in I_v$ is always allowed). Switch on the largest allowed $u<v$ and recurse
on it; the index strictly decreases, so the recursion halts at $0$. The resulting $R$ has $0\in
R$ and every $v\in R\setminus\{0\}$ an in-$R$-neighbour within $d$, so $R\in\Om$ and $R\cap T=P$. As $\sigma$ was arbitrary, $T$ is shattered; thus
$\VC(\Om(n,d))\ge g(n,d)$.

\emph{A shattered $T$ is $d$-gap-free up to a top block.} Suppose $T\supseteq
B:=\{t,\dots,t+d-1\}$ and some $w\in T$ exceeds $t+d-1$. Take $\sigma$ labelling all of $B$ with
$0$ and $w$ with $1$. If $R\in\Om$ realized $\sigma$ then $w\in R$ while $B\cap R=\emptyset$; the elements of $R$ below $w$ form a chain $0=y_0<\dots<y_s$ with all gaps
$\le d$, and $w-y_s\le d$ forces $y_s\ge w-d\ge t$. Descending, a single step of size $\le d$
must land the chain in $[t,t+d-1]=B$, exhibiting an element of $B\cap R=\emptyset$ --- a
contradiction. So $\sigma$ is unrealizable and $T$ is not shattered: any block a shattered $T$
contains sits at its very top, contributing at most $d-1$ extra indices, whence
$\VC(\Om(n,d))\le g(n,d)+d$.

\emph{Evaluation of $g(n,d)$.} Partitioning $[n]$ into $\lfloor n/d\rfloor$ length-$d$ blocks
(and a shorter tail), a $d$-gap-free set omits at least one integer per block, so $|T|\le
n-\lfloor n/d\rfloor$. Conversely $T^\star=[n]\setminus\{d,2d,\dots,d\lfloor n/d\rfloor\}$
contains no $d$ consecutive integers (every length-$d$ window holds a multiple of $d$) and has
size $n-\lfloor n/d\rfloor$. Hence $g(n,d)=n-\lfloor n/d\rfloor$.
\end{proof}

\begin{proof}[Proof of Corollary~\ref{cor:agnostic-bandwidth}]
The sandwich $\VC(\Om)\le\Ldim(\Om)\le\log_2|\Om|=\CL$ of Theorem~\ref{thm:halving}, applied to
$\Om=\Om(n,d)$, has left side $\big(1-\tfrac1d\big)n+O(1)$ by Lemma~\ref{lem:vc-bandwidth} and right
side $(\log_2\rho_d)\,n+O(1)$ by Theorem~\ref{thm:count}. Both are linear in $n$ with strictly positive
slope ($1-\tfrac1d>0$ and $\log_2\rho_d>0$ for $d\ge2$), so
$\Ldim(\Om(n,d))=\Theta(n)=\Theta(\CL)$. Substituting into Theorem~\ref{thm:agnostic}, the lower bound
$c_1\sqrt{\Ldim\,T}$ and the upper bound $\sqrt{(\ln2/2)\,\CL\,T}$ are both
$\Theta(\sqrt{\CL\,T})$, so the minimax agnostic $0/1$ regret is $\Theta(\sqrt{\CL\,T})$.
\end{proof}

\begin{remark}[The trapped constant]\label{rem:constant}
Corollary~\ref{cor:agnostic-bandwidth} settles the order but not
$\lim_n\Ldim(\Om(n,d))/n$, which is confined to $[\,1-\tfrac1d,\ \log_2\rho_d\,]$
($=[0.5,\,0.6942]$ for $d=2$). The interval is nonempty and strictly inside $(0,1)$, and its
width $\log_2\rho_d-(1-\tfrac1d)\to0$ as $d\to\infty$, so the constant is asymptotically
pinned for large reach. Exact computation of $\Ldim$ by the recursion
$\Ldim(\mathcal H)=\max_x\bigl(1+\min_b\Ldim(\mathcal H_{x\mapsto b})\bigr)$ gives, for $d=2$ and
$n=6,7,8,9$, the values $4,5,5,6$ against $\CL=5.04,5.75,6.46,7.16$.
\end{remark}

\subsection{Minimax log-loss regret is the Shtarkov sum (Theorem~\ref{thm:shtarkov})}
\label{app:th3}

\begin{proof}
Fix the query sequence $\gamma_{1:T}$. For an outcome $y\in\{0,1\}^T$ write
$M(y)=\max_{R\in\Om}P_R(y)$ and $S=\sum_{y}M(y)$ (the Shtarkov sum). A strategy with
sequence likelihood $\hat P$ suffers cumulative log loss $-\log_2\hat P(y)$, while the best
expert suffers $\min_R(-\log_2 P_R(y))=-\log_2 M(y)$; so its regret on $y$ is
$\log_2\!\bigl(M(y)/\hat P(y)\bigr)$ and its worst-case regret is
$\max_y\log_2\!\bigl(M(y)/\hat P(y)\bigr)$.

\emph{Achievability.} The normalized-maximum-likelihood distribution
$q^\star(y)=M(y)/S$ \citep{shtarkov1987universal} is a probability distribution on
$\{0,1\}^T$ and makes the ratio constant, $M(y)/q^\star(y)\equiv S$, so its worst-case
regret is exactly $\log_2 S$.

\emph{Optimality.} For any $\hat P$ with $\sum_y\hat P(y)\le1$, the mediant inequality gives
$\max_y M(y)/\hat P(y)\ge\sum_y M(y)/\sum_y\hat P(y)\ge S$. Hence
$\min_{\hat P}\max_y\log_2(M(y)/\hat P(y))=\log_2 S$, attained uniquely by $q^\star$.

\emph{Evaluating $S$.} Here each expert is deterministic:
$P_R(y)=\prod_t\mathbf1[y_t=\mathbf1[\gamma_t\in R]]\in\{0,1\}$ is the indicator that $y$ is
$R$'s label pattern on $\gamma_{1:T}$. Thus $M(y)\in\{0,1\}$ equals $1$ iff $y$ is some
content's pattern, and $S=\sum_y M(y)$ counts the distinct patterns,
$S=|\Om|_{\gamma_{1:T}}|$. Restriction can only merge contents, so $S\le|\Om|$, with equality
once the queried atoms separate all contents; and by Sauer--Shelah
$S\le\sum_{i\le\VC(\Om)}\binom{T}{i}\le(T+1)^{\VC(\Om)}$, giving
$\reg^\star_T\le\min\{\CL,\VC(\Om)\log_2(T+1)\}$. The precise minimax evaluation is that of \citet{drmota2004precise}.
\end{proof}

\subsection{Proof of Theorem~\ref{thm:vsp}} 
\label{app:vsp}

\begin{proof}[Proof of Theorem~\ref{thm:vsp}]
Write $V_t=\{R\in\Om:\ R\text{ consistent with }y_{<t}\}$ for the version space entering
round $t$, so $V_1=\Om$ and $V_{t+1}=\{R\in V_t:\mathbf 1[\gamma_t\in R]=y_t\}$. Since the
outcome is realizable, the generating target $R^\star\in\Om$ is consistent with every
label, hence $R^\star\in V_t$ for all $t$; as $w(R^\star)>0$ this gives $w(V_t)\ge
w(R^\star)>0$, so every prediction is well defined and every per-round loss is finite.

\emph{(i) Exact regret.} The algorithm predicts
$\hat p_t=w(\{R\in V_t:\gamma_t\in R\})/w(V_t)$ for label $1$. If $y_t=1$ then
$V_{t+1}=\{R\in V_t:\gamma_t\in R\}$ and $\hat p_t(y_t)=\hat p_t=w(V_{t+1})/w(V_t)$; if
$y_t=0$ then $V_{t+1}=\{R\in V_t:\gamma_t\notin R\}$ and
$\hat p_t(y_t)=1-\hat p_t=w(V_{t+1})/w(V_t)$. Either way
\[
\hat p_t(y_t)=\frac{w(V_{t+1})}{w(V_t)}.
\]
The sequential mass assigned to $y$ therefore telescopes,
\[
\widehat P(y)=\prod_{t=1}^{T}\hat p_t(y_t)
=\prod_{t=1}^{T}\frac{w(V_{t+1})}{w(V_t)}
=\frac{w(V_{T+1})}{w(V_1)}=\frac{w(V_{T+1})}{w(\Om)}.
\]
In the realizable case the best content incurs zero log-loss: any $R$ consistent with the
whole outcome (e.g.\ $R^\star$) predicts each observed label with probability $1$, so
$\min_{R\in\Om}\sum_t\ell^{\log}_t(p^R_t,y_t)=0$. Hence the cumulative regret on $y$ is
\[
\reg_T=-\log_2\widehat P(y)=\log_2\frac{w(\Om)}{w(V_{T+1})}
\ \le\ \log_2\frac{w(\Om)}{w(R^\star)},
\]
the inequality because $R^\star\in V_{T+1}$. Under $w\equiv1$ this reads
$\reg_T=\log_2(|\Om|/|V_{T+1}|)\le\log_2|\Om|=\CL$, since $|V_{T+1}|\ge1$.

\emph{(ii) Lower bound and optimality.} Let $\hat q=(\hat q_t)$ be any sequential log-loss
predictor. Its conditionals satisfy $\hat q_t(0\mid y_{<t})+\hat q_t(1\mid y_{<t})=1$, so
$\widehat Q(y):=\prod_t\hat q_t(y_t\mid y_{<t})$ defines a (sub-)probability on
$\{0,1\}^T$ with $\sum_{y}\widehat Q(y)\le1$. On a realizable outcome the best content again
has zero loss, so the predictor's regret there is $-\log_2\widehat Q(y)$, and the realizable
outcomes are exactly the $S_T=|\Om|_{\gamma_{1:T}}|$ distinct label patterns
(Theorem~\ref{thm:shtarkov}). If its worst-case regret over realizable outcomes were
$<\log_2 S_T$, then $\widehat Q(y)>1/S_T$ for each of the $S_T$ patterns; summing over them
gives $\sum_{y\ \mathrm{pattern}}\widehat Q(y)>1\ge\sum_y\widehat Q(y)$, a contradiction.
Hence every predictor has worst-case regret at least $\log_2 S_T=\reg^\star_T$, the minimax
value of Theorem~\ref{thm:shtarkov}.

Now compare Algorithm~\ref{alg:vsp} (uniform $w$) to this floor. The map sending each
$R\in\Om$ to its pattern on $\gamma_{1:T}$ partitions $\Om$ into the $S_T$ nonempty fibres
$\{V_{T+1}(y):y\text{ realizable}\}$, so $\sum_{y\ \mathrm{pattern}}|V_{T+1}(y)|=|\Om|$.
When the queries separate the contents, $S_T=|\Om|$ and every fibre is a singleton:
$|V_{T+1}(y)|=1$ for all realizable $y$, so by (i) the algorithm's regret is
$\log_2|\Om|=\log_2 S_T=\reg^\star_T$ on \emph{every} outcome; its worst case meets the
floor and it is minimax-optimal. In general (i) gives the data-dependent regret
$\log_2(|\Om|/|V_{T+1}(y)|)$, which differs from the constant minimax value $\log_2 S_T$ by
\[
\log_2\frac{|\Om|}{S_T\,|V_{T+1}(y)|},
\]
positive on small fibres and negative on large ones (the fibre sizes average $|\Om|/S_T$).
Thus Algorithm~\ref{alg:vsp} is anytime and horizon-free but equalizes the regret across
outcomes only when it separates the contents; the constant value $\log_2 S_T$ on every
outcome is attained by the horizon-dependent NML strategy of
Theorem~\ref{thm:shtarkov} alone.
\end{proof}

The only per-step primitive is a weighted count over the version space. For the
bandwidth-$d$ family this is a single $2^d\times 2^d$ transfer-matrix pass, $O(2^d n)$ per
step; on a tree (Appendix~\ref{app:wordnet}) it is one $O(n)$ dynamic program, so the mixture over
$|\Om|$ contents is evaluated without ever enumerating $\Om$. A non-uniform prior $w$
changes the guarantee to $\log_2\!\bigl(w(\Om)/w(R^\star)\bigr)$, the code length of the
true theory under $w$: a good language bias buys smaller redundancy, exactly as in
universal coding \citep{grunwald2007mdl}.

\section{Algorithm for the mismatch loss}\label{app:ewa}

Algorithm~\ref{alg:vsp} prunes: on each label it discards the contents inconsistent with it.
Pruning is correct in the realizable log-loss setting, where the target is never discarded, but
it is wrong for the agnostic mismatch loss, where a content that errs once may still be the best
comparator in hindsight; Section~\ref{sec:exp} shows that under label noise pruning loses the
target in every run. The fix is to reweight rather than prune: each content carries a weight that
decays geometrically in its accumulated mistakes, so a single error is penalised but not fatal.
This is the exponential-weights forecaster over the content class $\Om$, made explicit with a
prediction rule and an efficient implementation. The only difference from
Algorithm~\ref{alg:vsp} is the last line: Algorithm~\ref{alg:vsp} sets the weight of an
inconsistent content to $0$ (pruning), whereas Algorithm~\ref{alg:mismatch} multiplies it by
$\beta$ (discounting); taking $\beta\to0$ with the majority-vote prediction recovers
Algorithm~\ref{alg:vsp} and, exactly, Halving.

\begin{algorithm}[t]
\caption{Exponential-weights deductive predictor (mismatch loss)}\label{alg:mismatch}
\KwIn{content class $\Om$; query stream $\gamma_1,\gamma_2,\dots$; discount $\beta=e^{-\eta}\in[0,1)$
(default $\eta=\sqrt{8\ln|\Om|/T}$; $\beta\to0$ gives Halving)}
$w(R)\leftarrow1$ for all $R\in\Om$\tcp*{weight $=\beta^{\#\text{mistakes so far}}$}
\For{$t=1,2,\dots$}{
  receive query $\gamma_t$\;
  $s_1\leftarrow\sum_{R\in\Om:\,\gamma_t\in R}w(R)$,\quad
  $s_0\leftarrow\sum_{R\in\Om:\,\gamma_t\notin R}w(R)$\tcp*{weighted vote for/against entailment}
  predict $\hat y_t\leftarrow\mathbf1[s_1\ge s_0]$ (deterministic), or $\hat y_t\leftarrow1$
  w.p.\ $s_1/(s_1+s_0)$ (randomized)\;
  observe $y_t$\;
  $w(R)\leftarrow\beta\,w(R)$ for every $R$ with $\mathbf1[\gamma_t\in R]\ne y_t$\tcp*{discount inconsistent contents}
}
\end{algorithm}

\begin{theorem}[Mismatch guarantee]\label{thm:mismatch}
Run Algorithm~\ref{alg:mismatch} over $\Om$, and write $M=|\Om|$.
\begin{enumerate}[label=(\roman*)]
\item (Realizable limit.) With $\beta\to0$ and majority-vote prediction it is Halving, so in the
realizable case it makes at most $\log_2M=\CL$ mistakes.
\item (Agnostic regret.) With the randomized prediction and $\eta=\sqrt{8\ln M/T}$, against an
arbitrary label sequence,
\[
\E\,\reg^{0/1}_T=\E\Bigl[\sum_t\mathbf1[\hat y_t\ne y_t]\Bigr]-\min_{R\in\Om}\sum_t\mathbf1[h_R(\gamma_t)\ne y_t]
\ \le\ \sqrt{\tfrac{\ln2}{2}\,\CL\,T}.
\]
\end{enumerate}
\end{theorem}

\begin{proof}
Each content $R$ is the expert predicting $h_R(\gamma_t)=\mathbf1[\gamma_t\in R]$, with $0/1$ loss
$\ell_t(R)=\mathbf1[h_R(\gamma_t)\ne y_t]$ and mistake count $m_t(R)=\sum_{s<t}\ell_s(R)$. The
update multiplies by $\beta=e^{-\eta}$ on each mistake, so $w_t(R)=\beta^{m_t(R)}=e^{-\eta m_t(R)}$;
write $W_t=\sum_Rw_t(R)$ and $p_t=s_1/W_t$.

(i) With $\beta=0$ a content that errs is zeroed permanently, so at every round the positive-weight
set is exactly the version space of contents consistent with the labels seen, and the rule
$\mathbf1[s_1\ge s_0]$ is the majority vote over it: this is Halving over $\Om$. The target is
never removed, so $W_t>0$ throughout, and each mistake eliminates at least half the version space;
hence at most $\log_2M=\CL$ mistakes.

(ii) \emph{Loss identity.} For the randomized rule, if $y_t=1$ the prediction errs with probability
$s_0/W_t$, and if $y_t=0$ with probability $s_1/W_t$; in both cases this equals
$\hat\ell_t:=\sum_Rw_t(R)\ell_t(R)/W_t$, the weight-average expert loss, since the erring
contents are precisely $\{R:\gamma_t\notin R\}$ when $y_t=1$ and $\{R:\gamma_t\in R\}$ when
$y_t=0$. \emph{Potential argument.} By Hoeffding's lemma ($\ell_t\in[0,1]$),
\[
\frac{W_{t+1}}{W_t}=\E_{R\sim p_t}e^{-\eta\ell_t(R)}\le\exp\bigl(-\eta\hat\ell_t+\eta^2/8\bigr).
\]
Taking logarithms, summing over $t$, and using $W_1=M$ together with
$W_{T+1}\ge w_{T+1}(R)=e^{-\eta L_T(R)}$ for any fixed $R$ (where $L_T(R)=\sum_t\ell_t(R)$) gives
$\sum_t\hat\ell_t-\min_RL_T(R)\le\ln M/\eta+\eta T/8$. The left side is $\E\,\reg^{0/1}_T$ by the
loss identity. Optimising $\eta=\sqrt{8\ln M/T}$ yields $\sqrt{(T/2)\ln M}=\sqrt{(\ln2/2)\,\CL\,T}$,
using $\ln M=(\ln2)\CL$. (A time-varying $\eta_t=\sqrt{8\ln M/t}$, or a doubling trick, removes
the dependence on the horizon $T$ at the cost of an absolute constant.)
\end{proof}

\paragraph{Implicit implementation.} The weight of a content factorizes over atoms. Let
$a_t(\gamma)$ and $b_t(\gamma)$ count the rounds before $t$ in which atom $\gamma$ was labelled $0$
and $1$. A content $R$ errs on $\gamma$ exactly $a_t(\gamma)$ times if $\gamma\in R$ and
$b_t(\gamma)$ times if $\gamma\notin R$, so
\[
w_t(R)\ =\ \prod_{\gamma\in R}\beta^{a_t(\gamma)}\ \prod_{\gamma\notin R}\beta^{b_t(\gamma)}.
\]
This has the same product form as the prior: any prior that is a product of per-atom factors,
such as the content-uniform prior, the size prior $2^{-|R|}$, or the fibre-size prior on a tree,
can be multiplied in. The sums $s_1,s_0$ are therefore weighted counts of realizable contents
with per-atom weights. On a tree they are computed by the dynamic program of
Appendix~\ref{app:wordnet}, with each child either excluded (contributing the product of the
``out'' factors of its whole subtree) or included; a downward pass then gives the marginal
probability of every atom, in $O(n)$ time per round. For the bandwidth-$d$ family the same sums
are a $2^d\times2^d$ transfer-matrix product, in $O(2^dn)$ time. This is how
Algorithm~\ref{alg:mismatch} runs on the $10^{90.9}$ contents of \textsf{carnivore} in
Section~\ref{sec:exp}; the minimum $\min_RL_T(R)$ needed to report regret is computed by the
same recursion with $(\min,+)$ in place of $(+,\times)$.

\section{Further empirical results on Real Data}\label{app:exps}
This appendix supplements the Florentine experiment of Section~\ref{sec:exp}
(Figure~\ref{fig:florentine}): Appendix~\ref{app:florentine} probes its robustness and the
mechanism behind the exact regret, and Appendix~\ref{app:wordnet} carries the same two laws to a
real taxonomy three orders of magnitude larger, Appendix~\ref{app:baseline} compares
Algorithm~\ref{alg:vsp} with the same learner run over clause-sets, and
Appendix~\ref{app:newexp} gives the complete results of the experiments of
Section~\ref{sec:exp}.

We first spell out the two Florentine experiments: (i) the \emph{semantic collapse} of
syntactic rule-sets onto the content class $\Om$, and (ii) the behaviour of the online
predictor, Algorithm~\ref{alg:vsp}.

\textbf{Data.} Padgett's Florentine families \citep{padgett1993robust} is a real $15$-node network whose edges are marriage
ties between Renaissance families; it is a standard benchmark in social-network analysis. We use
it because our running example --- influence spreading through a network of ties --- is
exactly this kind of object.

\textbf{Template.} Fix one family as the \emph{seed} (default: Medici). The single rule schema
\[
\textsf{marked}(Y)\ \leftarrow\ \textsf{marked}(X),\ \textsf{tie}(X,Y)
\]
is instantiated over both directions of every marriage tie, giving $N=2\times 20 = 40$ ground
clauses. Here $\mathsf{marked}(X)$ is the unary predicate carrying the spread: $\mathsf{marked}(f)=1$ means family $f$ has been reached from the seed, i.e.\ $f$ lies in the current content $R$; it is the domain-neutral reading of the $\mathrm{cheat}$ predicate of Example~\ref{ex:main}, with the marriage tie $\mathsf{tie}(X,Y)$ as fixed background (the analogue of $\mathrm{friend}$).
A \emph{hypothesis} is a subset $S$ of these $40$ clauses --- a choice of which ties
actually conduct the marking. There are $2^{N}=2^{40}\approx 1.1\times 10^{12}$ hypotheses.

\textbf{Meaning.} A hypothesis $S$ does not stand on its own: its \emph{meaning} is the set of
families it ends up marking, i.e.\ the least Herbrand model $\mu(S)=\lhm(\mathrm{BK}\cup S)$
obtained by starting from the seed and firing the switched-on ties to a fixpoint. This is a set
of \emph{atoms} (marked families), not a set of clauses. The object of interest is the set of
all meanings the template can produce,
\[
\Om \;=\; \{\,\mu(S) : S\subseteq L\,\}\ \subseteq\ 2^{\{\text{15 families}\}} .
\]

\textbf{Which sets are meanings?} Not every set of families is realizable. By the realizability
characterization (Proposition~\ref{prop:realize}), for a symmetric tie relation $R\in\Om$ if and only
if $R$ contains the seed and is \emph{connected} in the network. Enumerating these gives exactly
$|\Om|=4{,}104$ meanings.

\textbf{Why the map is many-to-one (a worked case).} Take the meaning
$R=\{\text{Medici},\text{Salviati},\text{Pazzi}\}$, a path in the network. Every hypothesis that
(a) switches on the forward ties $\text{Medici}\!\to\!\text{Salviati}\!\to\!\text{Pazzi}$,
(b) switches \emph{off} any tie from $\{$Medici, Salviati$\}$ to a family outside $R$ (otherwise
the marking would spread further), and (c) does \emph{anything at all} on the reverse ties and on
ties whose source is never reached --- produces exactly this $R$. A whole ``fibre'' of distinct
rule-sets collapses to the one meaning. Summing the fibre sizes over the $4{,}104$ meanings
recovers all $2^{40}$ hypotheses.

\subsection{Florentine families: robustness, regret build-up, and the prior}\label{app:florentine}

\begin{figure}[t]
\centering
\includegraphics[width=0.88\linewidth]{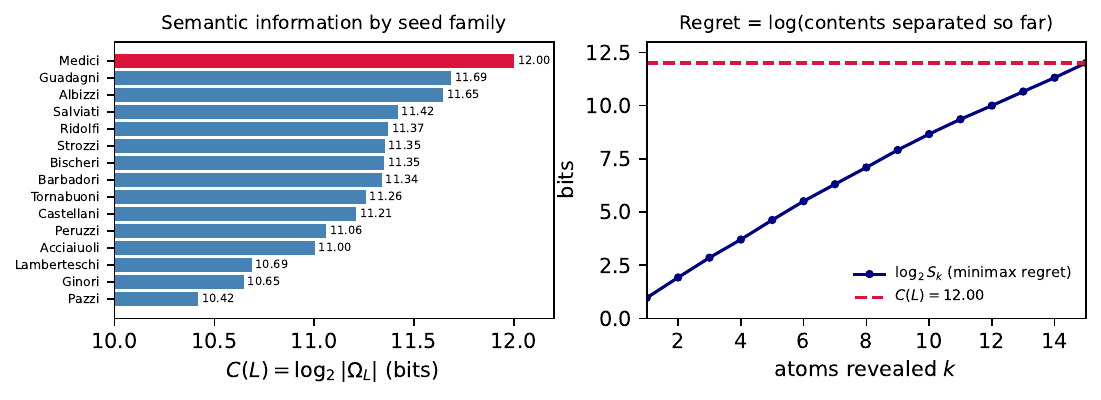}
\caption{\textbf{Left:} semantic information $\CL$ for each seed family (Table~\ref{tab:seed});
Medici (red) is largest. \textbf{Right:} regret build-up --- $\log_2 S_k$, the minimax value after
$k$ queries, rises to $\CL$ as the queries separate the $4{,}104$ contents (Medici seed).}
\label{fig:app-florentine}
\end{figure}

\paragraph{Robustness across the seed.} We repeat the construction with each of the $15$ families
as the seed $E$ (Table~\ref{tab:seed}, Figure~\ref{fig:app-florentine} left). The content count
$|\Om|$, and hence $\CL=\log_2|\Om|$, varies with how central the seed is --- Medici, the most
central family, induces the largest content class ($12.00$ bits), the peripheral Pazzi the
smallest ($10.42$) --- so the semantic information of the same template is a property of where the
evidence is injected. The two laws are nonetheless invariant across all $15$ seeds: the code
length of Algorithm~\ref{alg:vsp} equals $\CL$ exactly, and the mistake count never exceeds it
(Theorems~\ref{thm:vsp} and~\ref{thm:halving}).

\paragraph{How the regret is paid.} The exact-regret identity of Section~\ref{sec:exp} is not an
end-of-run coincidence; it is accrued query by query as the data separate contents. After $k$
atoms the minimax value (Theorem~\ref{thm:shtarkov}) is $\log_2 S_k$, where $S_k=|\Om|_{\gamma_{1:k}}|$
is the number of distinct label-patterns the first $k$ atoms induce on $\Om$. Averaged over query
orders (computed exactly over all $k$-subsets of the $15$ atoms),
$S_k = 1.9,\,3.7,\,7.0,\,13.2,\,24.5,\,44.7,\,80.2,\dots,1724,\,2683,\,4104$ for $k=1,\dots,15$, so
the average of $\log_2 S_k$ climbs monotonically from $0.93$ to $\CL=12.0$ (Figure~\ref{fig:app-florentine}, right). This is the
mechanism behind the identity: the regret is exactly the log-count of contents the queries can
tell apart, and the full content sub-base $\Bb$ separates all $4{,}104$, so the paid code length is
$\log_2|\Om|=\CL$.

\paragraph{Effect of the prior.} Under the uniform prior over contents the code length is the
constant $\CL$ for every target --- the minimax-optimal behaviour of Theorem~\ref{thm:vsp}.
Replacing it with a size prior $w(R)\propto 2^{-|R|}$ makes the code length target-dependent (mean
$13.44$ bits over the $4{,}104$ targets, shorter than $\CL$ on small contents and longer on large
ones): a matched language bias buys shorter codes on typical targets at the cost of worst-case
optimality, exactly the universal-coding trade-off \citep{grunwald2007mdl}. Consistent with
Proposition~\ref{prop:lift}, the capacity-achieving default is uniform on \emph{contents}, not on
the $2^{L}$ clause-sets.

\begin{table}[t]\centering\small
\begin{tabular}{@{}lrrcc@{}}
\toprule
seed family & $|\Om|$ & $\CL$ (bits) & code length & mistakes (mean / max)\\
\midrule
Medici & 4104 & 12.00 & $\CL$ & 5.3 / 10\\
Guadagni & 3294 & 11.69 & $\CL$ & 5.1 / \phantom{0}9\\
Albizzi & 3206 & 11.65 & $\CL$ & 5.3 / 10\\
Salviati & 2738 & 11.42 & $\CL$ & 5.0 / \phantom{0}8\\
Ridolfi & 2648 & 11.37 & $\CL$ & 5.1 / \phantom{0}8\\
Bischeri & 2614 & 11.35 & $\CL$ & 5.1 / \phantom{0}8\\
Strozzi & 2616 & 11.35 & $\CL$ & 4.9 / \phantom{0}8\\
Barbadori & 2590 & 11.34 & $\CL$ & 5.0 / \phantom{0}8\\
Tornabuoni & 2450 & 11.26 & $\CL$ & 4.7 / \phantom{0}8\\
Castellani & 2370 & 11.21 & $\CL$ & 4.8 / \phantom{0}8\\
Peruzzi & 2136 & 11.06 & $\CL$ & 4.8 / \phantom{0}9\\
Acciaiuoli & 2053 & 11.00 & $\CL$ & 4.9 / \phantom{0}9\\
Lamberteschi & 1648 & 10.69 & $\CL$ & 4.7 / \phantom{0}8\\
Ginori & 1604 & 10.65 & $\CL$ & 4.6 / \phantom{0}7\\
Pazzi & 1370 & 10.42 & $\CL$ & 4.6 / \phantom{0}8\\
\bottomrule
\end{tabular}
\caption{Algorithm~\ref{alg:vsp} on the Florentine network with each family as seed. The content
count and $\CL$ track the seed's centrality; across all seeds the code length equals $\CL$ exactly
and the mistakes stay below it.}
\label{tab:seed}
\end{table}

\subsection{The WordNet taxonomy at scale}\label{app:wordnet}

Section~\ref{sec:exp} enumerates $\Om$ outright, which is feasible only for small instances;
the complexity of computing $|\Om|$ for general templates is open
(Section~\ref{sec:position}). On a \emph{tree}, however, the version space factorizes
and Algorithm~\ref{alg:vsp} runs \emph{implicitly}, with no enumeration: an observed label
propagates (a concept marked forces its ancestors marked; a concept unmarked forces its whole
subtree unmarked), and the number of consistent contents together with the mixture weight
$\hat p_t$ of the next query are computed by a single $O(n)$ dynamic program over the tree. The
mixture over $|\Om|$ meanings is thus evaluated at every step, however large $|\Om|$ is.

\paragraph{Formal setup (WordNet).}
Fix a category and let its descendants in the WordNet \citep{miller1995wordnet} is-a (hypernymy) taxonomy
form a rooted tree $\mathcal{T}=(V,\text{parent})$ with root $r$ and $|V|=n$ concepts
($n=366$ for \textsc{carnivore}). The constants are the concepts; there are no function
symbols, so the Herbrand universe is $V$ itself. A single unary predicate
$\mathsf{under}$ carries the content: $\mathsf{under}(c)$ reads ``concept $c$ is included
in the selected sub-taxonomy.'' The content sub-base --- the atoms the learner is queried on --- is
\[
\mathcal{B}=\{\mathsf{under}(c):c\in V\},\qquad |\mathcal{B}|=n,
\]
the $\mathrm{parent}$ atoms being fixed background. The parent relation is fixed structural background
(the analogue of $\mathsf{tie}$ in the Florentine case) and is never predicted; the
background $BK$ additionally declares the root, $\mathsf{under}(r)$, as the seed
($V_0=\{r\}$).

\paragraph{Template and hypotheses.}
The reachability schema $\mathsf{under}(v)\leftarrow\mathsf{under}(\mathrm{parent}(v))$
grounds to one edge clause per non-root concept, so
\[
L=\bigl\{\,\mathsf{under}(v)\leftarrow\mathsf{under}(u):v\in V\setminus\{r\},\ u=\mathrm{parent}(v)\,\bigr\},
\qquad N=|L|=n-1,
\]
a plain edge clause $r_v\leftarrow r_u$ in the form of Proposition~\ref{prop:realize} (here $N=365$, so
there are $2^{365}\approx10^{110}$ clause-sets). A hypothesis $S\subseteq L$ switches on a
set of parent$\to$child edges; its meaning is
$\mu(S)=\mathrm{lhm}(BK\cup S)\subseteq\mathcal{B}$, obtained by firing the switched-on
edges from the root to a fixpoint.

\paragraph{Realizable contents.}
By Proposition~\ref{prop:realize}, on a tree each node's only in-neighbour is its parent, so
$R\in\Omega_L$ iff $r\in R$ and $R$ is ancestor-closed---a concept is included only if its
parent is:
\[
\Omega_L=\{R\subseteq V:\ r\in R,\ \mathrm{parent}(v)\in R\ \text{for all }v\in R\setminus\{r\}\},
\]
i.e.\ the subtrees of $\mathcal{T}$ hanging from the root (the is-a reading of the
reachability family; the Florentine instance is its social-network reading). The count
factorizes: writing $f(v)$ for the number of valid configurations of the subtree at $v$
given $v\in R$,
\[
f(v)=\prod_{c:\ \mathrm{parent}(c)=v}\bigl(1+f(c)\bigr),\qquad
|\Omega_L|=f(r),
\]
each child being dropped or kept-with-$f(c)$-options. For \textsc{carnivore} this gives
$|\Omega_L|=10^{90.9}$, hence the semantic capacity
$C(L)=\log_2|\Omega_L|=302.1$ bits, against the $2^{365}$ clause-sets---collapse ratio
$2^N/|\Omega_L|\approx10^{18.9}$. The product above is the $O(n)$ dynamic program by which
Algorithm~\ref{alg:vsp} evaluates the version-space mixture without ever enumerating $\Omega_L$.

\begin{figure}[t]
\centering
\includegraphics[width=0.9\linewidth]{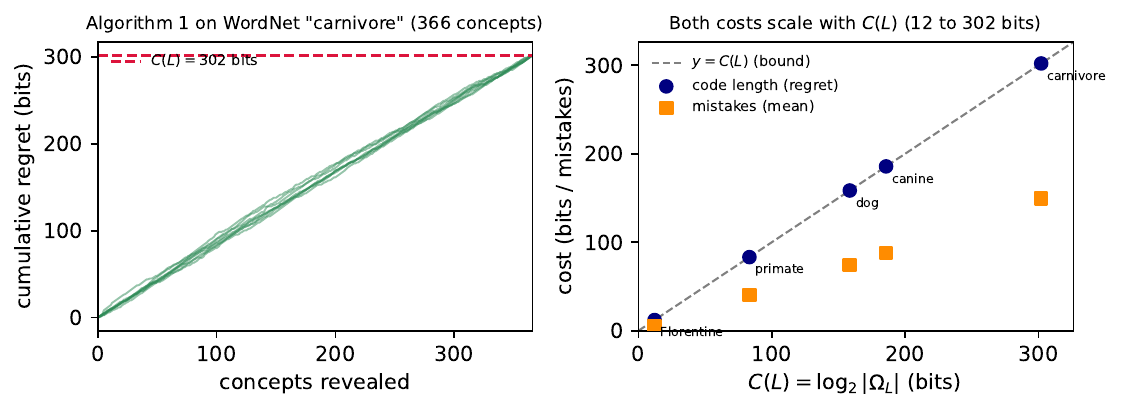}
\caption{\textbf{Left:} Algorithm~\ref{alg:vsp} on the \textsf{carnivore} taxonomy ($366$ concepts,
$|\Om|=10^{90.9}$); the cumulative regret climbs to $\CL=302$ bits. \textbf{Right:} across the five
instances of Table~\ref{tab:wordnet}, the code length lands exactly on the line $y=\CL$ from $12$
to $302$ bits, while the mistakes grow with $\CL$ but stay below it.}
\label{fig:wordnet}
\end{figure}

\paragraph{Results.} Table~\ref{tab:wordnet} runs Algorithm~\ref{alg:vsp} on categories spanning
$83$ to $302$ bits of semantic information, i.e.\ content classes from $10^{25}$ to $10^{90}$
meanings. In every case the cumulative code length equals $\CL$ \emph{exactly} (to $13$ decimal
places; the residual is floating-point), and the mistakes stay below $\CL$, at mean $\approx\CL/2$.
For \textsf{carnivore} ($366$ concepts) the $2^{365}\!\approx\!10^{110}$ clause-sets collapse onto
$|\Om|=10^{90.9}$ contents (collapse ratio $10^{18.9}$), and Algorithm~\ref{alg:vsp} pays exactly
$\CL=302$ bits without ever materializing a content (Figure~\ref{fig:wordnet}). The single number
$\CL$ therefore governs the online cost on real data across more than an order of magnitude of
scale --- and, since the tree case runs in linear time, whatever the complexity of counting
$\Om$ in general, it is no obstacle to \emph{using} it on these templates.

\begin{table}[t]\centering\small
\begin{tabular}{@{}lrrrcc@{}}
\toprule
category (root) & concepts & $|\Om|$ & $\CL$ (bits) & code length & mistakes (mean / max)\\
\midrule
Florentine (reference) & 15 & $4.1\times10^{3}$ & 12.0 & $\CL$ & 5.3 / \phantom{0}10\\
\textsf{primate} & 106 & $10^{25.0}$ & 83.2 & $\CL$ & 40.4 / \phantom{0}46\\
\textsf{dog} & 190 & $10^{47.7}$ & 158.5 & $\CL$ & 74.4 / \phantom{0}81\\
\textsf{canine} & 224 & $10^{55.9}$ & 185.7 & $\CL$ & 87.8 / \phantom{0}99\\
\textsf{carnivore} & 366 & $10^{90.9}$ & 302.1 & $\CL$ & 149.8 / 162\\
\bottomrule
\end{tabular}
\caption{Algorithm~\ref{alg:vsp} on the WordNet is-a taxonomy, run implicitly by dynamic
programming. The code length equals $\CL$ exactly at every scale, even for $|\Om|=10^{90}$;
mistakes stay under $\CL$.}
\label{tab:wordnet}
\end{table}

\subsection{Baseline: the same learner run over clause-sets}\label{app:baseline}

What does a learner lose by placing its prior on the syntax $2^L$ instead of the contents $\Om$?
The natural baseline is the same Bayes mixture as Algorithm~\ref{alg:vsp}, run over the $2^N$
clause-sets under the uniform prior. A clause-set predicts only through its content $\mu(S)$, so
this learner \emph{is} Algorithm~\ref{alg:vsp} with the non-uniform prior $w(R)=|\mu^{-1}(R)|$,
the fibre size, and Theorem~\ref{thm:vsp}(i) gives its regret exactly: once the queries separate
the contents it pays
\[
\reg^{\mathrm{syn}}_T(R^\star)=\log_2\frac{2^N}{|\mu^{-1}(R^\star)|}
\qquad\text{against}\qquad
\reg^{\mathrm{sem}}_T(R^\star)=\log_2|\Om|=\CL .
\]
The two learners therefore differ in one thing only, uniform on syntax versus uniform on
meanings, and the comparison is exact, with no tuning. Averaged over contents the clause-set
learner pays $\CL+D(\mathrm{unif}_{\Om}\,\|\,\pi)$ with $\pi=\mu_*\mathrm{unif}_{2^L}$
(Proposition~\ref{prop:lift}); averaged over its own prior it pays $H(\pi)\le\CL$.
In summary:

\begin{corollary}[Clause-level priors]
Let $P$ be a prior on $2^L$ and run the Bayes mixture over clause-sets.
It equals Algorithm~1 with $w=\pi:=\mu_*P$, so once the queries separate
the contents its regret on target $R^\star$ is $-\log_2 P(\mu^{-1}(R^\star))$;
for uniform $P$ this is $N-\log_2|\mu^{-1}(R^\star)|$. Its worst case over
targets is at least $C(L)$, with equality iff $\pi$ is uniform on $\Om$;
averaged over contents it pays $C(L)+D(\mathrm{unif}_{\Om}\,\|\,\pi)$, and
averaged over its own prior $H(\pi)\le C(L)$.
\end{corollary}

\begin{figure}[t]
\centering
\includegraphics[width=0.95\linewidth]{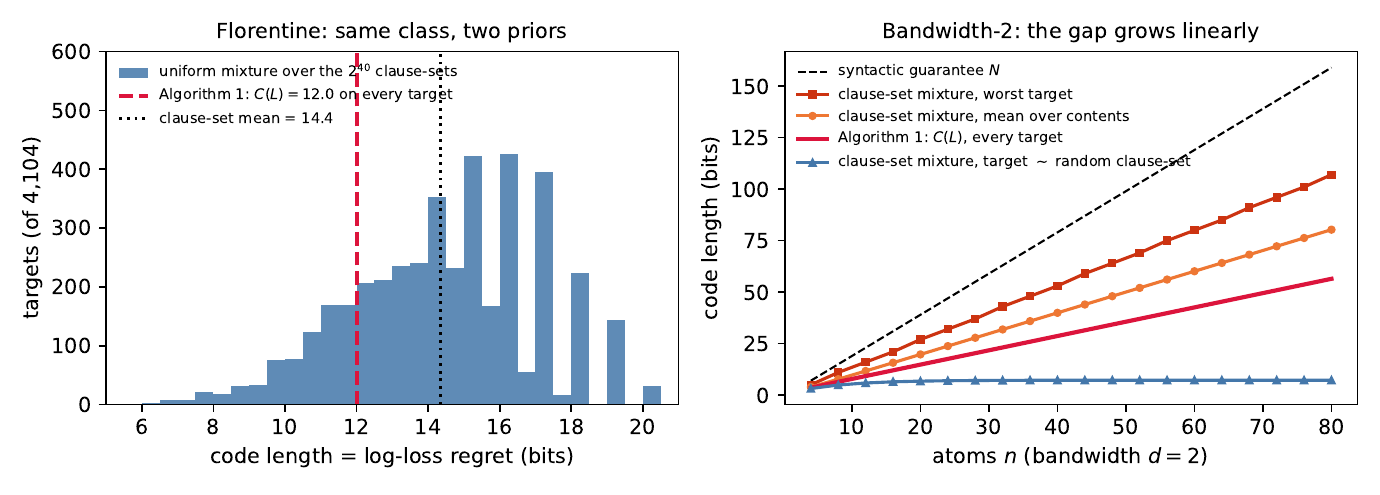}
\caption{\textbf{Left:} Florentine network. Code length of the uniform mixture over the $2^{40}$
clause-sets on each of the $4{,}104$ targets (histogram; mean $14.35$, max $20$), against the
constant $\CL=12.0$ of Algorithm~\ref{alg:vsp}. \textbf{Right:} bandwidth-$2$ template. The
clause-set mixture's worst-case and content-averaged code lengths grow linearly in $n$, above
$\CL$ and below the syntactic guarantee $N$; on targets drawn from random clause-sets it pays only
$H(\pi)=O(1)$.}
\label{fig:baseline}
\end{figure}

\textbf{Florentine network.} We compute all $4{,}104$ fibre sizes exactly (they sum to $2^{40}$).
Algorithm~\ref{alg:vsp} pays $\CL=12.00$ bits on every target. The clause-set mixture pays
between $6.00$ and $20.00$ bits: more than $\CL$ on $79\%$ of the contents, $14.35$ bits on average
over contents ($D(\mathrm{unif}\,\|\,\pi)=2.35$ bits), and $8$ bits ($67\%$) more in the worst case
(Figure~\ref{fig:baseline}, left). Under $0/1$ loss the corresponding weighted-majority learner
makes $5.58$ mistakes on average and up to $12$, against $5.29$ and $10$ for Halving over $\Om$.
The ordering reverses when the target is itself the content of a uniformly random clause-set:
those targets have large fibres, and the clause-set mixture pays $H(\pi)=9.81$ bits ($3.75$
mistakes) against $12.00$ ($4.84$).

\textbf{Bandwidth template.} For $L_{n,d}$ the fibre mass factorizes along the line,
$-\log_2\pi(R)=\sum_{j=1}^nc_j$ with $c_j=k_j$ if $r_j\notin R$ and $c_j=-\log_2(1-2^{-k_j})$ if
$r_j\in R$, where $k_j$ is the number of atoms of $R$ among $r_{j-d},\dots,r_{j-1}$; a transfer pass
gives the worst case and the mean exactly for any $n$ (Figure~\ref{fig:baseline}, right;
Table~\ref{tab:baseline}). The overpayment is linear in $n$: for $d=2$ the clause-set mixture pays
$1.01\,n$ bits on average over contents and $\tfrac43n$ in the worst case, against
$\CL=0.694\,n$. It stays well below the syntactic guarantee $N\approx dn$: the collapse ratio
$2^N/|\Om|$ is the gap between the two guarantees, of which the realized regret gap is a constant
fraction (about one quarter on average and one half in the worst case for $d=2$). On typical
random programs the clause-set prior is again the better code: by Corollary~\ref{cor:subcritical}
random reachability dies out, so $H(\pi)=O(1)$ ($7.25$ bits for $d=2$, uniformly in $n$) while
$\CL=\Theta(n)$.

\textbf{Reading.} Uniform-on-syntax is a strongly non-uniform prior on meanings: it concentrates
on the few contents with huge fibres (small outbreaks, or fully saturated ones) and starves the
exponentially many thin-fibre contents. The content-uniform Algorithm~\ref{alg:vsp} is the minimax
choice (Theorem~\ref{thm:vsp}(ii)): it pays exactly $\CL$ on every target, whereas the clause-set
learner's cost ranges over $[\,O(1),\Theta(N)\,]$ depending on how syntactically redundant the
truth happens to be. Section~\ref{sec:exp} shows the same pattern for Hedge under label noise.

\begin{table}[t]\centering\small
\begin{tabular}{@{}lcccc@{}}
\toprule
& $\CL$ (Alg.~\ref{alg:vsp}, & clause-set, & clause-set, & guarantee\\
& every target) & mean over $\Om$ & worst & $N$\\
\midrule
Florentine ($N=40$) & 12.00 & 14.35 & 20.00 & 40\\
bandwidth $d=2$, $n=20$ & 14.81 & 19.76 & 27.00 & 39\\
bandwidth $d=2$, per atom & 0.694 & 1.009 & 1.333 & 2\\
bandwidth $d=3$, per atom & 0.879 & 1.129 & 1.500 & 3\\
\bottomrule
\end{tabular}
\caption{Log-loss regret (bits) of the uniform mixture over clause-sets versus
Algorithm~\ref{alg:vsp}. All entries are exact (fibre counts / transfer pass); per-atom rows are
the $n\to\infty$ rates.}
\label{tab:baseline}
\end{table}

\subsection{Complete results for Section~\ref{sec:exp}}\label{app:newexp}

\textbf{Setup.} Florentine: seed Medici, $N=40$ clauses, all $4{,}104$ contents enumerated with
exact fibre sizes (they sum to $2^{40}$). WordNet: breadth-first spanning tree of the hyponym
closure of each root, one clause per edge; $|\Om|$ reproduces Table~\ref{tab:wordnet} exactly
($83.2$, $158.5$, $185.7$, $302.1$ bits). Held-out atoms: on Florentine, $400$ random reveal sets
per $q$, exact over all targets; on WordNet, $40$ targets from each target distribution and $3$
reveal sets each. Label noise: each run draws a target $R^\star$ and then, in each of $T$ rounds, a query atom
uniformly at random (with replacement) from the non-seed atoms; the true label
$\mathbf 1[\gamma_t\in R^\star]$ is flipped with probability $\epsilon$, and only the flipped
label is shown. $T=300$ on Florentine ($300$ runs) and $T=3000$ on \textsf{carnivore} ($20$ runs;
$10$ for rule-generated targets at $\epsilon=0.1$). Algorithm~\ref{alg:mismatch} uses
$\eta=\sqrt{8\ln|\Om|/T}$ and its randomized prediction; we report its \emph{expected} number of
mistakes, computed exactly from its predictive probabilities. Algorithm~\ref{alg:vsp} predicts by
majority vote over the version space; once the version space is empty, its tie rule predicts
$1$. Regret subtracts the mistakes of the best content in hindsight, computed exactly (by
enumeration on Florentine, and by the $(\min,+)$ tree recursion of Appendix~\ref{app:ewa} on
\textsf{carnivore}). The best content errs on about $\epsilon T$ labels, so random guessing has
regret about $(\tfrac12-\epsilon)T$: $120$ and $1200$ at $\epsilon=0.1$. Clean error is the
fraction of the last $T/4$ rounds in which the prediction disagrees with the true (unflipped)
label. ``Clause-uniform'' targets are $R=\mu(S)$ with $S$ uniform over the $2^N$
clause-sets; ``content-uniform'' targets are uniform over $\Om$.

\textbf{Held-out atoms.} The learner sees the labels of a random fraction $q$ of the atoms and
predicts the rest from the posterior of a mixture over $\Om$ whose prior is content-uniform,
clause-uniform (the fibre size $|\mu^{-1}(R)|$) or a size prior $2^{-|R|}$. Now the prior matters
(Tables~\ref{tab:full-flor} and~\ref{tab:full-wn}), and each prior is best on targets drawn from itself. On Florentine,
evaluated exactly over all $4{,}104$ targets, the content-uniform prior has the smallest
worst-case held-out loss at every $q$ ($0.91$--$1.20$ bits/atom); the clause prior's worst case is
$1.30$--$1.76$ and \emph{grows} with $q$. On \textsf{carnivore}, the clause prior's bias toward
small subtrees is the more robust one over the targets we sampled. Neither prior dominates: the
parameterization changes generalization, even though VC and Littlestone dimension are blind to it.

\begin{table}[h]\centering\small
\begin{tabular}{@{}llccccc@{}}\toprule
& & \multicolumn{5}{c}{fraction revealed $q$}\\\cmidrule(l){3-7}
targets & prior & 0.0 & 0.2 & 0.4 & 0.6 & 0.8\\\midrule
content-unif. & content-unif. & 0.95 & 0.91 & 0.87 & 0.84 & 0.79\\
 & clause-unif. & 0.98 & 1.07 & 1.07 & 1.04 & 0.98\\
 & size $2^{-|R|}$ & 1.11 & 1.04 & 0.97 & 0.93 & 0.86\\
\addlinespace
clause-unif. & content-unif. & 0.98 & 0.93 & 0.87 & 0.82 & 0.76\\
 & clause-unif. & 0.95 & 0.78 & 0.68 & 0.64 & 0.58\\
 & size $2^{-|R|}$ & 1.04 & 0.94 & 0.88 & 0.86 & 0.85\\
\addlinespace
worst case & content-unif. & 1.20 & 1.15 & 0.94 & 0.91 & 0.97\\
 & clause-unif. & 1.30 & 1.45 & 1.50 & 1.61 & 1.76\\
 & size $2^{-|R|}$ & 1.70 & 1.45 & 1.38 & 1.37 & 1.36\\

\bottomrule\end{tabular}
\caption{Florentine, held-out log-loss (bits/atom), all $q$. ``Worst case'' is the maximum over
all $4{,}104$ targets of the loss averaged over reveal sets.}
\label{tab:full-flor}
\end{table}

\begin{table}[h]\centering\footnotesize\setlength{\tabcolsep}{3.2pt}
\begin{tabular}{@{}llccccccccc@{}}\toprule
& & \multicolumn{3}{c}{content-unif.\ targets} & \multicolumn{3}{c}{clause-unif.\ targets} & \multicolumn{3}{c}{max (sampled)}\\
\cmidrule(lr){3-5}\cmidrule(lr){6-8}\cmidrule(l){9-11}
tree & prior & .2 & .5 & .8 & .2 & .5 & .8 & .2 & .5 & .8\\\midrule
\textsc{primate} (106) & content-unif. & 0.82 & 0.79 & 0.78 & 1.28 & 0.64 & 0.33 & 2.24 & 1.37 & 1.00\\
 & clause-unif. & 1.07 & 0.86 & 0.81 & 0.36 & 0.26 & 0.23 & 1.79 & 1.11 & 1.19\\
 & size $2^{-|R|}$ & 0.97 & 0.90 & 0.83 & 0.66 & 0.40 & 0.28 & 1.26 & 1.23 & 1.31\\
\addlinespace
\textsc{dog} (190) & content-unif. & 0.86 & 0.84 & 0.80 & 1.28 & 0.70 & 0.38 & 2.03 & 1.73 & 0.99\\
 & clause-unif. & 1.00 & 0.90 & 0.84 & 0.37 & 0.31 & 0.28 & 1.68 & 1.14 & 1.16\\
 & size $2^{-|R|}$ & 0.99 & 0.92 & 0.86 & 0.78 & 0.48 & 0.35 & 1.15 & 1.12 & 1.14\\
\addlinespace
\textsc{canine} (224) & content-unif. & 0.86 & 0.83 & 0.79 & 1.25 & 0.53 & 0.30 & 2.09 & 1.53 & 1.06\\
 & clause-unif. & 1.00 & 0.89 & 0.84 & 0.27 & 0.22 & 0.20 & 1.28 & 1.12 & 1.11\\
 & size $2^{-|R|}$ & 0.98 & 0.92 & 0.87 & 0.77 & 0.41 & 0.25 & 1.24 & 1.10 & 1.12\\
\addlinespace
\textsc{carnivore} (366) & content-unif. & 0.86 & 0.82 & 0.79 & 1.28 & 0.50 & 0.21 & 2.18 & 1.49 & 0.88\\
 & clause-unif. & 1.00 & 0.89 & 0.83 & 0.21 & 0.16 & 0.14 & 1.38 & 1.11 & 1.03\\
 & size $2^{-|R|}$ & 0.97 & 0.92 & 0.89 & 0.72 & 0.30 & 0.18 & 1.18 & 1.05 & 1.13\\
\bottomrule\end{tabular}
\caption{WordNet, held-out log-loss (bits/atom) for $q\in\{0.2,0.5,0.8\}$ on four taxonomies
(number of concepts in parentheses). ``Max'' is over all $240$ sampled (target, reveal-set) pairs
of both target families.}
\label{tab:full-wn}
\end{table}

\begin{table}[h]\centering\footnotesize\setlength{\tabcolsep}{4pt}
\begin{tabular}{@{}llcccc@{}}\toprule
& & \multicolumn{4}{c}{flip rate $\epsilon$}\\\cmidrule(l){3-6}
data / targets & learner & 0 & 0.05 & 0.1 & 0.2\\\midrule
Florentine / contents & Alg.~1 (prune) & 5.2 / 0.0 & 116.6 / 45.1 & 109.9 / 46.8 & 84.1 / 47.5\\
 & Alg.~2, content-unif. & 20.7 / 0.1 & 21.1 / 0.2 & 21.6 / 0.4 & 22.8 / 2.0\\
 & Alg.~2, clause-unif. & 24.5 / 0.2 & 24.6 / 0.4 & 25.0 / 0.8 & 26.1 / 3.1\\
 & version space empties & 0\% & 100\% & 100\% & 100\%\\
Florentine / rules & Alg.~2, content-unif. & 20.7 / 0.1 & 21.1 / 0.2 & 21.6 / 0.4 & 22.7 / 1.9\\
 & Alg.~2, clause-unif. & 17.0 / 0.1 & 16.7 / 0.2 & 17.8 / 0.4 & 18.6 / 1.5\\
\addlinespace
\textsc{carnivore} / contents & Alg.~1 (prune) & 143 / 0.0 & 1214 / 45.2 & 1056 / 44.7 & 786 / 43.4\\
 & Alg.~2, content-unif. & 353 / 1.7 & 359 / 2.6 & 363 / 4.4 & 366 / 9.4\\
 & Alg.~2, clause-unif. & 390 / 1.8 & 396 / 2.8 & 401 / 4.6 & 404 / 9.7\\
 & version space empties & 0\% & 100\% & 100\% & 100\%\\
\textsc{carnivore} / rules & Alg.~2, content-unif. & -- & -- & 356 / 1.1 & --\\
 & Alg.~2, clause-unif. & -- & -- & 81 / 0.8 & --\\
\bottomrule\end{tabular}
\caption{Label noise, all $\epsilon$: expected $0/1$ regret / clean error (\%). Rule-generated
targets on \textsf{carnivore} were run at $\epsilon=0.1$ only. The bound
$\sqrt{(\ln2/2)\,\CL\,T}$ is $35.3$ (Florentine) and $560$ (\textsf{carnivore}).}
\label{tab:full-noise}
\end{table}

\begin{table}[h]\centering\footnotesize\setlength{\tabcolsep}{3pt}
\begin{tabular}{@{}c c ccc c ccc c@{}}\toprule
& & \multicolumn{3}{c}{lower bound, witness of the proof} & ${}+$ inside $G$ & \multicolumn{3}{c}{upper bound} & \\
\cmidrule(lr){3-5}\cmidrule(lr){6-6}\cmidrule(lr){7-9}
$c$ & $2\zeta^2(1-\zeta)$ & $m{=}500$ & 1000 & 2000 & 2000 & 500 & 1000 & 2000 & $\E|K|$ ($m{=}2000$)\\\midrule
0.25 & 0.0000 & 0.0000 & 0.0000 & 0.0000 & 0.0000 & 0.0000 & 0.0024 & 0.0000 & 0.0\\
0.50 & 0.0000 & 0.0003 & 0.0001 & 0.0000 & 0.0000 & 0.0047 & 0.0028 & 0.0007 & 0.4\\
0.75 & 0.0000 & 0.0016 & 0.0004 & 0.0001 & 0.0001 & 0.0220 & 0.0158 & 0.0118 & 10.2\\
1.00 & 0.0000 & 0.0095 & 0.0076 & 0.0102 & 0.0111 & -- & -- & -- & 118.1\\
1.25 & 0.1734 & 0.0970 & 0.1052 & 0.1045 & 0.1279 & -- & -- & -- & 964.0\\
1.50 & 0.2834 & 0.2009 & 0.2140 & 0.2204 & 0.3244 & -- & -- & -- & 2009.5\\
2.00 & 0.2580 & 0.2147 & 0.2174 & 0.2196 & 0.4951 & -- & -- & -- & 3687.1\\
2.50 & 0.1711 & 0.1459 & 0.1515 & 0.1487 & 0.5682 & -- & -- & -- & 4850.2\\
3.00 & 0.1053 & 0.0957 & 0.0940 & 0.0946 & 0.6057 & -- & -- & -- & 5931.8\\
\bottomrule\end{tabular}
\caption{Semantic compression transition, $F\sim\mathcal D(m,c/m)$: Monte-Carlo estimates of the
lower bound on the semantic redundancy $\E\bigl[\sum_{x\notin G}\log_2\binom{|G|}{a_x}+\sum_{y\notin G}\log_2\binom{|G|}{b_y}\bigr]/H(F)$
from the proof of Theorem~\ref{prop:collapse}(ii); of the same quantity plus
$\E[(e(G)-|G|)\log_2(1/p)]/H(F)$, which adds the edges inside $G$ (a valid lower bound for each $m$; see the
remark after the proof); and of the upper bound $(H(|K|)+\E|K|\log_2(m(m-1)))/H(F)$ from part (i)
(informative only for $c<1$). Lower bounds: $40$ replicates for $m\le1000$, $20$ for $m=2000$;
upper bounds: $16$ and $8$. The witness approaches $2\zeta^2(1-\zeta)$ from below; convergence is
slowest near $c=1$, where $\log_2|G|/\log_2m$ is still far from $1$.}
\label{tab:full-collapse}
\end{table}

\begin{table}[h]\centering\footnotesize\setlength{\tabcolsep}{4pt}
\begin{tabular}{@{}c ccc cc c@{}}\toprule
& \multicolumn{3}{c}{true value (MC, exact $P(R)$)} & \multicolumn{2}{c}{lower bounds, $m{=}2000$} & proved limit\\
\cmidrule(lr){2-4}\cmidrule(lr){5-6}
$c$ & $m{=}200$ & $m{=}500$ & $m{=}1000$ & boundary & ${}+$ inside $G$ & $2\zeta^2(1-\zeta)$\\\midrule
0.50 & 0.003 & 0.002 & 0.001 & 0.000 & 0.000 & 0.000\\
0.75 & 0.013 & 0.006 & 0.004 & 0.000 & 0.000 & 0.000\\
1.00 & 0.061 (0.006) & 0.042 & 0.020 & 0.010 & 0.011 & 0.000\\
1.25 & 0.238 (0.016) & 0.236 (0.012) & 0.254 (0.012) & 0.105 & 0.128 & 0.173\\
1.50 & 0.516 (0.016) & 0.543 (0.010) & 0.570 (0.008) & 0.220 & 0.324 & 0.283\\
2.00 & 0.834 (0.007) & 0.842 & 0.850 & 0.220 & 0.495 & 0.258\\
2.50 & 0.917 & 0.933 & 0.943 & 0.149 & 0.568 & 0.171\\
3.00 & 0.964 & 0.965 & 0.967 & 0.095 & 0.606 & 0.105\\
\bottomrule\end{tabular}
\caption{The true semantic redundancy $H(F\mid R)/H(F)=1-H(R)/H(F)$, $F\sim\mathcal D(m,c/m)$, estimated by
Monte Carlo with the exact $P(R)$ of Appendix~\ref{app:collapse} (standard errors in
parentheses when at least $0.005$), against the lower bounds of the proof at $m=2000$ and the proved
limit. Below $c=1$ the redundancy decreases with $m$; above $c=1$ it increases slightly with $m$.
The compression rate $H(R)/H(F)$ is one minus each entry.}
\label{tab:true-rate}
\end{table}

\begin{table}[h]\centering\small
\begin{tabular}{@{}cccccc@{}}\toprule
& \multicolumn{5}{c}{$m$}\\\cmidrule(l){2-6}
$c$ & 500 & 1000 & 2000 & 4000 & 8000\\\midrule
0.50 & $1.70\pm0.37$ & $1.38\pm0.31$ & $0.73\pm0.22$ & $1.80\pm0.41$ & $1.05\pm0.27$\\
0.75 & $7.80\pm1.38$ & $7.58\pm1.03$ & $7.60\pm1.41$ & $7.45\pm0.92$ & $7.28\pm1.07$\\
0.90 & $20.35\pm2.13$ & $19.30\pm2.02$ & $23.73\pm2.90$ & $27.53\pm3.41$ & $28.77\pm2.81$\\
\bottomrule\end{tabular}
\caption{Subcritical check of $\E|K|$, the number of edges of $F$ not determined by $R$ (proof of
Theorem~\ref{prop:collapse}(i)); $60$ replicates, $\pm$ one standard error. It is flat in $m$
for $c\le0.75$; at $c=0.9$ it is still approaching its limit, as expected near criticality.}
\label{tab:full-ek}
\end{table}

\subsection{A differentiable-ILP-style learner}\label{app:dilp}

Section~\ref{sec:related} notes that differentiable ILP optimizes a soft version of the $2^L$
template. This appendix runs such a learner on the same data as Hedge over $\Om$ and asks two
questions: does its \emph{prediction} match a learner over contents, and does the \emph{program}
it outputs?

\textbf{The learner.} In the spirit of \citet{evans2018learning}, every clause $\gamma\in L$ gets a
real weight $w_\gamma$ and a soft truth value $\sigma(w_\gamma)$. Forward chaining is made soft by the
probabilistic-OR amalgamation: on Florentine, the truth value of $\mathrm{reach}(v)$ is updated as
$x_v\leftarrow1-\prod_{\gamma=(u\to v)}\bigl(1-\sigma(w_\gamma)\,x_u\bigr)$, with the seed fixed to $1$, and the
update is iterated $n$ times; on \textsf{carnivore} the tree gives the closed form
$x_v=\prod_{\gamma\text{ on the path to }v}\sigma(w_\gamma)$. The weights start from $\mathcal N(0,0.5^2)$
and are trained by Adam (learning rate $0.1$, $400$ steps) on the binary cross-entropy of the
labelled examples. This is a ``dILP-style'' learner: it keeps the parameterization and the soft
semantics of differentiable ILP but not the full system (rule templates over predicates,
the amalgamation schedule, or the search over program templates).

\textbf{Protocol.} Batch data, identical for every learner: a target $R^\star$ is drawn (uniformly from
$\Om$, ``contents'', or as the content of a uniformly random clause-set, ``rules''); $T$ atoms are
queried uniformly from the non-seed atoms and each label is flipped with probability $\epsilon$. We
report the error of each output against the \emph{true} target over all non-seed atoms ($40$ runs on
Florentine, $10$ on \textsf{carnivore}). The learners are: the dILP-style learner (predict an atom
iff its soft truth exceeds $\tfrac12$); Hedge over $\Om$ with the content-uniform prior (ours) and
with the clause-uniform prior (posterior weight $w(R)\,\beta^{\#\mathrm{mistakes}(R)}$, prediction
by weighted majority).

\begin{figure}[h]\centering
\includegraphics[width=0.95\linewidth]{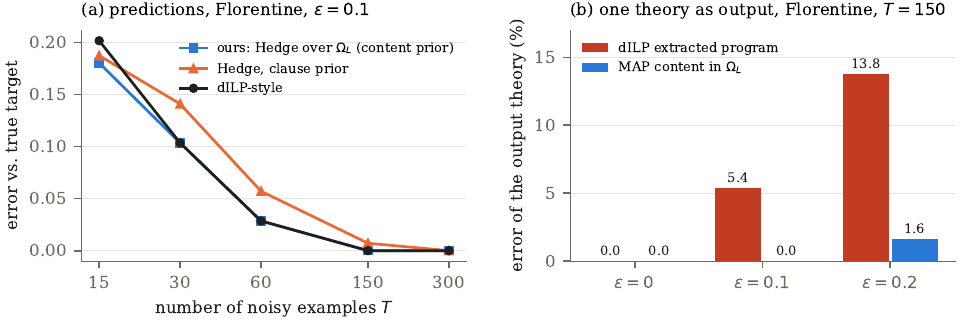}
\caption{(a) Prediction error against the true target on Florentine, $\epsilon=0.1$, content-generated
targets: the dILP-style learner tracks Hedge over $\Om$ with the content prior. (b) Error of the
single output theory at $T=150$: the program extracted from the dILP-style weights vs.\ the MAP
content in $\Om$.}
\label{fig:dilp}
\end{figure}

\textbf{Predictions.} The dILP-style learner predicts essentially as well as Hedge over $\Om$
(Figure~\ref{fig:dilp}(a), Table~\ref{tab:dilp-flor}). On content-generated targets it matches the
content prior (e.g.\ $10.4\%$ vs.\ $10.4\%$ at $T=30$ and $2.9\%$ vs.\ $2.9\%$ at $T=60$,
$\epsilon=0.1$); on rule-generated targets it lies between the two priors. The same holds on
\textsf{carnivore} ($7.1\%$ vs.\ $7.1\%$ at $\epsilon=0.1$, $T=1000$). This is expected:
gradient descent on a product-form likelihood finds the same predictions as a mixture over the
meanings that the likelihood supports.

\textbf{The output program.} The difference appears when the learner must output \emph{one}
theory, as ILP systems do. Thresholding the weights ($\gamma\in S$ iff $\sigma(w_\gamma)>\tfrac12$)
and deducing exactly gives the \emph{extracted program}; the learner over contents outputs the
\emph{MAP content}, the content with fewest mistakes on the data. Table~\ref{tab:dilp-extract} and
Figure~\ref{fig:dilp}(b) compare them. On Florentine under noise the extracted program is much
worse: $5.4\%$ vs.\ $0.0\%$ at $\epsilon=0.1$ and $13.8\%$ vs.\ $1.6\%$ at $\epsilon=0.2$ ($T=150$),
and it does not improve with more data ($5.0\%$ vs.\ $0.0\%$ at $\epsilon=0.2$, $T=300$), although
the soft predictions are perfect there. Almost all of its errors are false negatives ($84$ false
negatives against $3$ false positives over the $80$ runs at $T=150$): the soft model supports an
atom through several clause-sets with the same meaning, each carrying only part of the evidence,
and in a third of these atoms two or more parallel rules into the atom have
$0.15<\sigma(w_\gamma)<\tfrac12$; thresholding each clause separately removes them all. On
\textsf{carnivore}, a tree with exactly one rule into each atom, every content has a single
clause-set, and the extracted program matches the MAP content to within $0.6$ points. The gap thus
appears exactly where the fibres $|\mu^{-1}(R)|$ are large.

\textbf{Redundancy of the parameters.} On Florentine, for $20$ content-uniform targets, we trained
$30$ restarts each from $w\sim\mathcal N(0,1.5^2)$ on clean labels of all atoms. The meaning is
identified in $99.8\%$ of the $600$ restarts, but the $30$ restarts reach on average $29$
\emph{distinct} clause-sets, consistent with fibres of $2\times10^6$ to $2\times10^9$ clause-sets
per target (median $4.2\times10^7$). The parameters of a soft clause template are therefore not
identifiable, while the content is; this is the semantic compression of
Section~\ref{sec:phase} seen from the optimizer's side.

\textbf{Reading.} None of this contradicts differentiable ILP, which was designed for other goals
(learning programs over predicates from few examples). The point is narrower: its
predictions are those of a learner over $\Om$, while its single extracted program pays for the
redundancy of the parameterization; computing over contents (here, taking the MAP content) removes
that cost. Scripts: \texttt{dilp.py}, \texttt{dilp\_extract.py}, \texttt{dilp\_mechanism.py},
\texttt{dilp\_redundancy.py}.

\begin{table}[h]\centering\footnotesize
\begin{tabular}{@{}lcccccc@{}}\toprule
& & \multicolumn{5}{c}{number of noisy examples $T$}\\\cmidrule(l){3-7}
targets, $\epsilon$ & learner & 15 & 30 & 60 & 150 & 300\\\midrule
contents, 0.1 & dILP-style & 20.2 & 10.4 & 2.9 & 0.0 & 0.0\\
 & ours (Hedge, content) & 18.0 & 10.4 & 2.9 & 0.0 & 0.0\\
 & Hedge, clause & 18.7 & 14.1 & 5.7 & 0.7 & 0.0\\
\addlinespace
contents, 0.2 & dILP-style & 24.8 & 17.3 & 9.5 & 0.9 & 0.0\\
 & ours (Hedge, content) & 23.2 & 15.2 & 9.8 & 0.7 & 0.0\\
 & Hedge, clause & 25.9 & 16.6 & 11.8 & 1.4 & 0.4\\
\addlinespace
rules, 0.1 & dILP-style & 18.0 & 9.8 & 3.4 & 0.2 & 0.0\\
 & ours (Hedge, content) & 15.2 & 9.8 & 3.2 & 0.2 & 0.0\\
 & Hedge, clause & 14.6 & 8.4 & 2.9 & 0.9 & 0.0\\
\addlinespace
rules, 0.2 & dILP-style & 23.8 & 15.2 & 7.7 & 0.4 & 0.4\\
 & ours (Hedge, content) & 23.8 & 15.9 & 8.0 & 0.5 & 0.2\\
 & Hedge, clause & 21.8 & 12.1 & 6.1 & 0.9 & 0.5\\
\bottomrule\end{tabular}
\caption{Florentine, error (\%) of the predictions against the true target, for content- and
rule-generated targets and $\epsilon\in\{0.1,0.2\}$; $40$ runs per cell.}
\label{tab:dilp-flor}
\end{table}

\begin{table}[h]\centering\footnotesize
\begin{tabular}{@{}llccc@{}}\toprule
data & $\epsilon$ / $T$ & dILP (soft) & dILP program & MAP content\\\midrule
Florentine & 0.0 / 150 & 0.0 & 0.0 & 0.0\\
Florentine & 0.1 / 150 & 0.0 & 5.4 & 0.0\\
Florentine & 0.2 / 150 & 1.6 & 13.8 & 1.6\\
Florentine & 0.1 / 300 & 0.0 & 5.2 & 0.0\\
Florentine & 0.2 / 300 & 0.0 & 5.0 & 0.0\\
\textsf{carnivore} & 0.1 / 1000 & 6.6 & 7.0 & 6.8\\
\textsf{carnivore} & 0.2 / 1000 & 13.3 & 13.3 & 13.1\\
\textsf{carnivore} & 0.1 / 3000 & 0.5 & 0.7 & 0.6\\
\textsf{carnivore} & 0.2 / 3000 & 3.2 & 3.8 & 3.0\\
\bottomrule\end{tabular}
\caption{Error (\%) of one output theory: soft dILP-style predictions, the extracted program
(threshold at $\tfrac12$, exact deduction), and the MAP content in $\Om$. Florentine: $40$ runs;
\textsf{carnivore}: $10$ runs.}
\label{tab:dilp-extract}
\end{table}
\FloatBarrier

\vspace{-0.1in}
\section{Semantic Information: the deterministic-channel lift}\label{sec:info}

The quantities of Section~\ref{sec:object} are combinatorial including a count $|R|$ and a
Hartley log $\log_2|\Om|$. They are the deterministic shadows of information-theoretic
objects that appear the moment a prior is placed on hypotheses. Because the closure map
$\mu$ is deterministic, the lift is exact and forced. Let $P$ be a law on $2^L$; through
the deterministic channel $S\mapsto R=\mu(S)$ it induces a random content with pushforward
$\pi=\mu_*P$ on $\Om$. That is, $\pi=\mu_*P$ is the distribution of the content $R$ (on $\Om$) induced by pushing the
clause-set prior $P$ (on $2^L$) through $\mu$. $S$ carries the input prior; $R$ carries
$\pi$.

\begin{definition}[Probabilistic semantic quantities]\label{def:prob}
For a prior $P$ on $2^L$ with pushforward $\pi=\mu_*P$: the \emph{semantic entropy} is
$H(\Om):=H(R)$; the \emph{semantic information} is $I_P:=I(S;R)$ across the generation
channel; the \emph{semantic capacity} is $\CL:=\sup_P I(S;R)$.
\end{definition}

\begin{proposition}[The deterministic channel collapses the triple]\label{prop:lift}
For $R=\mu(S)$ and any prior $P$:
\begin{enumerate}[label=(\roman*)]
\item $I(S;R)=H(R)$ and $H(S)=H(R)+H(S\mid R)$; the semantic compression gain
$H(S\mid R)$ is the syntactic redundancy that $\mu$ \emph{compresses
away}, and it is $0$ iff $\mu$ is injective on the support of $P$;
\item $\CL=\sup_P I(S;R)=\log_2|\Om|$, attained exactly when $\pi$ is uniform on $\Om$;
this recovers the combinatorial capacity of Definition~\ref{def:cec};
\item under that capacity-achieving prior the self-information
$-\log_2\pi(R)\equiv\log_2|\Om|$ is constant --- an exact equipartition for realizable
theories --- while a generic prior falls short of capacity by
$\log_2|\Om|-H(R)=D(\pi\,\|\,\mathrm{unif}_{\Om})$.
\end{enumerate}
\end{proposition}

\begin{proof}[Proof sketch]
$R=\mu(S)$ is deterministic, so $H(R\mid S)=0$ and $I(S;R)=H(R)$; the chain rule gives the
decomposition. As $P$ ranges over laws on $2^L$, $\pi=\mu_*P$ ranges over all laws on
$\Om$, so $\sup_P H(R)=\log_2|\Om|$ at the uniform $\pi$; the deficit is the KL divergence
to uniform.
\end{proof}

This fixes the dictionary: semantic content, after taking logs, is entropy $H(R)$;
semantic information is $I(S;R)$, here equal to $H(R)$ because generation is noiseless; and
semantic capacity is the Shannon capacity $\sup_P I(S;R)$, here equal to $\log_2|\Om|$
because the channel is deterministic.

\begin{example}[The cheating class, made probabilistic]\label{ex:cheatprob}
In reality, not every friend of a cheated person is a cheater. This is the \emph{dual} of
the epidemic in Example~\ref{ex:epidemic}: there the friendship \emph{graph} was random and
the spreading rule fixed; here we keep the fixed ten-tie graph of Example~\ref{ex:main} and
randomize the \emph{rules} instead (``$S$ random'', a bond percolation on the fixed graph).
To model it, we
put a prior on the template of Example~\ref{ex:main} by letting each friendship conduct
cheating independently with probability $\theta$ (Bernoulli on clauses). The pushforward
$\pi_\theta$ gives each content the total mass of its fibre; the fibre sizes $|\mu^{-1}(R)|$
over the ten contents are $256,128,128,32,64,32,160,32,32,160$ (summing to $2^{10}$). At
$\theta=\tfrac12$ (uniform  $1024$ clause-sets),
\[
H(S)=N=10,\qquad H(R)=I(S;R)=2.962,\qquad H(S\mid R)=7.038\ \text{bits},
\]
so semantics compresses away $7.038$ of the $10$ hypothesis bits: they are syntactic redundancy,
removed by the quotient. The capacity is prior-free, $\CL=\log_2 10=3.322$ bits, attained only by the
content-uniform prior; the Bernoulli$(\tfrac12)$ prior falls $0.360$ bit short, exactly
$D(\pi_{1/2}\,\|\,\mathrm{unif}_{\Om})$, because its pushforward is uneven (fibres range
from $32$ to $256$). Equivalently, in the friendship-percolation reading (each friendship
conducts w.p.\ $\theta$), the middle student $C$ --- reachable through $A$ or through $D$ ---
cheats with the \emph{reliability} probability $1-(1-\theta^2)^2$, a redundancy no chain can
express. Capacity, entropy, and extent are three different bits; the mean extent is the
subject of Section~\ref{sec:phase}.
\end{example}

\section{A primer for readers from either side: machine learning \texorpdfstring{$\leftrightarrow$}{<->} logic}\label{app:dict}

This paper sits between two communities that use overlapping words for different things. A
reader trained in learning theory may never have met a least Herbrand model; a reader trained in
logic programming may never have met a Littlestone tree. This appendix is written so that either
reader can follow the main text. Section~\ref{app:dict-ml} explains the logic for learning
theorists, Section~\ref{app:dict-logic} explains the learning theory for logicians,
Section~\ref{app:dict-example} works one tiny example through every quantity in the paper,
Section~\ref{app:dict-table} collects the full dictionary, and Section~\ref{app:dict-false}
lists the words that mean different things on the two sides.

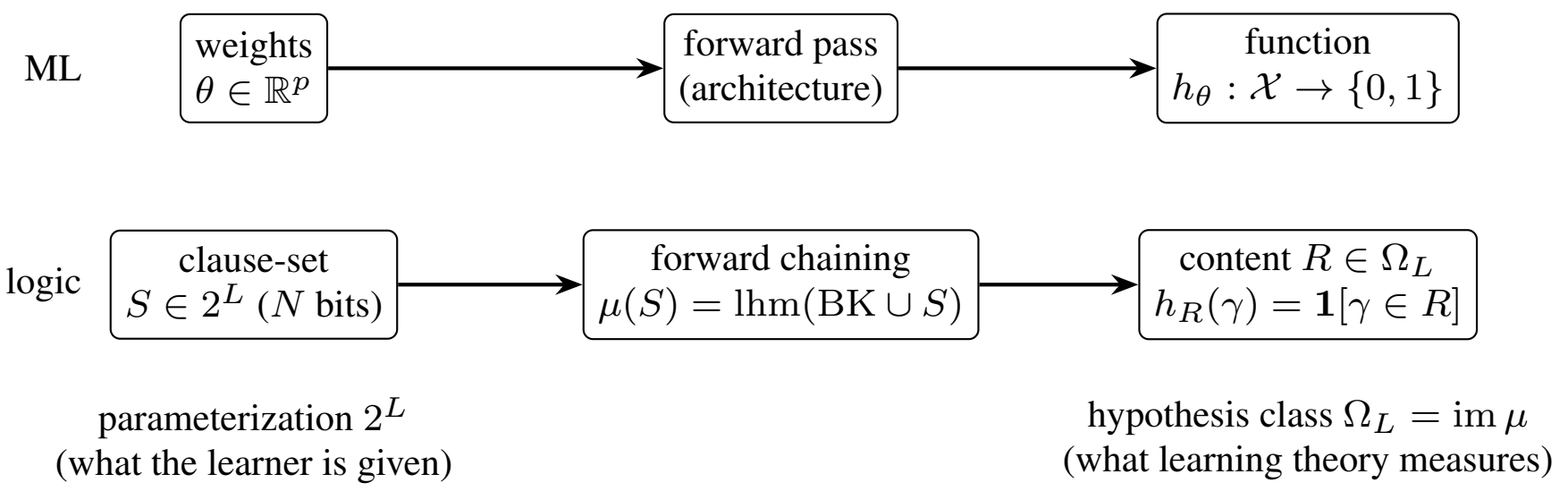
\begin{figure}[h]
\centering
\begin{tikzpicture}[
  box/.style={draw,rounded corners=2pt,minimum height=8mm,align=center,font=\footnotesize,inner sep=3pt},
  lab/.style={font=\footnotesize,align=center},
  arr/.style={-{Stealth[length=2mm]},thick}]
\node[box] (w)  at (0,1.6)   {weights\\$\theta\in\mathbb R^p$};
\node[box] (f)  at (3.9,1.6) {forward pass\\(architecture)};
\node[box] (h)  at (7.8,1.6) {function\\$h_\theta:\mathcal X\to\{0,1\}$};
\node[box] (s)  at (0,0)     {clause-set\\$S\in2^L$ ($N$ bits)};
\node[box] (mu) at (3.9,0)   {forward chaining\\$\mu(S)=\lhm(\BK\cup S)$};
\node[box] (r)  at (7.8,0)   {content $R\in\Om$\\$h_R(\gamma)=\mathbf1[\gamma\in R]$};
\draw[arr] (w)--(f); \draw[arr] (f)--(h);
\draw[arr] (s)--(mu); \draw[arr] (mu)--(r);
\node[lab,anchor=east] at (-1.15,1.6) {ML};
\node[lab,anchor=east] at (-1.15,0) {logic};
\node[lab] at (7.8,-1.15) {hypothesis class $\Om=\im\mu$\\(what learning theory measures)};
\node[lab] at (0,-1.15) {parameterization $2^L$\\(what the learner is given)};
\end{tikzpicture}
\caption{The same pipeline in two vocabularies. In both rows many parameter settings give the same
function (top: symmetric or ``dead'' weights; bottom: rules that never fire, or that re-derive
facts already deduced). Learning-theoretic quantities depend only on the right-hand column; the
learner is handed only the left-hand one.}
\label{fig:dict-pipeline}
\end{figure}

\subsection{Logic programming for learning theorists}\label{app:dict-ml}

\textbf{Atoms are the instances.} Fix finitely many predicate symbols (relations, such as
$\mathrm{friend}/2$ or $\mathrm{cheat}/1$) and finitely many constants (individuals, such as the
five students of Example~\ref{ex:main}). A \emph{ground atom} is a relation applied to constants,
such as $\cheat{A}$ or $\friendf{A}{C}$; it is a yes/no statement about the world. The
\emph{Herbrand base} $\Bb$ is the finite set of all ground atoms. In learning terms, $\Bb$ is the
instance space $\mathcal X$: a query is ``is $\gamma$ true?'' for one atom $\gamma$, and a label
is the answer. A set of atoms $R\subseteq\Bb$ is the same thing as a binary classifier on $\Bb$,
namely its indicator $h_R(\gamma)=\mathbf1[\gamma\in R]$. Readers who know databases may read
everything in this paper as \emph{Datalog}: function-free definite programs over finite
relations.

\textbf{Rules are switchable computations.} A \emph{definite (Horn) clause} is a rule
``$\mathit{head}\leftarrow\mathit{body}_1,\dots,\mathit{body}_k$'': if all body atoms hold, the
head holds. A rule with variables, such as $\cheat{Y}\leftarrow\cheat{X},\friendf{X}{Y}$, stands
for all its \emph{ground instances}, obtained by substituting constants for $X,Y$. A rule with an
empty body is simply a fact. Rules are the only operation; there is no negation, so adding rules
can only add conclusions (the semantics is \emph{monotone}).

\textbf{Forward chaining is the forward pass.} Given a program $P$ (a finite set of ground
clauses), the immediate-consequence operator $\TP$ fires, in parallel, every rule whose body
already holds. Starting from the empty set and iterating, $\emptyset\subseteq\TP(\emptyset)\subseteq
\TP^2(\emptyset)\subseteq\cdots$, one reaches in at most $|\Bb|$ rounds a fixpoint, the
\emph{least Herbrand model} $\lhm(P)$: exactly the atoms deducible from $P$. For the
reachability templates of this paper, forward chaining is literally breadth-first search: an atom
becomes true when a switched-on edge reaches it from an atom that is already true. Three classical
semantics (the least model, the least fixpoint of $\TP$, and the set of provable atoms) coincide
for definite programs \citep{vanemden1976semantics}, so ``the meaning of $P$'' is unambiguous: it
is the set $\lhm(P)$.

\textbf{Template, background, and hypothesis.} Inductive logic programming (ILP) fixes
\emph{background knowledge} $\BK$ (facts and rules that are always present, such as the friendship
graph) and a \emph{language bias}: a finite template $L$ of candidate clauses. A hypothesis is a
subset $S\subseteq L$, i.e.\ a binary vector of $N=|L|$ switches. So $L$ plays the role of an
architecture, $S$ the role of a binary weight vector, and the map $\mu(S)=\lhm(\BK\cup S)$ the role
of the forward pass that turns weights into a function (Figure~\ref{fig:dict-pipeline}).
Differentiable ILP \citep{evans2018learning} makes this literal by relaxing the switches to
continuous weights.

\textbf{Why $\mu$ is many-to-one.} As in a neural network, where permuting hidden units or
zeroing the outgoing weight of a dead unit leaves the function unchanged, many clause-sets have
the same meaning. A rule may never fire, because its body is never derived; or it may re-derive a
fact that another rule already derives. In Example~\ref{ex:main}, switching on the reverse edge
$\cheat{B}\leftarrow\cheat{A},\friendf{A}{B}$ changes nothing, because $B$ is already true. The
content class $\Om=\im\mu$ is the quotient of the parameter space by this redundancy, and it is
the only thing a learner scored on its predictions can ever identify.

\textbf{What ``semantic'' means here.} \emph{Semantics} is the rule that assigns meanings to
programs: here the map $\mu(S)=\lhm(\BK\cup S)$, fixed by the background knowledge (for transitive
closure, ``reachability''). The \emph{meaning} of a particular program is what this rule assigns
to it, $R=\mu(S)$: the set of facts $S$ implies, as opposed to how $S$ is written (its syntax).
Programs with the same meaning are synonyms. The ``semantic
information'' $\CL=\log_2|\Om|$ counts, in bits, how many different meanings the template can
express. The terminology follows \citet{carnap1952semantic}, who first proposed measuring the
information in what a statement says rather than in the symbols that encode it.

\subsection{Learning theory for logicians}\label{app:dict-logic}

\textbf{The protocol.} We use \emph{online} learning, which needs no probabilistic assumptions. At
each round $t$ an adversary (or nature) picks a query atom $\gamma_t$; the learner predicts whether
$\gamma_t$ is entailed by the unknown target theory; then the true bit $y_t$ is revealed. The
learner is judged over a whole sequence of $T$ rounds. This matches the ILP setting of learning
from positive and negative examples, presented one at a time.

\textbf{Realizable versus agnostic.} The problem is \emph{realizable} if the labels are those of
some content $R^\star\in\Om$, i.e.\ some program in the language bias explains the data exactly;
this is ILP without noise. It is \emph{agnostic} if the labels are arbitrary, as with noisy
examples. In the agnostic case one cannot hope for few mistakes, so one measures the
\emph{regret}: the learner's mistakes minus those of the best content in hindsight.

\textbf{Version space and Halving.} The \emph{version space} is the set of contents consistent
with all labels seen so far, the analogue of the set of programs that cover all positive and no
negative examples. The \emph{Halving} algorithm predicts by majority vote over the version space.
Each mistake removes at least half of it, so Halving makes at most $\log_2|\Om|=\CL$ mistakes in
the realizable case. This is the oldest bound in online learning, and it is why $\log_2$ of the
number of \emph{distinct} hypotheses matters.

\textbf{VC and Littlestone dimension.} A set of atoms is \emph{shattered} if the template can
realize every true/false pattern on it. The VC dimension is the size of the largest shattered set;
in logic terms, it is the largest number of atoms whose truth values the rules can set completely
independently. The Littlestone dimension $\Ldim$ is the adversarial version: the depth of the
deepest questioning strategy (a binary ``mistake tree'') in which every path forces the learner to
err at every step while some content stays consistent. Always
$\VC\le\Ldim\le\log_2|\Om|$. In the realizable case $\Ldim$ is exactly the optimal worst-case
number of mistakes \citep{littlestone1988learning}, and in the agnostic case the optimal regret is
$\Theta(\sqrt{\Ldim\,T})$ \citep{bendavid2009agnostic,alon2021adversarial}.

\textbf{Log loss, coding, and mixtures.} Instead of a yes/no guess the learner may output a
probability $\hat p_t$ that $\gamma_t$ holds, and pay $-\log_2$ of the probability it gave to the
truth. Summed over rounds this is the length, in bits, of a code for the label sequence, so online
log-loss learning \emph{is} data compression. The best achievable worst-case regret is the
logarithm of the \emph{Shtarkov sum} \citep{shtarkov1987universal}; for a finite class of
deterministic predictors it is the log of the number of distinct label patterns. The \emph{Bayes
mixture} with a prior $w$ over contents predicts with the posterior-weighted vote, and its regret
against $R^\star$ is exactly $\log_2(\text{total prior mass}/w(R^\star))$ once the queries
separate the contents (Theorem~\ref{thm:vsp}). This is where the prior matters: a uniform prior
over contents pays $\CL$ on every target, while a prior induced from clause-sets pays more on some
targets and less on others.

\textbf{Exponential weights (Hedge).} For the agnostic $0/1$ loss, Halving is too brittle: one
noisy label can eliminate the true content. Hedge keeps every content but multiplies its weight by
$\beta<1$ at each mistake (Algorithm~\ref{alg:mismatch}). Its regret is at most
$\sqrt{(\ln2/2)\,\CL\,T}$.

\textbf{Why none of this depends on the parameterization.} Each quantity above is defined from the
set of distinct classifiers, here $\Om$, and not from how they are written down. Two programs with
the same least model are the same hypothesis to a learning theorist. This is why the paper's
learning results are standard once $\Om$ is known; the work lies in constructing $\Om$ from the
template, since a learner is never given $\Om$ directly.

\subsection{One example through every quantity}\label{app:dict-example}

Take three atoms $r_0,r_1,r_2$, with $r_0$ declared true in $\BK$, and the template of three
edge rules
\[
L=\{\,a:\ r_1\leftarrow r_0,\quad b:\ r_2\leftarrow r_0,\quad c:\ r_2\leftarrow r_1\,\},\qquad N=3 .
\]
This is the bandwidth-$2$ template on $n=2$ atoms. Its $2^3=8$ clause-sets have only four
meanings:
\begin{center}\footnotesize
\begin{tabular}{@{}llccc@{}}
\toprule
content & clause-sets $S$ & fibre & pattern & clause-prior\\
$R=\mu(S)$ & with this meaning & $|\mu^{-1}(R)|$ & on $(r_1,r_2)$ & regret (bits)\\
\midrule
$\{r_0\}$ & $\emptyset,\ \{c\}$ & 2 & $(0,0)$ & $\log_2(8/2)=2$\\
$\{r_0,r_1\}$ & $\{a\}$ & 1 & $(1,0)$ & $\log_2(8/1)=3$\\
$\{r_0,r_2\}$ & $\{b\},\ \{b,c\}$ & 2 & $(0,1)$ & $\log_2(8/2)=2$\\
$\{r_0,r_1,r_2\}$ & $\{a,b\},\ \{a,c\},\ \{a,b,c\}$ & 3 & $(1,1)$ & $\log_2(8/3)=1.42$\\
\bottomrule
\end{tabular}
\end{center}
Rule $c$ alone does nothing, since $r_1$ is never derived; with $b$ present it is redundant. Every
quantity of the paper can be read off this table.
\begin{itemize}[leftmargin=*,itemsep=1pt]
\item \emph{Realizability (Proposition~\ref{prop:realize}).} $\{r_0,r_1\}$ is realizable; a set
such as $\{r_1\}$ is not, because it omits the source.
\item \emph{Capacity.} $|\Om|=4$, so $\CL=2$ bits, against $N=3$ syntactic bits.
\item \emph{VC and Littlestone dimension.} All four patterns on $\{r_1,r_2\}$ occur, so these two
atoms are shattered: $\VC=\Ldim=\CL=2$. In larger templates the three separate
(Remark~\ref{rem:constant}).
\item \emph{Halving.} At most $2$ mistakes to identify the target, whatever the query order.
\item \emph{Log loss.} With a uniform prior on the four contents, the Bayes mixture pays exactly
$2$ bits on every target after both atoms are queried (Theorem~\ref{thm:vsp}). With the prior
induced by a uniform choice of clause-set it pays the last column: $1.42$ to $3$ bits, $2.10$ on
average over contents: more than $\CL$ on the thin-fibre content $\{r_0,r_1\}$, less on the
heavily redundant $\{r_0,r_1,r_2\}$.
\item \emph{Random rules and semantic compression.} If each rule is on independently with probability $1/2$,
the content has distribution $\pi=(2,1,2,3)/8$, so $H(R)=1.91$ bits while $H(S)=3$ bits; the
semantic compression gain $H(S\mid R)=1.09$ bits is the part of the random program its meaning
makes redundant
(Remark~1 and Theorem~\ref{prop:collapse}). Averaged over its own prior, the clause-set
mixture pays $H(\pi)=1.91<\CL$ bits: the clause prior is a good bet exactly when targets really
come from random rules (Section~\ref{sec:exp}).
\end{itemize}

\subsection{The dictionary}\label{app:dict-table}

\begin{table}[h]
\centering\small
\begin{tabular}{@{}p{0.40\linewidth}p{0.55\linewidth}@{}}
\toprule
\textbf{machine learning} & \textbf{logic programming / this paper}\\
\midrule
\multicolumn{2}{@{}l}{\emph{Objects}}\\
instance space $\mathcal X$ & Herbrand base $\Bb$ (queried on the content sub-base)\\
instance $x$ & a ground atom $\gamma$ (a variable-free fact)\\
label $y\in\{0,1\}$ & entailment bit $y=\mathbf1[\gamma\in R^\star]$\\
architecture & template (language bias) $L$ with background $\BK$\\
parameters / weights & clause-set $S\in2^L$, a vector of $N$ switches\\
forward pass & forward chaining, $\mu(S)=\lhm(\BK\cup S)$\\
semantic function & semantics $\mu$; the meaning of $S$ is $\mu(S)$\\
parameter symmetry (dead or permuted units) & rules that never fire or are redundant (fibres of $\mu$)\\
hypothesis / concept class $\mathcal H$ & content class $\Om=\im\mu$ (never $2^L$)\\
hypothesis $h$ & content $R\in\Om$, read as $h_R(\gamma)=\mathbf1[\gamma\in R]$\\
target concept $h^\star$ & target theory $R^\star$; realizable iff $R^\star\in\Om$\\
\midrule
\multicolumn{2}{@{}l}{\emph{Processes}}\\
online prediction & classifying query atoms one at a time\\
realizable learning & noise-free ILP: some program in the bias explains the data\\
agnostic learning & ILP with noisy or inconsistent examples\\
version space & programs covering all positives and no negatives (up to meaning)\\
prior over hypotheses & prior over programs (Bayesian/MDL program induction)\\
\midrule
\multicolumn{2}{@{}l}{\emph{Quantities}}\\
$\log_2|\mathcal H|$ (Occam/Halving bound) & semantic capacity $\CL=\log_2|\Om|$\\
number of parameters & $N=|L|$, the syntactic bits\\
VC dimension & most atoms whose truth values the rules can set independently\\
Littlestone dimension & depth of the hardest adversarial questioning strategy\\
minimax log-loss regret (Shtarkov) & $\CL$, once the queries separate the contents (Theorem~\ref{thm:shtarkov})\\
code length & bits needed to describe which theory is true\\
\bottomrule
\end{tabular}
\caption{Machine learning $\leftrightarrow$ logic dictionary. The map $\mu:2^L\to\Om$ is the
parameter-to-classifier map; semantically equal clause-sets are one hypothesis.}
\label{tab:dict-objects}
\end{table}

\subsection{Words that mean different things on the two sides}\label{app:dict-false}

\begin{description}[leftmargin=0pt,itemsep=2pt,font=\normalfont\itshape]
\item[Hypothesis.] In ILP, a program (a set of clauses). In learning theory, a function. In this
paper a hypothesis $S$ is a program, and the function it computes is its content $R=\mu(S)$; the
hypothesis \emph{class} of learning theory is $\Om$, not $2^L$.
\item[Model.] In logic, an interpretation that makes a program true; the least Herbrand model is
the smallest one. In machine learning, a trained predictor. We use ``model'' only in the logical
sense.
\item[Capacity.] In learning theory, a complexity measure of a class, such as VC dimension. In
information theory, the maximal mutual information of a channel. Our $\CL=\log_2|\Om|$ is both a
log-cardinality and the capacity of the noiseless closure channel $S\mapsto R$ (Remark~1).
\item[Semantics, meaning, content.] The semantics is the map $\mu$ from programs to meanings,
fixed by the background rules; the meaning (content) of one program $S$ is its image $R=\mu(S)$, the
set of facts it deduces; its \emph{extent} $|R|$ is how many there are. Capacity counts distinct
contents; extent measures the size of one.
\item[Semantic compression.] Not the compression of text or messages studied in semantic
communication, but the identity $H(S)=H(R)+H(S\mid R)$: an optimal code for a random program uses
$H(S)$ bits, an optimal code for its meaning only $H(R)$ (it identifies the meaning without listing
everything that follows from it), and the \emph{semantic compression gain} $H(S\mid R)$ is the
redundant syntax the meaning removes (Remark~1, Theorem~\ref{prop:collapse}).
\item[Bandwidth.] The largest index distance a single rule can bridge in the reachability template
$L_{n,d}$ (Section~\ref{sec:count}), as in the bandwidth of a banded matrix. It is the locality of
the language bias, not a communication bandwidth.
\item[Closure.] The set of all consequences, here the least Herbrand model, as in ``deductive
closure''. For the transitive-closure template it coincides with the graph-theoretic transitive
closure.
\item[Grounding.] Replacing variables by constants. A template with variables is shorthand for its
finite set of ground clauses, and $N$ counts ground clauses.
\item[Realizable.] In learning theory, the target lies in the class. In this paper also: a set of
atoms is a \emph{realizable closure} if some clause-set deduces exactly it. The two uses agree:
the problem is realizable iff the target is a realizable closure.
\end{description}

\end{document}